\PassOptionsToPackage{sort&compress}{natbib}
\documentclass{article}
\usepackage{corl_2023} % Use this for the initial submission.
\usepackage{amsmath,amsfonts,amssymb}
\usepackage{notoccite}
\usepackage[ruled,vlined]{algorithm2e}
\usepackage{multicol}
\usepackage{xcolor}
\usepackage{svg}
\usepackage{subcaption}
\usepackage{multirow}
\usepackage[font=small,skip=2pt]{caption}
\usepackage{float}
\usepackage{diagbox}
\usepackage{bm}
\usepackage{changepage}
\usepackage{enumitem}
\setitemize{label=\textbullet, leftmargin=15pt, nolistsep}

\begin{document}
\nolinenumbers
% \title{KODex: Learning Koopman Operators for \\ Dexterous Manipulation }
% \title{KODex: Using Koopman Operator Theory to Learn Dexterous Manipulation Skills from Demonstrations}
% \title{KODex: Encoding Dexterous Manipulation Skills using Learned Koopman Operators}
% \title{Is Lifted Linearization All You Need for Learning Dexterous Manipulation Skills?}
\title{On the Utility of Koopman Operator Theory in Learning Dexterous Manipulation Skills}
\maketitle

\begin{abstract}
% This paper study on utility of Koopman operator theory for high-dimensional dexterous manipulation tasks. We conduct extensive experiments on KODex 
Despite impressive dexterous manipulation capabilities enabled by learning-based approaches, we are yet to witness widespread adoption beyond well-resourced laboratories.
% a few well-resourced laboratories. 
This is likely due to practical limitations,
% However, these advances are only accessible to those with significant resources due to a number  of practical barriers, 
such as significant computational burden, inscrutable learned behaviors, sensitivity to initialization, and the considerable technical expertise required for implementation. 
In this work, we investigate the utility of Koopman operator theory in alleviating these limitations. Koopman operators are simple yet powerful control-theoretic structures to represent complex \textit{nonlinear} dynamics as \textit{linear} systems in higher dimensions. 
Motivated by the fact that complex nonlinear dynamics underlie dexterous manipulation, we develop a Koopman operator-based imitation learning framework to learn the desired motions of both the robotic hand and the object simultaneously. 
We show that Koopman operators are surprisingly effective for dexterous manipulation and offer a number of unique benefits. 
Notably, policies can be learned \textit{analytically}, drastically reducing computation burden and eliminating sensitivity to initialization and the need for painstaking hyperparameter optimization.
% Second, the learned reference dynamics can be combined with a \textit{task-agnostic} tracking controller such that task changes and variations can be handled with ease.
Our experiments reveal that a Koopman operator-based approach can perform comparably to state-of-the-art imitation learning algorithms in terms of success rate and sample efficiency, while being \textit{an order of magnitude} faster. Policy videos can be viewed at \href{https://sites.google.com/view/kodex-corl}{https://sites.google.com/view/kodex-corl}.
% In addition, we discuss a number of avenues for future research exposed by our work. 
\end{abstract}
% Koopman learning for high-dimensional robotic system, including hybrid systems
% Koopman linear evolution is good for optimal control or contraction analysis with performance guarantee
% Koopman states space embed the object states (task-related, like policy-based RL method), combined the underlying the data-driven dynamics system with the policy mapping(state -> action)
% We consider it as an autonomous system without control input
% Demonstrate on new robot hand

\section{Introduction}
%%%%%%%%%%%%%%%%%%%% Start HR draft %%%%%%%%%%%%%%%%%%%%
% Dexterous manipulation is an important and useful problem
Autonomous dexterous manipulation is a necessary skill for robots operating in a physical world built by and for humans. However, achieving reliable robotic dexterous manipulation skills has been a long-standing challenge~\cite{okamura2000overview} due to numerous factors, such as complex nonlinear dynamics, high-dimensional action spaces, and expertise required to design bespoke controllers.
% that do not generalize to task variations.

% Recent learning-based approaches show a lot of promise
Over the past decade, learning-based solutions have emerged as promising solutions that can address the challenges in acquiring dexterous manipulation skills. Indeed, these methods have been shown to be capable of impressive feats, such as solving Rubik's cubes~\cite{OpenAI2018Cube}, manipulation Baoding balls~\cite{nagabandi2020deep}, retrieving tool trays~\cite{xieneural}, and reorienting complex objects~\cite{chen2021a}.
\begin{figure*}[t]
\centering
\includegraphics[width=\columnwidth]{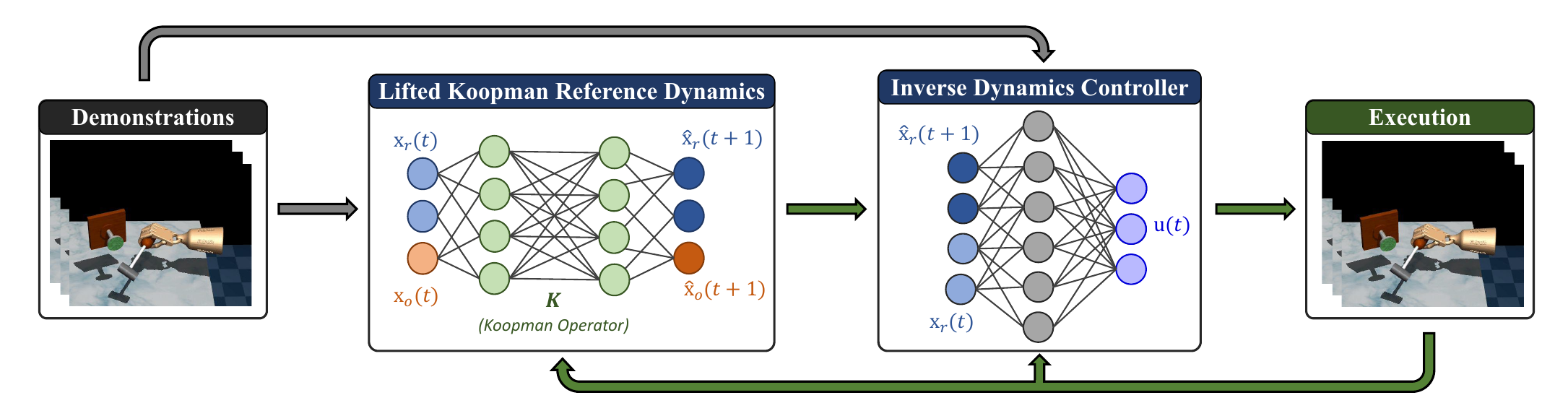}
% \includesvg{images/KODex_block_diagram.svg}
\caption{KODex simultaneously encodes complex nonlinear dynamics of the desired motion of both the robot state ($\mathrm{x}_r$) and the object state ($\mathrm{x}_o$) as a \textit{linear} dynamical system in a higher-dimensional space by learning a Koopman operator $\bm{K}$ directly from demonstrations. Further, KODex learns an inverse dynamics controller to track the robot reference trajectory ($\{\hat{\mathrm{x}}_r(t)\}_{t=1}^T$) generated by the lifted linear system.}
\label{fig:framework}
\end{figure*}
% Existing methods are computationally expensive, hard to implement, and require painstaking hyper parameter tuning
However, existing learning approaches suffer from practical limitations that hinder their widespread adoption. First, implementing existing algorithms requires significant technical expertise and modern machine learning infrastructure, \textcolor{black}{such as the advanced computing resources (e.g., GPUs), and the well-established deep learning softwares (e.g., PyTorch).} Second, training policies consume significant computational resources. \textcolor{black}{Third, while existing approaches have achieved impressive SOTA performance, these results tend to require painstaking efforts to tune hyperparameters and architectures}. Fourth, performance tends to be highly sensitive to parameter initialization. 
% (see Section~\ref{sec:related_work} for a detailed discussion).

% In this work, we explore the use of Koopman operator theory - How Koopman captures the underlying nonlinear dynanmics as linear dynamics in lifted spaces (kernal trick for dynamical system)
In this work, we investigate the utility of Koopman operator theory in alleviating the limitations of existing learning-based approaches as identified above. The Koopman operator theory helps represent arbitrary \textit{nonlinear} dynamics in finite dimensional spaces as \textit{linear} dynamics in an infinite-dimensional Hilbert space~\cite{Koopman1931Koopman}. While this equivalence is exact and fascinating from a theoretical standpoint, it is not tractable. However, recent advances have enabled the approximation of this equivalence in higher but \textit{finite}-dimensional spaces by learning the operator directly from data~\cite{williams2015data}. 

% Why Koopman? Desired robot trajectory *and* object trajectory can be seen solutions to underlying dynamics.
We develop a novel imitation learning framework, dubbed \textit{Koopman Operator-based Dexterous Manipulation (KODex)}, to evaluate the utility of Koopman operator theory for dexterous manipulation (see Fig. \ref{fig:framework}). Specifically, we model desired behaviors as solutions to nonlinear dynamical systems and learn Koopman operators that define approximately-equivalent linear dynamics in higher-dimensional spaces. 
% While we use polynomial functions to lift the state space in this work, KODex is agnostic to the specific lifting function.
Note that it is insufficient to exclusively focus on the robot's motion as the objective of dexterous manipulation is centered on the object's motion~\cite{okamura2000overview}.
% Dexterous manipulation is concerned more directly with the movement of the object, and only implicitly with that of the robot~\cite{okamura2000overview}.
% Since dexterous manipulation is object-centric, the desired motion of the robot must also depend on the object state. 
As such, KODex \textit{simultaneously} learns the desired motions of both the robot and the object from demonstrations. To eliminate the need for an expertly-tuned \textcolor{black}{PD} controller, KODex relies on a learned inverse dynamics controller to track the reference trajectory generated by the learned dynamical system. 
% Note that the tracking controller is \textit{task-agnostic}, and as such, can be preserved even if we wish to learn an entirely new task on the same robot platform.
% and can be learned via self supervision.

% Benefits of Koopman include analytical solution, low computational cost, low variance/predictablity/easy to inspect. 
% KODex offers a number of unique benefits.
% A number of reasons motivate our exploration of Koopman operator theory within the context of dexterous manipulation. 
A significant benefit of learning Koopman operators from data is that it lends itself to an \textit{analytical} solution. As such, KODex is simple to implement and does not require expertise and familiarity with state-of-the-art (SOTA) machine learning infrastructure. Instead of painstaking hyperparameter optimization, we show that generic and task-agnostic polynomial lifting functions are sufficient for KODex to learn diverse dexterous manipulation skills.
% More importantly, KODex incurs significantly lower computational costs to existing approaches, and does not rely on hyperparameters that have to be painstakingly tuned by an expert. 
Further, KODex offers consistent and predictable performance since the learning process is analytical and thus not sensitive to parameter initialization. Finally, given that KODex learns a \textit{linear} dynamical system, one can readily inspect the learned behaviors using a wide array of control theoretic tools.

% Our results demonstrate that ...
We carry out extensive evaluations of KODex within the context of four dexterous manipulation skills on the simulated Adroit hand, an established experimental platform for dexterous manipulation~\cite{Rajeswaran2018DAPG}. Further, we compare KODex against SOTA imitation learning approaches in terms of general efficacy, computational efficiency, and sample efficiency.
% and robustness to changes to physical properties that could result from sim-to-real gap or task variations. 
% task success rate, imitation error, training time, and sample efficiency. 
Our results demonstrate that KODex is at least an \textit{order of magnitude} faster to train than SOTA imitation learning algorithms, while achieving comparable sample efficiency and task success rates.
These results suggest that Koopman operators can be effective, efficient, and reliable tools to learn dexterous manipulation skills and to reduce the barriers to wide-spread adoption.

\section{Related Work}\label{sec:related_work}
In this section, we contextualize our contributions within relevant sub-fields. 
% and their connection to our work.

% \subsubsection*{Learning Manipulation Skills as Dynamical Systems}
% \label{sec:rw_learning_DS}
% talk about dynamical systems for 7DoF arm , not high-dimsional
% to learn dynamical system, instead of learning in action space
\textbf{Learning Manipulation Skills as Dynamical Systems}: 
Our work falls into the category of dynamical-system-based imitation learning methods for manipulation~\cite{Harish2022Survey}. Made popular by the Dynamics Movement Primitives (DMPs)~\cite{ijspeert2013dynamical}, these methods model robot motions as solutions to a learnable dynamical system. The past decade witnessed a plethora of approaches built upon the same principle (e.g., \cite{khansari2011learning,neumann2015learning,Harish2017Dynamical,rana2020learning, rana2020euclideanizing,figueroa2022locally}), creating increasingly-capable LfD tools for manipulation. Robustness to perturbations, high sample efficiency, and provable convergence are all but a few examples of the many advantages of dynamical-system-based approaches. These approaches tend to be highly structured and leverage control-theoretic and topological tools to learn complex desired motions with unparalleled sample efficiency.
Recent work also embeded the dynamical systems structure into deep neural networks to enable end-to-end learning~\cite{bahl2020neural}.
% In fact, most of these approaches have demonstrated that goal-directed motions can be learned from as few as six demonstrations. 
% Recently, \citet{bahl2020neural} developed Neural Dynamic Policy (NDP) which embeds the dynamical systems structure into deep neural network-based policies to enable end-to-end learning.
These approaches were primarily designed to capture low-dimensional end-effector skills for serial-link manipulators. 
% As such, they deal with low-dimensional systems with limited degrees-of-freedom (DOFs). We discuss a few recent exceptions to this limitation in Section~\ref{sec:rw_learning_DM}. 
In contrast, our work investigates the utility of Koopman operators in learning dexterous manipulation skills on high-DOF platforms.
\textbf{Learning Dexterous Manipulation Skills}:
\label{sec:rw_learning_DM}
Deep Reinforcement Learning (RL) has been dominating the field of dexterous manipulation recently, enabling an impressive array of skills~\cite{nagabandi2020deep, qi2022hand, zhu2019dexterous}. A popular approach demonstrated that a multi-finger hand can learn to solve the Rubik's cube~\citet{OpenAI2018Cube}. Recently, \citet{chen2021a} developed a model-free RL framework capable of reorienting over 2000 differently-shaped objects. 
% and showed strong zero-shot transfer to new objects. 
% Several other recent approaches have demonstrated the effectiveness of RL for dexterous manipulation. 
Despite impressive successes, RL-based algorithms suffer from poor sample efficiency and notoriously-difficult training procedures. 
In contrast, Imitation learning (IL) aims to improve sample efficiency by leveraging expert demonstrations~\cite{osa2018algorithmic,Harish2022Survey}. However, most existing IL-based methods (including those discussed in Section~\ref{sec:rw_learning_DM}) focus on lower-DOF manipulators and do not scale well to high-dimensional systems. Indeed, there are at least two recent notable exceptions to this limitation: \citet{xieneural} developed a highly structured IL method to learn dexterous manipulation skills from demonstrations in the form of a dynamical system; \citet{arunachalam2022holo} introduced a mixed-reality framework to collect high-quality demonstrations and learn dexterous manipulation skills by leveraging visual representations and motion retargeting. 
Researchers have also combined IL with RL to get the best of both approaches and has been able to achieve impressive performance (e.g.,~\cite{kumar2016learning,Rajeswaran2018DAPG}).  
A common attribute of all existing learning approaches to dexterous manipulation
% these methods (irrespective of whether they leverage RL, IL, or both) 
is that they rely on 
% large neural networks that have to be trained via numerical optimization. As such, typically require 
significant computational resources and user expertise for implementation and hyperparameter tuning.
% there are much fewer IL-based approaches focused on dexterous manipulation (\cite{xieneural, arunachalam2022holo}), perhaps due to the prohibitive cost involved in collecting a large number of demonstrations. 
Further, the effectiveness of these approaches is highly sensitive to parameter initialization~\cite{henderson2018deep}. 
In stark contrast, KODex encodes complex skills as dynamical systems which are \textit{analytically} extracted from demonstrations. As such, KODex incurs a significantly smaller computational burden and eliminates the dependence on painstaking hyperparameter tuning and unreliable numerical optimization. Further, unlike opaque deep neural networks, KODex learns \textit{linear} dynamical systems that can be inspected via control-theoretic tools.

% \subsubsection*{Koopman Operator for Robotics}
% talk about the Koopman robotics work, not high-dimensional
\textbf{Koopman Operators in Robotics}:
Recently, Koopman operator theory has proven beneficial in various robotic systems, such as differential drive robots~\cite{abraham2017model}, spherical and serial-link manipulators~\cite{Abraham2019ActiveKoopman}, autonomous excavators~\cite{Sotiropoulos2022Bucket, Selby2021Excavation}, and soft robotic manipulators~\cite{bruder2019modeling,Bruder2021Soft}. However, the systems investigated in these works are low-dimensional. In contrast, our work is focused on evaluating the effectiveness of Koopman operators in learning skills for a high-dimensional system with complex dynamics (i.e., a multi-fingered hand). Further, prior works have not sufficiently investigated the relative benefits of leveraging Koopman operators over SOTA neural network-based approaches, and the circumstances under which these benefits hold.
% This leaves us a question of whether or under what tasks/metrics Koopman operator can better model complex system dynamics than neural networks. Therefore, 
In our work, we thoroughly evaluate KODex against SOTA imitation learning methods within the context of multiple dexterous manipulation tasks.
% in terms of various performance metrics.
% acknowledge that our method belongs to learning dynamics from demonstrations and there are lots of cool work along this direction

%
% \vspace{-0.3cm}
\section{Preliminary: Koopman Operator Theory}
We begin by providing a brief introduction to Koopman operator theory~\cite{Koopman1931Koopman}.
% -- a control-theoretic approach used to represent nonlinear dynamical systems as a linear system of equations~\cite{Koopman1931Koopman}.
% We first describe the infinite-dimensional Koopman operator representation of the autonomous dynamical system in Section.~\ref{sec:koopman}.
% Then, we present a practical system identification method to extract the matrix approximation of the Koopman operator on the finite-dimensional lifted state space via linear regression in Section.~\ref{sec:approximation}.

\textbf{Koopman Representation}:
Consider a discrete-time autonomous nonlinear dynamical system
\begin{equation}
\label{eqn:system}
\mathrm{x}(t+1) = F(\mathrm{x}(t)), 
\end{equation}
where $\mathrm{x}(t) \in \mathcal{X} \subset \mathbb{R}^n$ is the state at time $t$, and $F(\cdot):\mathbb{R}^n \rightarrow \mathbb{R}^n$ is a nonlinear function.

To represent the the nonlinear dynamical system in (\ref{eqn:system}) as a linear system, we begin by introducing a set of \textit{observables} using the so-called \textit{lifting function} $g:\mathcal{X} \rightarrow \mathcal{O}$,
where $\mathcal{O}$ is the space of observables. We can now define the \textit{Koopman Operator} $\mathcal{K}$, an infinite-dimensional operator on the lifting function $g(\cdot)$ for the discrete time system defined in (\ref{eqn:system}) as follows
% \begin{equation}
% \label{eqn:KG}
% [\mathcal{K}g] = g \circ F(\mathrm{x}(t))
% \end{equation}
% where $\circ$ is the compositional operator. For the discrete time system defined in (\ref{eqn:system}), the above equation can be rewritten as
\begin{equation}
\label{eqn:KG_2}
[\mathcal{K}g] = g(F(\mathrm{x}(t))) = g(\mathrm{x}(t+1))
\end{equation}
If the observables belong to a vector space, the Operator $\mathcal{K}$ can be seen as an infinite-dimensional linear map that describes the evolution of the observables as follows
\begin{equation}
\label{eqn:observable_system}
g(\mathrm{x}(t+1)) = \mathcal{K}g(\mathrm{x}(t))
\end{equation}
% Now, let's define the lifting function as , which maps original states from $X$ to lifted states $\mathbb{C} \in \mathbb{R}^p$. Put it in other words, $g(\mathrm{x}_t)$ is a vector of $p$ scalar-valued functions $\phi(\mathrm{x}_t)$:
% \begin{equation}
% \label{eqn:gx}
% g(\mathrm{x}_t) = [\phi_1(\mathrm{x}_t), \phi_2(\mathrm{x}_t), \ldots, \phi_p(\mathrm{x}_t)]^\top,
% \end{equation}
% If $p = \infty$, the Koopman operator $\mathcal{K}$ is an infinite dimensional operator that can directly act on the element of $\mathbb{C}$ to describe the system evolution in the lifted state space
% As long as $g(\mathrm{x}_t)$ is defined in vector space as illustrated in Eqn.~\ref{eqn:gx}, $\mathcal{K}$ is a linear operator, so Eqn.~\ref{eqn:KG} can be rewritten as
% \begin{equation}
% \label{eqn:newKG}
% % [\mathcal{K}g](\mathrm{x}_t) = 
% g(\mathrm{x}_{t+1}) = \mathbf{K}g(\mathrm{x}_t),
% \end{equation}
% where $\mathbf{K} \in \mathbb{R}^{p \times p}$. 
Therefore, the Koopman operator $\mathcal{K}$ linearly propagates forward the infinite-dimensional lifted states (i.e., observables). 
In practice, we do not benefit from this representation since it is infinite-dimensional. However, we can approximate $\mathcal{K}$ using a matrix $\mathbf{K}\in\mathbb{R}^{p \times p}$ and define a finite set of observables $\phi (t) \in \mathbb{R}^p $. Thus, we can rewrite the relationship in (\ref{eqn:observable_system}) as
\begin{equation}
\label{eqn:approximate_KG}
 \phi(\mathrm{x}(t+1)) =  \mathbf{K}\phi(\mathrm{x}(t)) + r(\mathrm{x}(t)),
\end{equation}
where $r(\mathrm{x}(t)) \in \mathbb{R}^p$ is the residual error caused by the finite dimensional approximation, which can be arbitrarily reduced based on the choice of $p$. 

\textbf{Learning Koopman Operator from Data}:
The matrix operator $\mathbf{K}$ can be inferred from a dataset $D = [\mathrm{x}(1), \mathrm{x}(2), \cdots, \mathrm{x}(T)]$, which contains the solution to the system in (\ref{eqn:system}) for $T$ time steps.
% Now, suppose we have $N$ pieces of demonstration data and each piece is the full trajectory of system evolution, which is assumed to be an exact solution of Eqn.~\ref{eqn:newsystem}. Each piece of demonstration data consists of $T$ datapoints, where $T$ represents the task-specific time horizon. One datapoint contains the robot states and object states at one time step $t$. The demonstration data $D$ can be defined as
% \begin{equation}
% D = [\{\hat{\mathrm{x}}_t^{1}\}^{t=T}_{t=1}, \{\hat{\mathrm{x}}_t^2\}^{t=T}_{t=1}, \cdots, \{\hat{\mathrm{x}}_t^N\}^{t=T}_{t=1}],
% \end{equation}
% It should be noted that a recorded $n$th trajectory with missing datapoints is also acceptable, as long as all the state pairs $(\hat{\mathrm{x}}_t^n, \hat{\mathrm{x}}_{t+1}^n)$ are temporally consecutive. Additionally, we also record the actuated torque $\mathrm{\tau}_t$ in each datapoint for the controller design discussed in Section.~\ref{sec:control}. 
Given the choice of observables $\phi(\cdot)$, the finite dimensional Koopman matrix $\mathbf{K}$ is computed by minimizing the approximation error defined in (\ref{eqn:approximate_KG}). Specifically, we can obtain $\mathbf{K}$ from $D$ by minimizing the cost function $\mathbf{J}(\mathbf{K})$ given below
\begin{equation}
\label{eqn:goal}
\mathbf{J}(\mathbf{K}) = \frac{1}{2}\sum_{t=1}^{t=T-1} \Vert r(\mathrm{x}(t)) \Vert ^2 =\frac{1}{2}\sum_{t=1}^{t=T-1} \Vert \phi(\mathrm{x}(t+1)) - \mathbf{K}\phi(\mathrm{x}(t)) \Vert ^2
\end{equation}
% via minimizing $\mathcal{J}$, which is the $L_2$ sum of residual error $r(\mathrm{x}_t)$ on all datapoints in $D$

Note minimizing $\mathbf{J}(\mathbf{K})$ amounts to solving a least-square problem, whose solution is given by~\cite{williams2015data}
\begin{equation}
\label{eqn:k}
\mathbf{K} = \mathbf{A}\mathbf{G}^\dagger;\ \ \mathbf{A} = \frac{1}{T-1}\sum_{t=1}^{t=T-1}\phi(\mathrm{x}(t+1)) \otimes \phi(\mathrm{x}(t)),\ \  
\mathbf{G} = \frac{1}{T-1}\sum_{t=1}^{t=T-1}\phi(\mathrm{x}(t)) \otimes \phi(\mathrm{x}(t))
\end{equation}
where $\mathbf{G}^\dagger$ denotes the Moore–Penrose inverse of $\mathbf{G}$ 
\footnote{\textcolor{black}{It could be efficiently solved using scipy.linalg.pinv($\mathbf{G}$) function from Scipy library.}}, and 
% \begin{equation}
% \label{eqn:k_comp}
% \mathbf{A} = \frac{1}{T-1}\sum_{t=1}^{t=T-1}\phi(\mathrm{x}(t+1)) \otimes \phi(\mathrm{x}(t)),
% \quad   
% \mathbf{G} = \frac{1}{T-1}\sum_{t=1}^{t=T-1}\phi(\mathrm{x}(t)) \otimes \phi(\mathrm{x}(t))
% \end{equation}
$\otimes$ denotes the outer product. 

% Thus, we can utilize the computed Koopman matrix $\mathbf{K}$ to make predictions in lifted state space via linear evolution (Eqn.~\ref{eqn:approximate_KG}) starting from $g(\mathrm{x}_1)$.

% In order to capture more underlying dynamics of the original system, the higher dimension of the lifted states is preferred. Furthermore, the data distribution in $D$ has a dominant effect on the generalizability of computed Koopman matrix $\mathbf{K}$. Therefore, there is a trade-off between the number of demonstration data and the cost of date collection, especially when data is expensive, e.g. collecting data on a physical robot. To grab a hint on how much demonstration data is "enough" to obtain desirable results, we first initialize an empty data set $\bar{D}$ and another empty error list $E$ in experiments. During each iteration ($N$ in total), we add one piece of demonstration data $\{\hat{\mathrm{x}}_t^{n}\}^{t=T}_{t=1}$ selected from $D$ into $\bar{D}$ and adaptively obtain the Koopman matrix $\mathbf{K}_{\bar{D}}$ using Eqn.~\ref{eqn:k} \ref{eqn:k_comp}. Then, we compute the $L_2$ sum of residual error $e_n$ on full demonstration data $D$ using $\mathbf{K}_{\bar{D}}$ and add $e_n$ into $E$. At the last iteration, we will have $\mathbf{K}_{D} = \mathbf{K}_{\bar{D}}$.

\section{Learning Koopman Operators for Dexterous Manipulation}
% In this section, we introduce our learning framework KODex. 
We begin by introducing our framework to model dexterous manipulation skills as nonlinear dynamics and discuss the importance of incorporating object states into the system (Section~\ref{sec:modeling}). Next, we describe how KODex learns the reference dynamics for a given skill from demonstrations (Section~\ref{sec:demons}). Then, we discuss how to learn a low-level controller, also from demonstrations, in order to faithfully track the reference trajectories generated by KODex (Section~\ref{sec:control}). \textcolor{black}{Finally, we discuss policy execution (Section~\ref{sec:implementation}).} 
% we formulate the learning framework for dexterous manipulation using Koopman operator theory. First, in Section.~\ref{sec:incorp} and \ref{sec:demons}, we demonstrate the modified components of Koopman operator, which are necessary to the complex dexterous manipulation tasks.
% Therefore, the unknown nonlinear dynamical model of the dexterous manipulation system is completely constructed by the linear model in the lifted space. Besides, the hand states can be easily retrieved from the lifted states, which serve as the referenced hand trajectories. 
% Next, 
% in order to execute the trajectories on a dexterous hand, 
% we show how to separately learn an inverse dynamics controller from demonstration data in Section~\ref{sec:control}, which takes the input as the consecutive pair of robot states and outputs the actuation torques. Finally, we describe how to execute the learned Koopman operator in Section~\ref{sec:implementation}. The framework overview can be found in 
Further, an overall pseudo-code for KODex can be found in Appendices~\ref{sec:pseudo-code}.
\subsection{Modeling Dexterous Manipulation Skills}
\label{sec:modeling}
A central principle behind KODex is that the desired behavior of a robot can be represented using a dynamical system. 
% To that end, we begin by defining the state of the dynamical system. 
Note that, unlike other kinds of manipulation skills (e.g., end-effector skills of multi-link manipulators), dexterous manipulation is explicitly concerned with how an object moves as a result of the robot's motion~\cite{okamura2000overview}. As such, KODex captures the desired motion of the robot along with that of the object. To this end, we define the state at time $t$ as $\mathrm{x}(t) = [{\mathrm{x}_r(t)}^\top, {\mathrm{x}_o(t)}^\top]^\top$, where ${\mathrm{x}_r(t)} \in \mathcal{X}_r \subset \mathbb{R}^n$ and ${\mathrm{x}_o(t)} \in \mathcal{X}_o \subset \mathbb{R}^m$ represent the state of the robot and the object, respectively, at time $t$. As such, the dynamical system we wish to capture is
\begin{equation}
\label{eqn:F*}
    \mathrm{x}(t+1) = F^*(\mathrm{x}(t))
\end{equation}
where $F^*(\cdot):\mathcal{X}_r \times \mathcal{X}_o \rightarrow \mathcal{X}_r \times \mathcal{X}_o $ denotes the true dynamics that govern the \textit{interdependent} motions of the robot and the object. Note that this system is time-invariant. Indeed, time-invariant dynamical systems provide a natural way to capture manipulation skills that are more robust to intermittent perturbations than those that explicitly depend on time~\cite{Harish2022Survey}.

A key challenge in learning the dynamical system in (\ref{eqn:F*}) is that it can be arbitrarily complex and highly nonlinear, depending on the particular skill of interest. KODex leverages Koopman operator theory to learn a \textit{linear} dynamical system that can effectively approximate such complex nonlinear dynamics. To this end, we first define a set of observables as follows
\begin{equation}
\label{eqn:gx}
\phi(\mathrm{x}(t)) = [{\mathrm{x}_r(t)}^\top, \psi_r({\mathrm{x}_r(t)}), {\mathrm{x}_o(t)}^\top, \psi_o({\mathrm{x}_o(t)})]^\top, \forall t
\end{equation}
where $\psi_r:\mathbb{R}^n \rightarrow \mathbb{R}^{n\prime}$ and $\psi_o:\mathbb{R}^m \rightarrow \mathbb{R}^{m\prime}$ are vector-valued lifting functions that transform the robot and object state respectively. 
% Now, the linear dynamical system in the space of observables will take the same form as in (\ref{eqn:approximate_KG}).

In our implementation, we use polynomial functions up to a finite degree in our lifting function since polynomial functions allow for flexible definition of complex functions. However, it is important to note that KODex is agnostic to the specific choice of observables. 
Further, we do not assume that we know the ideal set of observables for any given skill. Instead, as we demonstrate in Section \ref{sec:exp_eval}, KODex can learn different dexterous manipulation skills on the same space of observables. 

\subsection{Learning Reference Dynamics}
\label{sec:demons}
% For complex tasks,
% (e.g. dexterous manipulation in this work)
% learning a robust policy requires a significant amount of system trajectories. Therefore, we deploy a well-trained RL agent \cite{Rajeswaran2018DAPG} in simulation and record its rollouts as demonstration data.

We now turn to the challenge of learning the Koopman operator $\mathbf{K}$ from demonstrations. Let 
% suppose we have $N$ demonstrated trajectories and each trajectory has the full system evolution, which is assumed to be an exact solution of (\ref{eqn:F*}). 
% Each trajectory has $T$ time steps, and $T$ depends on specific skills. At each time step $t$, we record the robot states and object states, which give us
$
D = [\{{\mathrm{x}}^{(1)}(t),\mathrm{\tau}^{(1)}(t)\}^{t=T^{(1)}}_{t=1}, \cdots, \{{\mathrm{x}}^{(N)}(t),\mathrm{\tau}^{(N)}(t)\}^{t=T^{(N)}}_{t=1}]$
denote a set of $N$ demonstrations containing trajectories of state-torque pairs.
% denote a set of $N$ demonstrations, where $\{\mathrm{x}^{(n)}(t)\}^{t=T^{(n)}}_{t=1}$ is the $n$th demonstration containing the evolution of both the robot and the object states for $T^{(n)}$ time steps.
Now, we can compute the Koopman matrix as $\mathbf{K} = \mathbf{A}\mathbf{G}^\dagger$, where $\mathbf{A}$ and $\mathbf{G}$ can be computed by modifying the expressions in (\ref{eqn:k}) as follows
\begin{equation}
\label{eqn:k_comp_new}
\mathbf{A} =\sum_{n=1}^{n=N}\sum_{t=1}^{t=T^{(n)}-1}  \frac{\phi(\mathrm{x}^{n}(t+1)) \otimes \phi(\mathrm{x}^n(t))}{N(T^{(n)}-1)},\ 
\mathbf{G} = \sum_{n=1}^{n=N}\sum_{t=1}^{t=T^{(n)}-1} \frac{\phi(\mathrm{x}^{n}(t)) \otimes \phi(\mathrm{x}^n(t))}{N(T^{(n)}-1)}   
\end{equation}
It is worth noting that KODex can also leverage partial trajectories that do not complete the task, as long as all the state pairs $(\mathrm{x}^n(t), \mathrm{x}^n(t+1))$ are temporally consecutive. Additionally, we also record the actuated torque $\mathrm{\tau}(t)$ at each time step for the controller design discussed in Section~\ref{sec:control}. 

We use the learned reference dynamics to generate rollouts in the observables space. However, we need to obtain the rollouts in the original robot states to command the robot. Since we designed $\phi(\mathrm{x}(t))$ such that both robot state $\mathrm{x}_r(t)$ is a part of the observables in  (\ref{eqn:gx}), we can retrieve the desired robot trajectory $\{\hat{\mathrm{x}}_r(t)\}$ by selecting the corresponding elements in $\phi(\mathrm{x}(t))$. 
% Note that in practice, we are only interested in the predicted robot states $\mathrm{x}_r(t+1)$ retrieved from $g(\mathrm{x}(t+1))$.

% In order to capture more underlying dynamics of the original system, the higher dimension of the lifted states ($\hat{g}(\hat{\mathrm{x}}_t)$) is preferred.

Indeed, the data distribution in $D$ has a considerable effect on the generalizability of the computed Koopman matrix $\mathbf{K}$. Therefore, there is an inevitable trade-off between the number of demonstrations and the cost of data collection - a challenge shared by most imitation learning algorithms.

\subsection{Learning a Tracking Controller}
\label{sec:control}
% also, mention the lifting functions include the original states themselves, from the first n columns.
% To execute the learned skill on the robot, we first need to retrieve the robot states from the lifted states.
% For convenience, we define
% \begin{equation}
% \label{eqn:gx}
% g(\mathrm{x}(t)) = [{\mathrm{x}_r(t)}^\top, \psi_r({\mathrm{x}_r(t)}), {\mathrm{x}_o(t)}^\top, \psi_o({\mathrm{x}_o(t)})]^\top,
% \end{equation}
% where $\psi_r$ and $\psi_o$ are both set of vector-valued lifting functions.
To track the desired trajectories generated from the learned reference dynamics, we design an inverse dynamics controller $C$ \cite{hanna2017grounded} \cite{bahl2020neural}. 
% The robot state can be read from simulators or robot sensors. 
Indeed, a \textcolor{black}{PD} controller could be used instead of learning a tracking controller. However, one would have to painstakingly tune the control gains and frequency, and do so for each task independently.

We use a multi-layer perception (MLP) as the tracking controller and train it using the recorded state-torque pairs $(\mathrm{x}_r^n(t), \mathrm{x}_r^n(t+1),\mathrm{\tau^n}(t))$ by minimizing
\begin{equation}
\label{eqn: controller}
\mathcal{L}_{\text{control}} =\sum_{n=1}^{n=N}\sum_{t=1}^{t=T^{(n)}-1} \frac{|C(\mathrm{x}_r^n(t), \mathrm{x}_r^n(t+1)) - \mathrm{\tau}^n(t)|^2}{N(T^{(n)}-1)}
\end{equation}
The learned controller takes as input the current robot state $\mathrm{x}_r(t)$ and the desired next state from the reference trajectory  $\hat{\mathrm{x}}_r(t+1)$, and generates the torque $\mathrm{\tau}(t)$ required for actuation.
% The alternative is to fine-tune a PID controller, which requires extra manual work but does not need access to the demonstrated torques. Thus, during data collection, we only record the robot and object states and no control information is needed.
% (original) Note that the controller can also be trained without the demonstrations by collecting self-play data. 
% for each task due to the different types of hand-object interactions.
% It should be noted that the PID controller should run at a higher frequency than the trajectory generation.
% Indeed, it is also possible to train a tracking controller via reinforcement learning~\cite{peng2020learning}. Note that KODex does not require torque information to learn the reference dynamics. As such, if the controller is a PD controller or being trained via reinforcement learning, KODex can learn skills from state-only observations~\cite{torabi2019recent}.
\textcolor{black}{
\subsection{Execution}
\label{sec:implementation}
% From the previous sections, we already have the lifting function $g(\cdot)$, the computed Koopman operator $\mathbf{K}$, and the controller $C$. 
With the reference dynamics and the tracking control learned, we specify how to execute the policy in this section.
Suppose $\mathrm{x}(1) = (\mathrm{x}_r(1), \mathrm{x}_o(1))$ is the given initial state, we can then generate the reference trajectory ($\{\hat{\mathrm{x}}_r(t)\}_{t=1}^T$) by integrating the learned reference dynamics $\hat{\mathrm{x}}(t+1)=\mathbf{K}\hat{\mathrm{x}}(t)$. Further, at time step $t$, we pass the current robot state $\mathrm{x}_r(t)$ and the desired next robot state $\hat{\mathrm{x}}_{r}(t+1)$ to the learned controller $C$ to compute the required torque $\mathrm{\tau}(t)$.
% \vspace{-0.3cm}
}
\section{Experimental Evaluation}\label{sec:exp_eval}
We evaluated KODex along with existing approaches in terms of their general efficacy, computational efficiency, sample efficiency, and scalability when learning dexterous manipulation skills.
% To evaluate KODex and compare it to existing approaches, we designed and carried out a series of experiments. We detailed our experiments design and baselines in Sec.~\ref{sec:exp_design}. We then make thorough comparisons between KODex and baselines in term of its i) general efficacy (Sec. \ref{sec:general_efficacy}), ii) sample efficiency (Sec.~\ref{sec:sample_efficiency}). We lastly mention the additional experiments in Sec.~\ref{sec:additional_exp}.

\subsection{Experimental Design}\label{sec:exp_design}

\textbf{Evaluation Platform}: 
We conducted all our experiments on the widely-used ADROIT Hand~\cite{Rajeswaran2018DAPG} -- a 30-DoF simulated system (24-DoF hand + 6-DoF floating wrist base) built with MuJoCo~\cite{Todorov2012MUJuCO}.

% The goal is to efficiently learn the Koopman operators and inverse dynamics controllers from the demonstration data. 

\textbf{Baselines}: 
We compared KODex against the following baselines:
\begin{itemize}
    \item \textit{NN}: Fully-connected neural network policy
    \item \textit{LSTM}: Recurrent neural network policy with Long Short-Term Memory (LSTM) units
    \item \textit{NDP}: Neural Dynamic policy~\cite{bahl2020neural}
    % \item \textit{RMP}: Riemannian Motion policy \cite{mukadam2020riemannian}
    \item \textit{NGF}: Neural Geometric Fabrics policy~\cite{xieneural}
    % \item \textit{NPG}: Natural Policy Gradient policy \cite{kakade2001natural}
    % \item \textit{DAPG}: Demo Augmented Policy Gradient policy \cite{Rajeswaran2018DAPG}
\end{itemize}
Note that NN and LSTM are unstructured baselines which help question the need for structured policies. NDP and NGF are highly-structured SOTA imitation learning methods for manipulation. 
% We evaluated each policy's performance in modeling complex reference dynamics associated with dexterous manipulation skills. 

We undertook several precautions to ensure a fair comparison. First, we designed the robot and object state space for all baselines and KODex to be identical. Second, we carefully designed the baselines policies and tuned their hyper-parameters for each baseline method (Appendices~\ref{sec:policy_designs} and \ref{appendix:layer_size}).
% (Appendices~\ref{sec:policy_designs} and \ref{appendix:layer_size}). 
Third, we trained each baseline policy over five random seeds to control for initialization effects. For all tasks, we saved the baseline policies that performed the best on a validation set of 50 held-out demonstrations. Note that KODex utilizes an analytical solution and thus does not require parameter initialization or hyper parameter optimization.

\begin{figure*}
     \centering
     % \begin{subfigure}[b]{1.0\textwidth}
     %    \centering
    \includegraphics[width=0.95\linewidth]{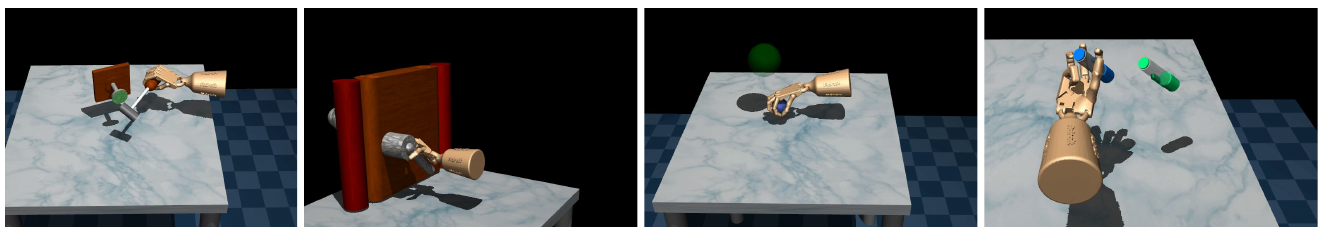}
     % \end{subfigure}
     \caption{We evaluate KODex on four tasks from \cite{Rajeswaran2018DAPG}: Tool Use, Door Opening, Relocation, and Reorientation.}
      \label{fig:tasks}
\end{figure*}

% \vspace{3pt}

\textbf{Tasks}: 
We evaluated all algorithms on a set of four tasks originally proposed in \cite{Rajeswaran2018DAPG} (see Fig. \ref{fig:tasks}).
% \begin{figure}[t!]
%      \centering
%      \begin{subfigure}[b]{0.5\textwidth}
%         \centering
%         \includegraphics[width=1\linewidth, height = 2cm]{images/pen.pdf}
%         \caption{In-hand reorientation - Reorient the blue pen to the goal orientation represented by the green pen.}
%         \label{fig:pen}
%      \end{subfigure}
%      \hfill
%      \begin{subfigure}[b]{0.5\textwidth}
%         \centering
%         \includegraphics[width=1\linewidth, height = 2cm]{images/relocate.pdf}
%         \caption{Object relocation - Move the blue ball to the green target.}
%         \label{fig:relocate}
%      \end{subfigure}
%      \hfill
%      \begin{subfigure}[b]{0.5\textwidth}
%         \centering
%         \includegraphics[width=1\linewidth, height = 2cm]{images/door.pdf}
%         \caption{Door opening - Undo the latch and drag the door open.}
%         \label{fig:door}
%      \end{subfigure}
%      \hfill
%      \begin{subfigure}[b]{0.5\textwidth}
%         \centering
%         \includegraphics[width=1\linewidth, height = 2cm]{images/hammer.pdf}
%         \caption{Tool use - Pick up the hammer and use it to forcefully drive the nail into the board.}
%         \label{fig:tool}
%      \end{subfigure}
%      \caption{An illustration of different tasks adopted from \cite{Rajeswaran2018DAPG}. Each cases shown above are considered successful and the ADROIT Hand is actuated using our approach as detailed in Alg.~\ref{alg: overview}.}
% \end{figure}
\begin{itemize}
    \item \textit{Tool use}: Pick up the hammer to drive the nail into the board placed at a randomized height.
    % The initial poses of the hand and the hammer are fixed. Given a randomized nail position, the task is considered successful if the entire nail is inside the bored. The time horizon $T$ for this task is 100.
    \item \textit{Door opening}:
    Given a randomized door position, undo the latch and drag the door open.
    % the task is considered successful if the door opening angle is larger than a threshold. The time horizon $T$ for this task is 70.
    \item \textit{Object relocation}: Move the blue ball to a randomized target location (green sphere).
    % (second right scene in Fig.~\ref{fig:tasks}). 
    % The initial position of the blue ball is fixed. Given a randomized target, the task is considered successful if the position difference is within tolerance. The time horizon $T$ for this task is 100.
    \item \textit{In-hand reorientation}:  Reorient the blue pen to a randomized goal orientation (green pen).
    % (right-most scene in Fig.~\ref{fig:tasks}). 
    % The task is considered successful if the orientation difference is within tolerance. The time horizon $T$ for this task is 100.
\end{itemize}
See Appendices~\ref{Appendix:state_design} for the state space design of all tasks.
For all tasks, the reference dynamics was queried at 100HZ, and the controller ran at 500HZ.
% For each task, we consider an execution to be successful if the following criteria are met within the time horizon. 
% \textit{In-hand Reorientation}: the robot manipulates the object to the goal orientation; \textit{Object Relocation}: the robot grasps the object from the table and moves it to the target position; \textit{Door Opening}: the robot opens the door; \textit{Tool Use}: the robot grasps the hammer and uses it to drive the nail into the board. The quantitative criteria for each task used in programming are detailed in Appendix.~\ref{Task_Success}.

\textbf{Metrics}: 
% \label{subsec:metrics}
% We use the following metrics to evaluate KODex and the baseline methods:
We quantify performance in terms of i) \textit{Training time}: Time taken to train a policy, ii) \textit{Imitation error}: The $L1$ distance between generated joint trajectories and the demonstrations, and iii) \textit{Task success rate}: Percentage of successful trials (see Appendices~ \ref{appendix:Task_Success_criteria} for success criteria).

% \begin{itemize}
% \item \textit{Training time}: The number of seconds taken by each method to train a policy.
% \item \textit{Imitation error}: The $L1$ distance between generated joint trajectories and the demonstrations.
% \item \textit{Task success rate}: Percentage of successful trials (see Appendix \ref{appendix:Task_Success_criteria} for success criteria).
% \end{itemize}

% Further, we investigate each method's \textit{sample efficiency} by reporting how the above metrics vary as a function of the number of demonstrations.
% For baseline methods, we report the mean and mean $\pm$ standard deviation of all three metrics across 5 random seeds.

\textbf{Expert Policy}: 
For each task, we trained an expert RL agent using DAPG~\cite{Rajeswaran2018DAPG} to generate 250 expert demonstrations (200 for training and 50 for validation). See Appendices~\ref{Appendix:distribution} for further details.

\textbf{Inverse Dynamics Controller}: 
To standardize controller performance across methods, we trained a common inverse dynamics controller for each task using $250$ demonstrations (see Appendices~\ref{appendix:controller}).

\subsection{General Efficacy}
\label{sec:general_efficacy}
\begin{figure*}[t]
     \centering
     \begin{subfigure}[t]{\textwidth}
     \centering
        \raisebox{-\height}{\includegraphics[width=0.6\textwidth]{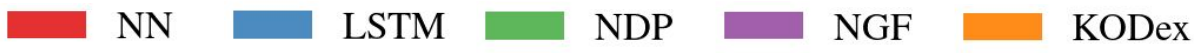}}
    \end{subfigure}
%%%%%%%%%%%%%%%%%%%%%%%%%%%%%%%%%%%%first row
        \begin{subfigure}[t]{0.32\textwidth}
        \raisebox{-\height}{\includegraphics[width=\textwidth]{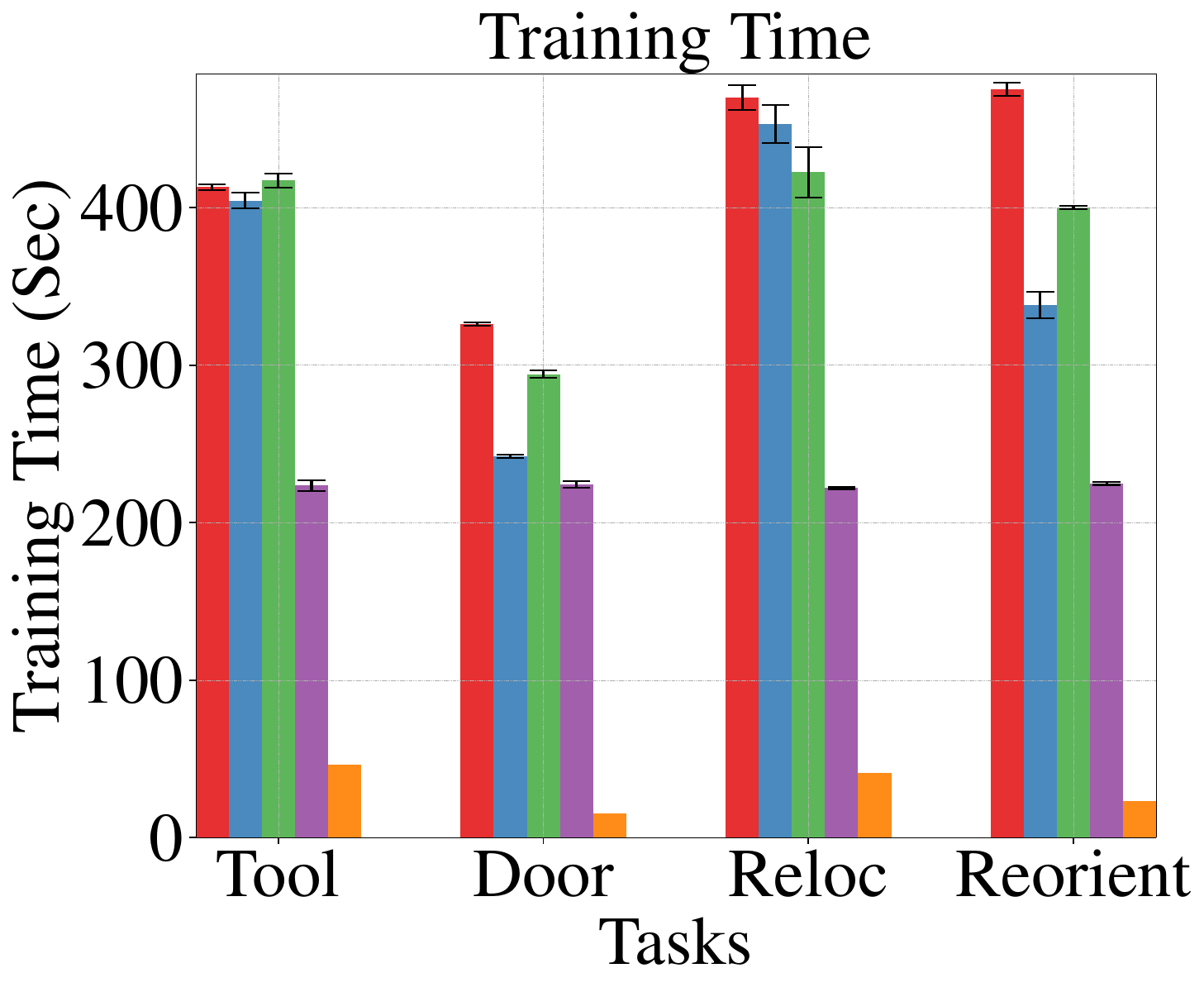}}
    \end{subfigure}
        \begin{subfigure}[t]{0.32\textwidth}
        \raisebox{-\height}{\includegraphics[width=\textwidth]{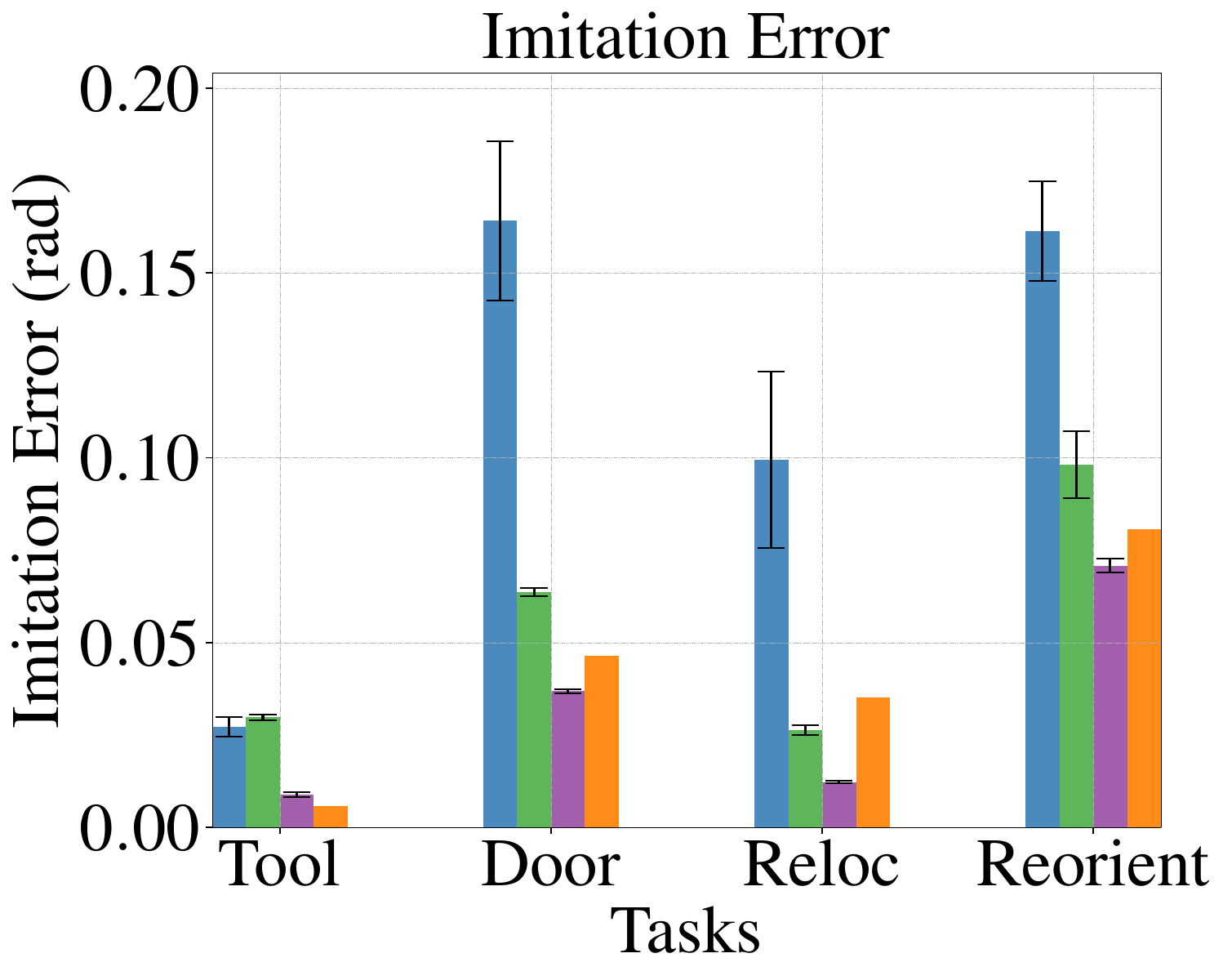}}
    \end{subfigure}
    \begin{subfigure}[t]{0.32\textwidth}
        \raisebox{-\height}{\includegraphics[width=\textwidth]{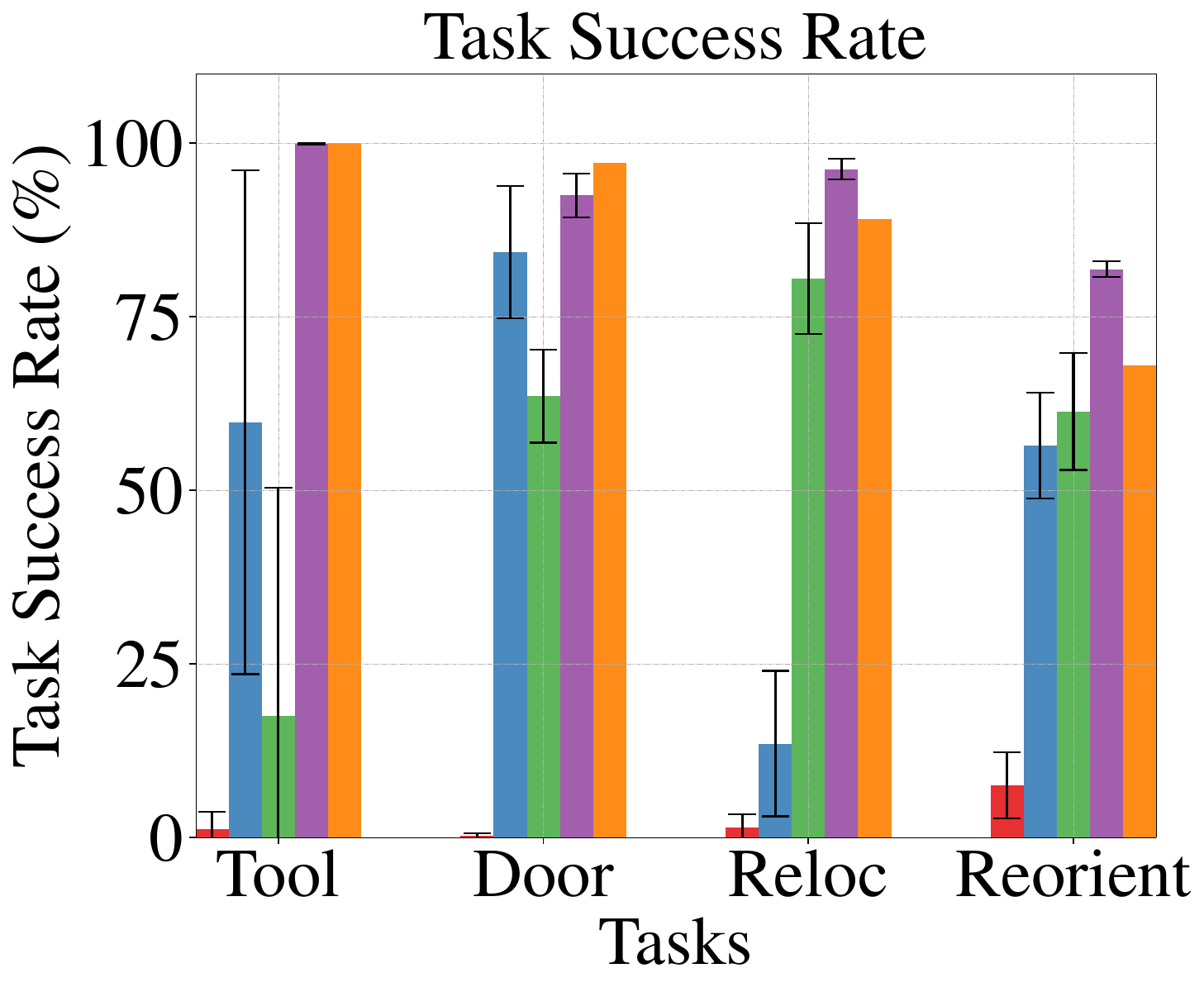}}
    \end{subfigure}
    \caption{
    We report training time (left), imitation error (middle), and success rate (right) for methods on each task when trained on 200 demonstrations and evaluated on an independent set of 10,000 samples. Error bars for baseline methods, show the standard deviation over five random seeds.
    % Policy performances on all tasks based on three metrics: Training Time (left figure), Imitation Error (middle figure), and Task Success Rate (right figure). Each method was trained on 200 demonstrations and evaluated on an independent set of 10,000 samples. For baseline methods, the error bars show the standard deviation over five random seeds.
    }
\label{fig:performance_on_200}
\end{figure*}

% We generated 10,000 new rollouts with the expert RL policy (within the same distribution as the demonstrations) as the test set.
In Fig. \ref{fig:performance_on_200}, we report the training time, imitation error, and task success rate for each method on each task, when trained on 200 demonstrations and tested on 10,000 testing instances.
% (see Appendices ~\ref{Appendix:distribution}). 

\textbf{Training time}: As can be seen, KODex is \textit{an order of magnitude} faster than both unstructured baselines (NN, LSTM) and SOTA IL methods (NDP, NGF). Further, this trend holds across all the tasks. This is to be expected since KODex analytically computes the Koopman operator unlike all the baselines, which rely on gradient descent and numerical optimization.

\textbf{Imitation error}: In general, all methods (except NN) achieve low imitation error for the Tool Use task with negligible difference across methods. In the three remaining tasks, we see that all structured methods (NDP, NGF, KODex) considerably outperform the unstructured baseline (LSTM).  We have excluded the significantly larger imitation errors generated by NN from the plot to preserve the resolution necessary to distinguish between the other methods. These results reinforce the effectiveness of structured methods in imitating demonstrations. Importantly, KODex is able to achieve imitation performance comparable to the SOTA IL methods despite its simplicity, while remaining an order of magnitude more computationally efficient.

\begin{figure*}[t]
     \centering
     \begin{subfigure}[t]{0.96\textwidth}
     \centering
        \raisebox{-\height}{\includegraphics[width=0.7\textwidth]{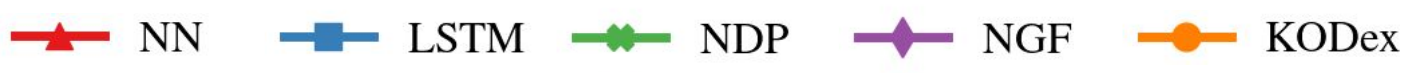}}
    \end{subfigure}
%%%%%%%%%%%%%%%%%%%%%%%%%%%%%%%%%%%%first row
        \begin{subfigure}[t]{0.24\textwidth}
        \raisebox{-\height}{\includegraphics[width=\textwidth]{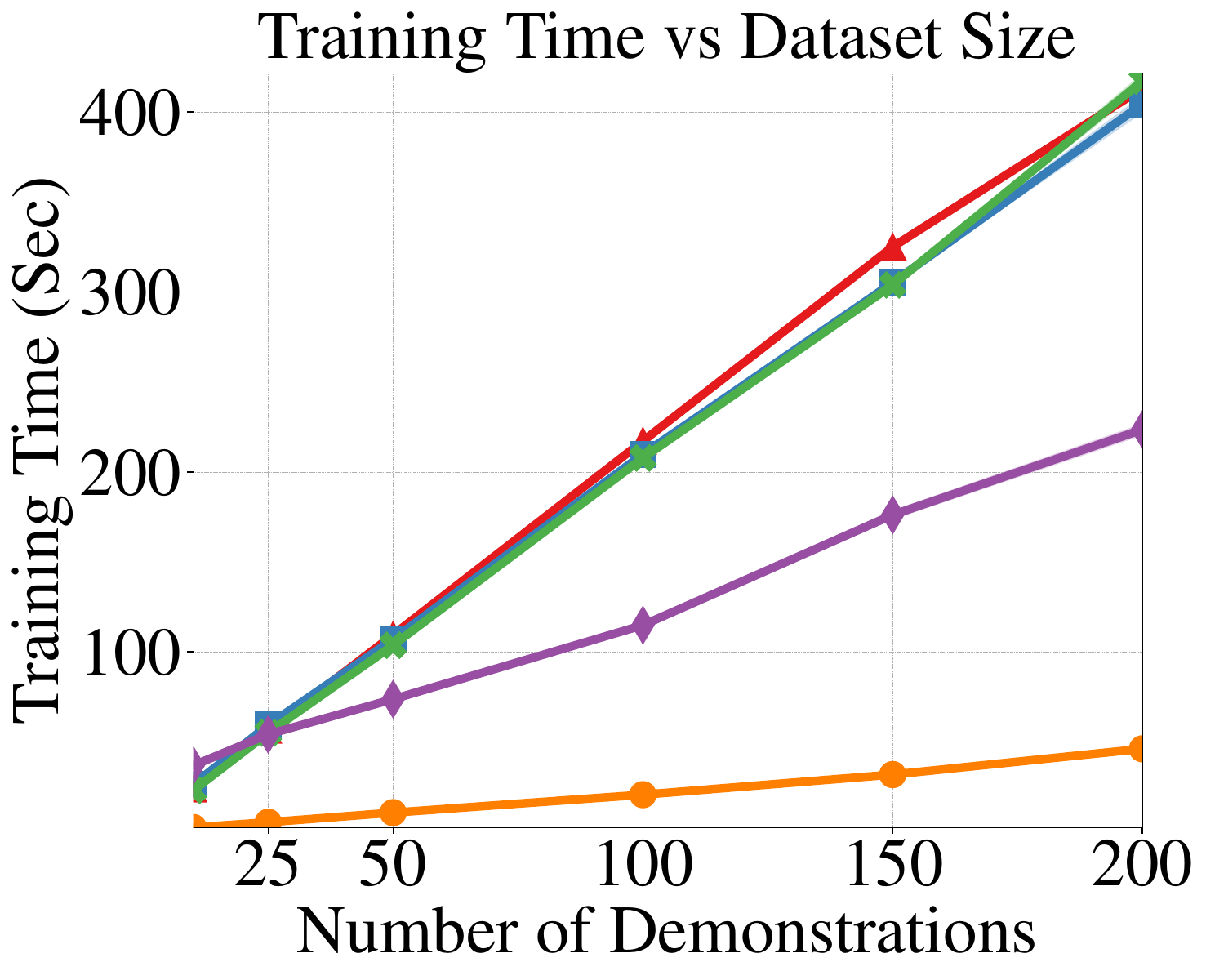}}
    \end{subfigure}
        \begin{subfigure}[t]{0.24\textwidth}
        \raisebox{-\height}{\includegraphics[width=\textwidth]{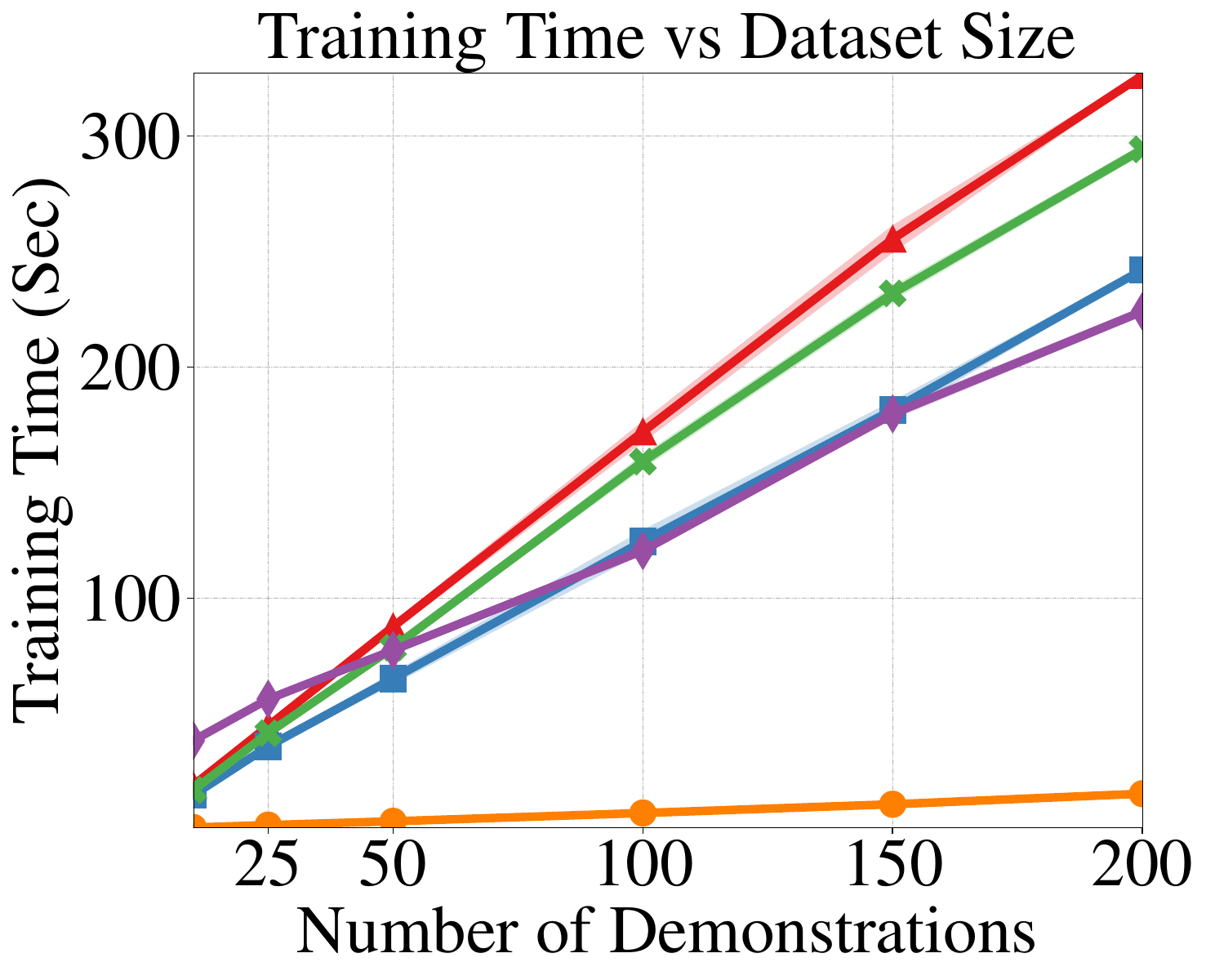}}
    \end{subfigure}
    \begin{subfigure}[t]{0.24\textwidth}
        \raisebox{-\height}{\includegraphics[width=\textwidth]{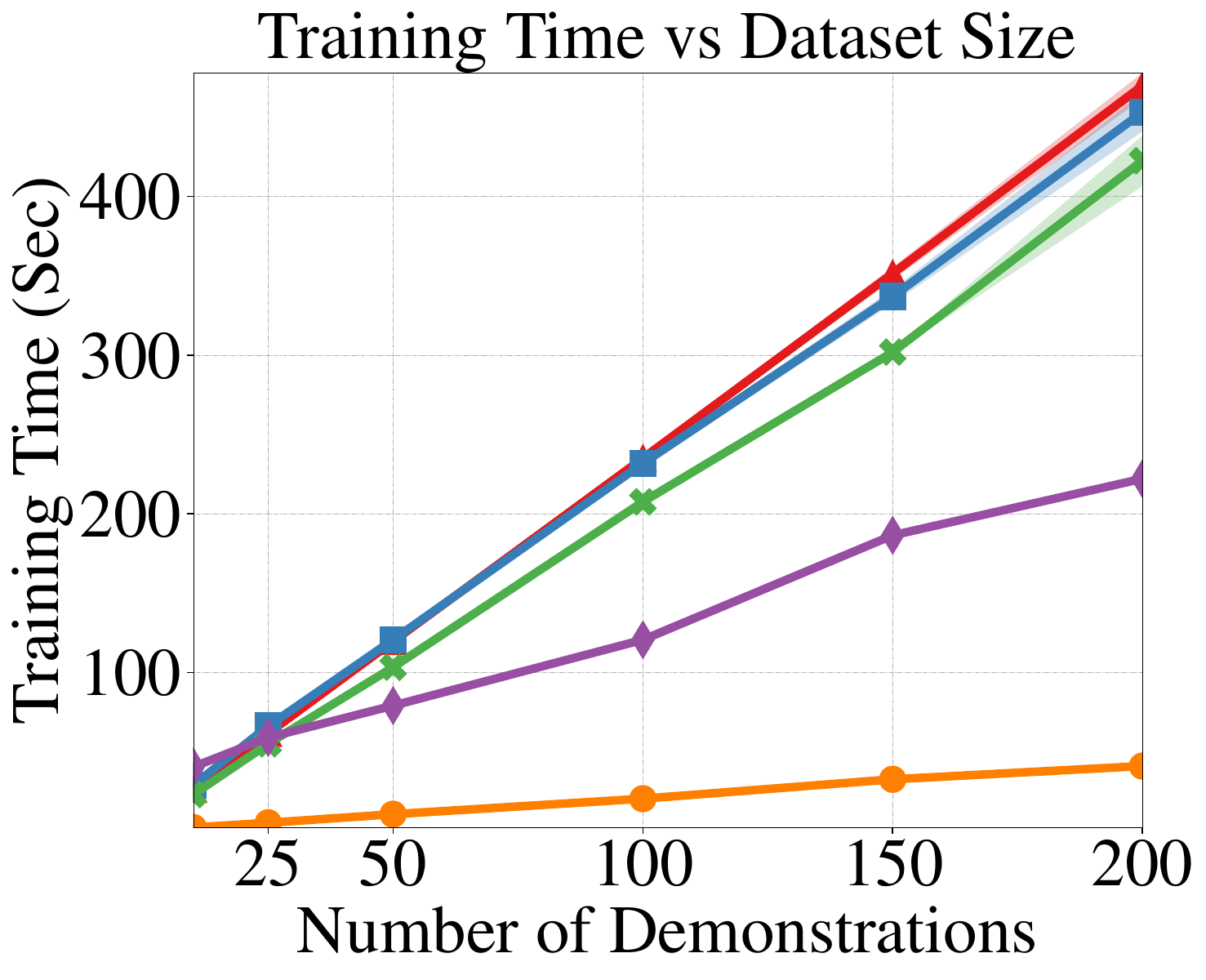}}
    \end{subfigure}
    \begin{subfigure}[t]{0.24\textwidth}
        \raisebox{-\height}{\includegraphics[width=\textwidth]{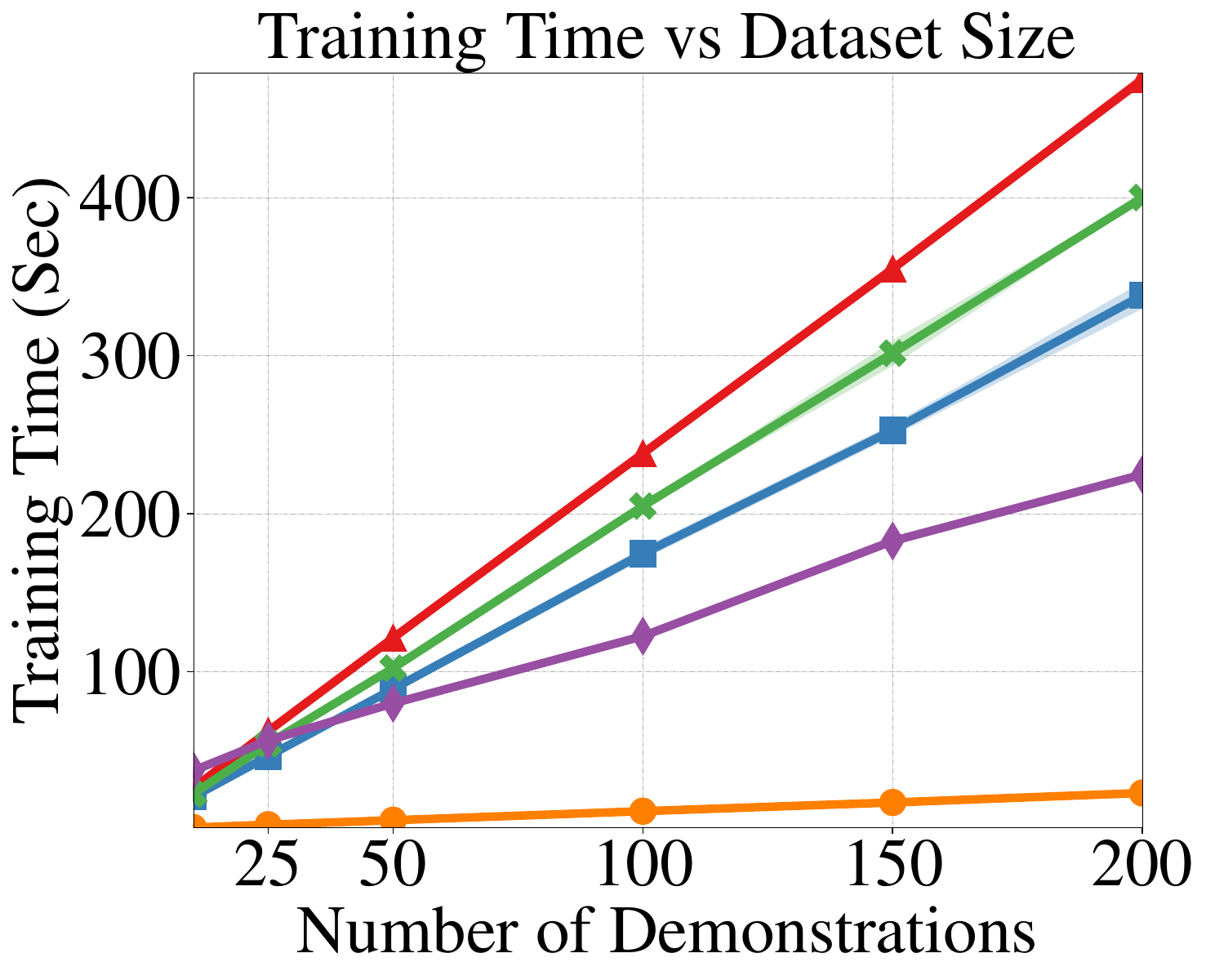}}
    \end{subfigure}
    %%%%%%%%%%%%%%%%%%%%%%%%%%%%%%%%%%%%second row
        \begin{subfigure}[t]{0.24\textwidth}
        \raisebox{-\height}{\includegraphics[width=\textwidth]{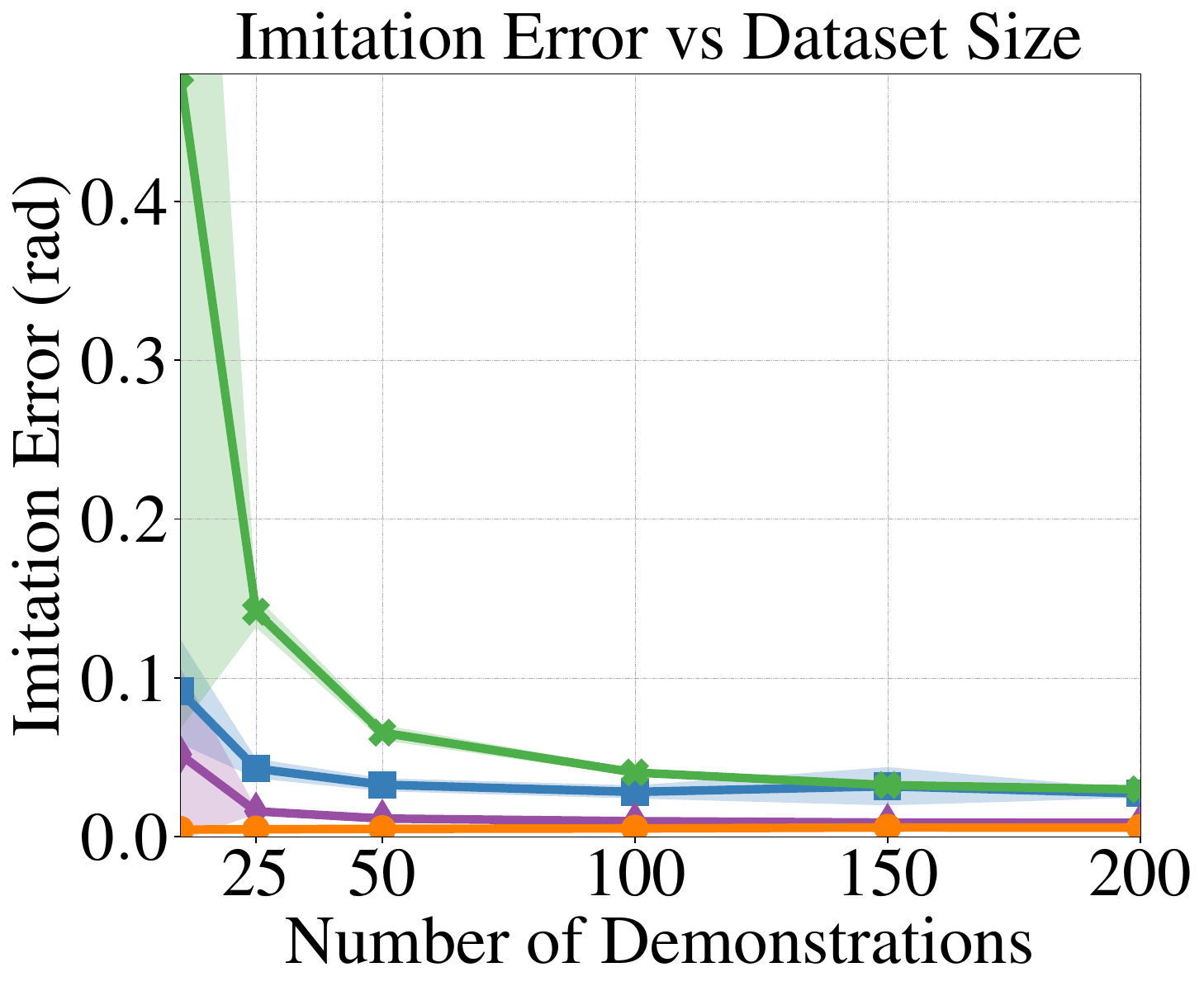}}
    \end{subfigure}
        \begin{subfigure}[t]{0.24\textwidth}
        \raisebox{-\height}{\includegraphics[width=\textwidth]{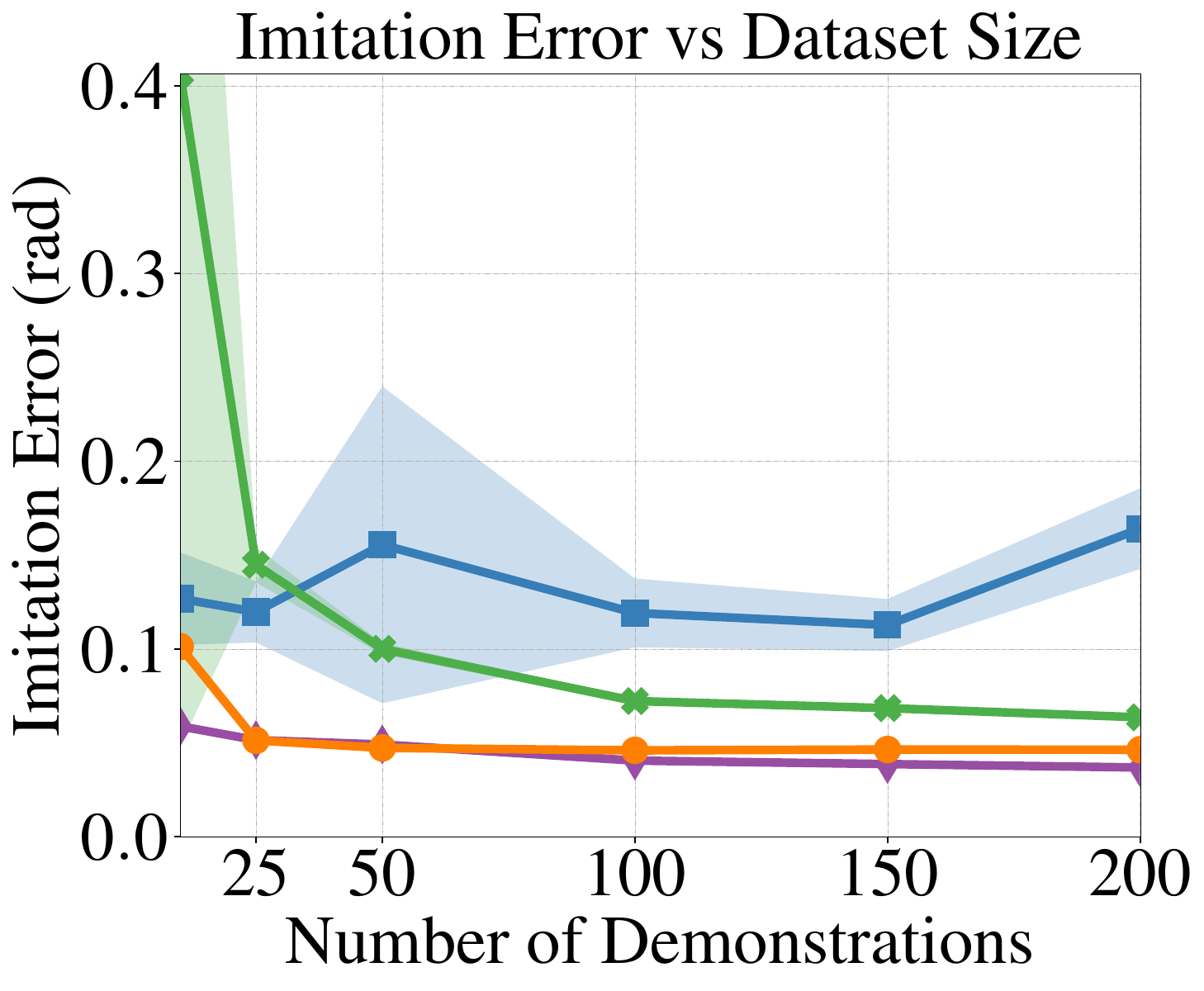}}
    \end{subfigure}
    \begin{subfigure}[t]{0.24\textwidth}
        \raisebox{-\height}{\includegraphics[width=\textwidth]{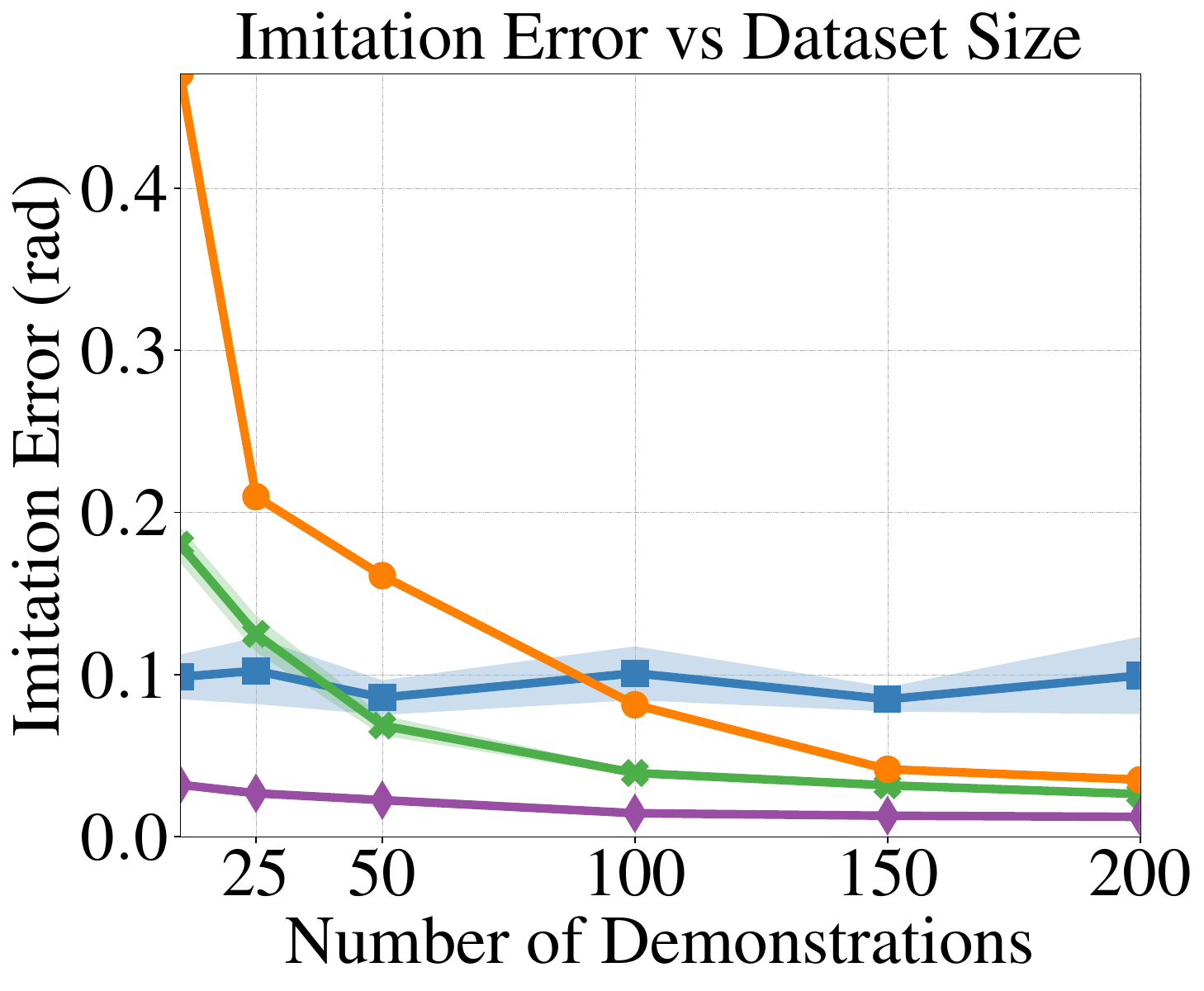}}
    \end{subfigure}
    \begin{subfigure}[t]{0.24\textwidth}
        \raisebox{-\height}{\includegraphics[width=\textwidth]{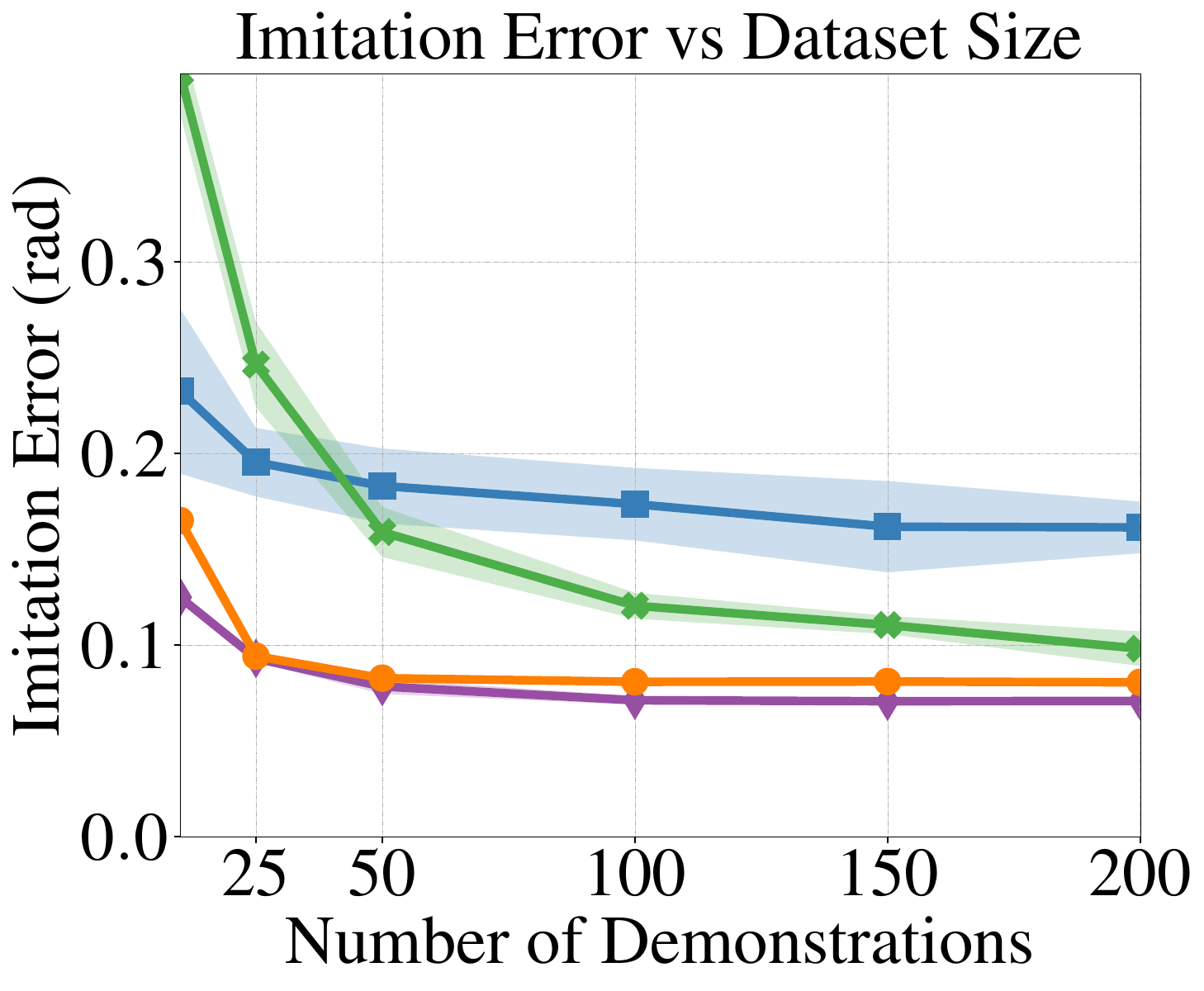}}
    \end{subfigure}
        %%%%%%%%%%%%%%%%%%%%%%%%%%%%%%%%%%%%third row
        \begin{subfigure}[t]{0.24\textwidth}
        \raisebox{-\height}{\includegraphics[width=\textwidth]{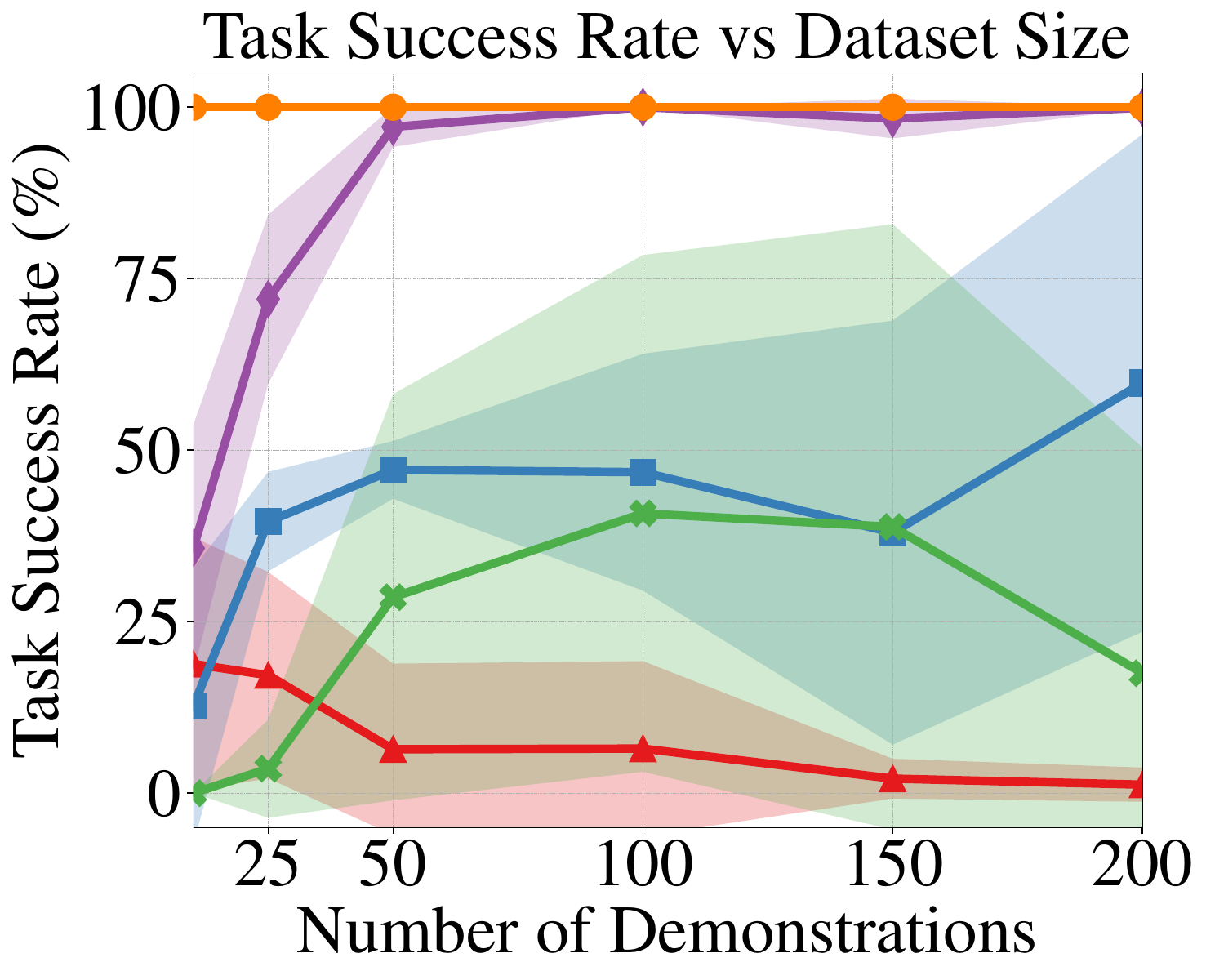}}
        \caption{Tool Use (Hammer)} 
    \end{subfigure}
        \begin{subfigure}[t]{0.24\textwidth}
        \raisebox{-\height}{\includegraphics[width=\textwidth]{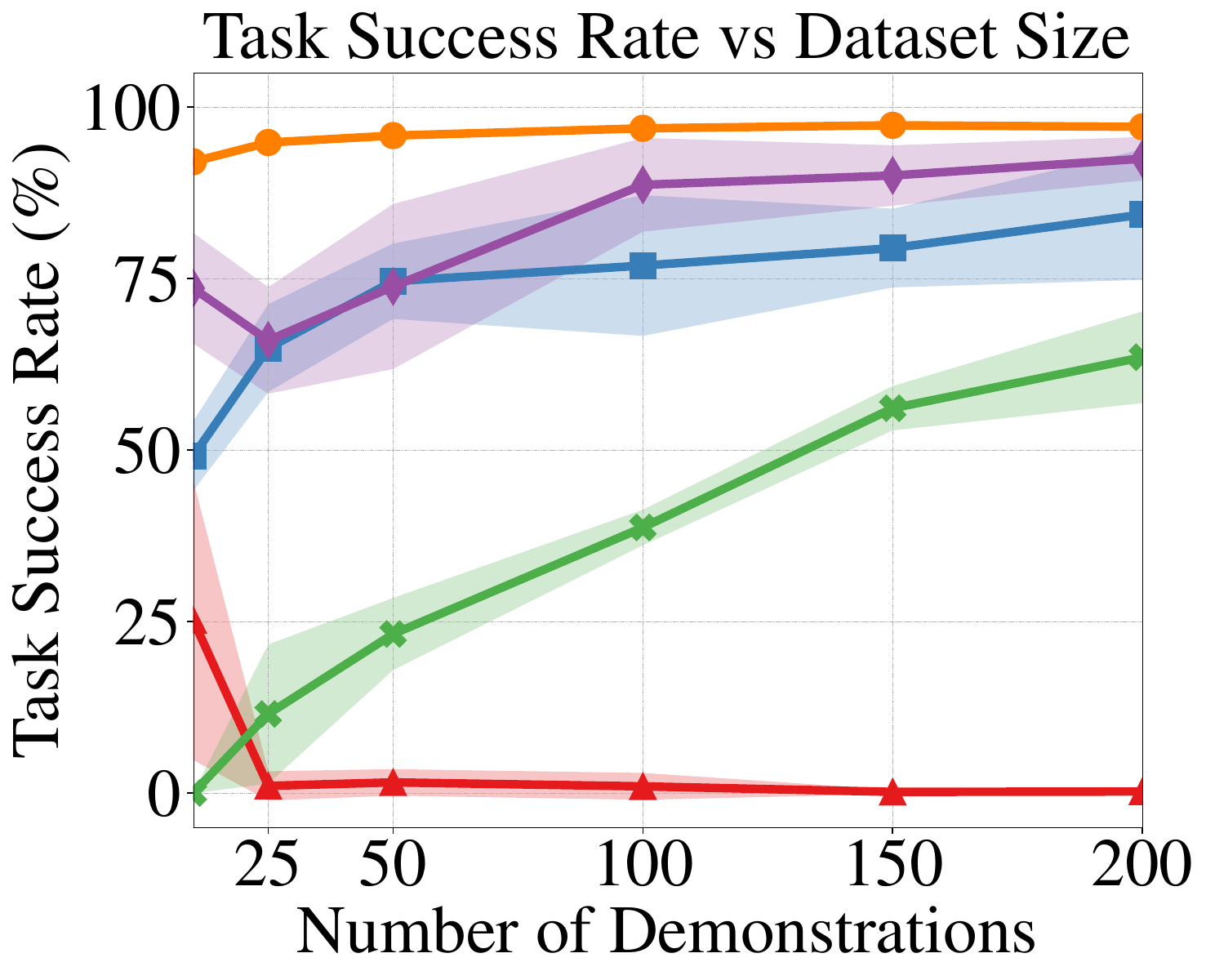}}
        \caption{Door Opening} 
    \end{subfigure}
    \begin{subfigure}[t]{0.24\textwidth}
        \raisebox{-\height}{\includegraphics[width=\textwidth]{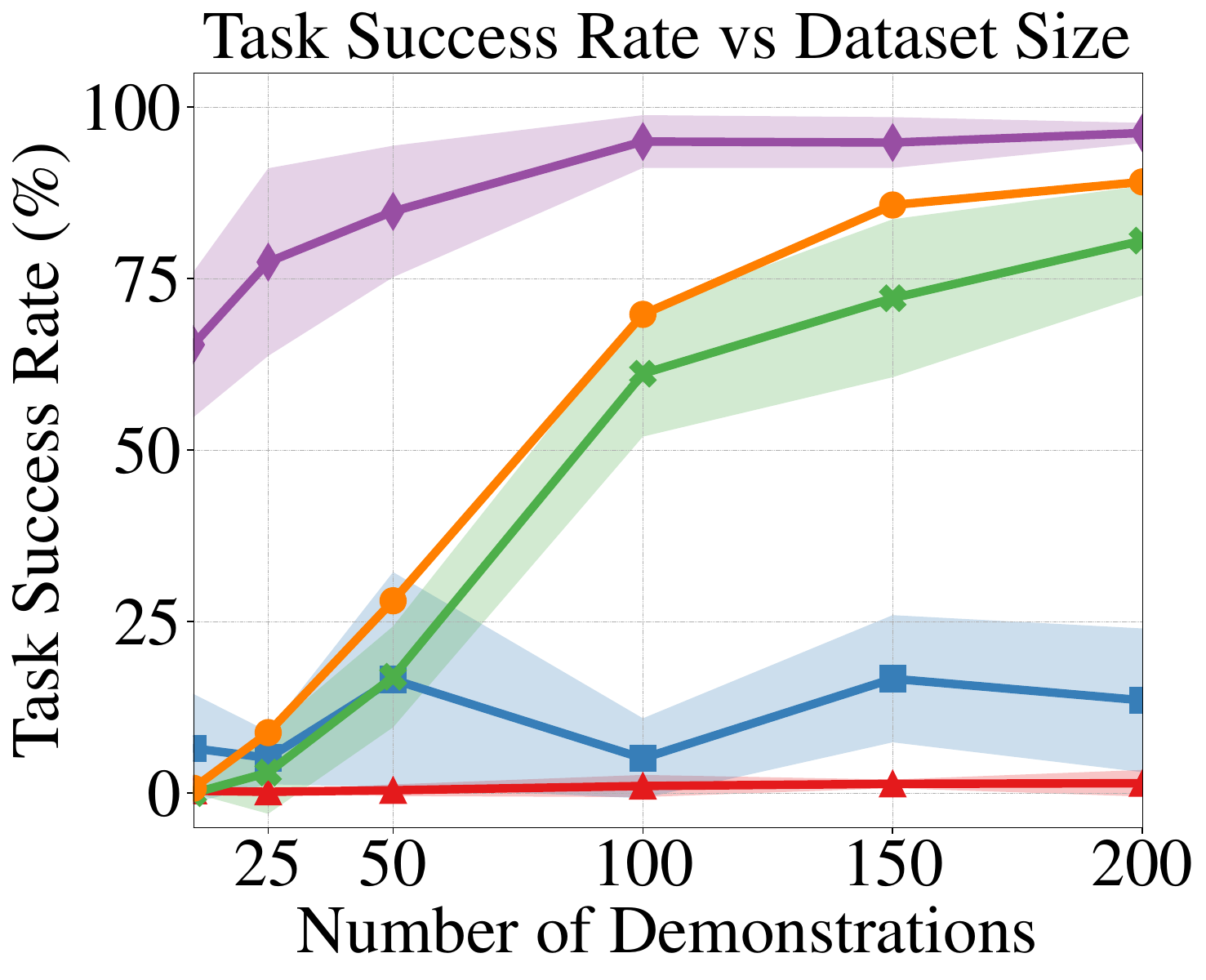}}
        \caption{Object Relocation} 
    \end{subfigure}
    \begin{subfigure}[t]{0.24\textwidth}
        \raisebox{-\height}{\includegraphics[width=\textwidth]{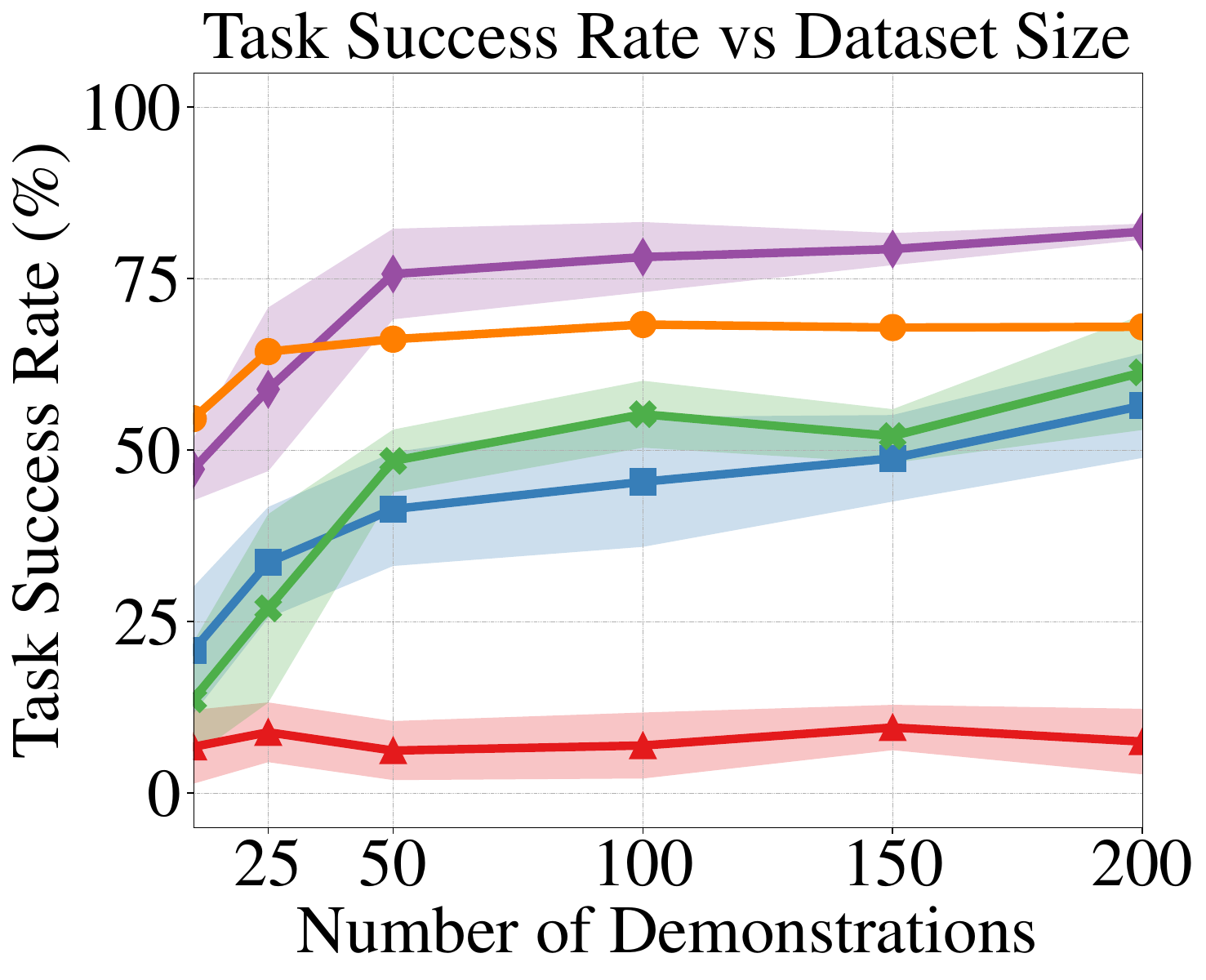}}
        \caption{In-hand Reorientation} 
    \end{subfigure}
    \caption{The effects of number of demonstrations on training time (top row), imitation error (middle row), and success rate (bottom row) for all methods on each task. Solid lines indicate mean trends and shaded areas show $\pm$ standard deviation over five random seeds. 
    % Note that we did not plot NN's imitation errors as they were out of scale.
    }
\label{fig:imitation_learning}
\end{figure*}
\vspace{3pt}

\textbf{Task success rate}: 
% Following the trends we have observed in other metrics, KODex considerably outperforms or matches all baselines in terms of task success rate. Specifically, 
As expected, the NN policy performs significantly worse than all other methods.  On the other hand, LSTM achieves impressive task success rates, even outperforming NDP in two of the four tasks. This is in stark contrast to its high imitation error. While counter-intuitive, this observation follows the recent finding that imitation error and task success rate might not necessarily be correlated ~\cite{mandlekar2021matters}.
We observe that KODex and NGF perform comparably, with one achieving a higher task success rate than the other in two of the four tasks. 
% KODex is also able to consistently outperform NN, LSTM, and NDP policies and match NGF policy at 200 demonstrations, as shown in Fig.~\ref{fig:performance_on_200}.
Importantly, KODex results in the most consistent and predictable performance due to its lack of sensitivity to initialization. 
% As we would expect, the naive NN baseline results in the lowest performance, while LSTM is able to occasionally match the performance of other structured methods. We also see that imitation error and task success rate do not necessarily correlate, matching the findings from a recent review paper~\cite{mandlekar2021matters}.

% These results suggest that \textbf{(i).} KODex achieves the highest training efficiency, which is not surprising as KODex analytically computes the optimal solution instead of running gradient descent for hundreds of iterations. \textbf{(ii).} KODex consistently offers the best performance against all baselines over all tasks, indicating the benefits of introducing Koopman operator for dexterous manipulation tasks. We believe that the observables' strong power of modelling the nonlinearity of each dynamical system is the main key of the magic performances. While other baseline models are less effective at capturing the system nonlinearity, e.g. using activation functions (NN, LSTM) or predefined dynamics structures (NDP), resulting in the worse and inconsistent performance across all tasks. \textbf{(iii).} KODex is the most sample-efficient as it results in the lowest mean imitation error and largest task success rate (and both with smallest variance) across almost all sizes of demonstrations. Take Door opening and Tool use as examples, with only 50 demonstrations, KODex can almost achieve 100\% success rate. 

% \vspace{3pt}
% \noindent \textbf{Sample Efficiency}: 
\vspace{-0.1cm}
\subsection{Scalability and Sample Efficiency}
\label{sec:sample_efficiency}
To investigate scalability and sample efficiency, we trained policies on a varying number of demonstrations ([10, 25, 50, 100, 150, 200]). In Fig.~\ref{fig:imitation_learning}, we report the training time, imitation error, and task success rate for each method as a function of the number of demonstrations when tested on the same 10,000 instances used to evaluate general efficacy. 

\textbf{Training time}: We observe that KODex scales with the number of demonstrations significantly better than the baselines, as evidenced by its training time growing at considerably lower rates. 
% As one would expect, NGF scales better than the other baselines.

\textbf{Imitation error and Success rate}: We find that unstructured models (NN and LSTM) fail to demonstrate a consistent monotonic decrease (increase) in imitation error (task success rate) as the number of demonstrations increase. In stark contrast, structured methods (NDP, NGF, and KODex) are able to consistently drive down imitation error and improve task success rate. KODex almost consistently achieves the lowest imitation error and the highest task success rate with the fewest number of demonstrations, and is closely followed by NGF. These observations suggest that KODex tends to be comparably, if not more sample efficient than the baselines, thanks to the rich structure induced by the Koopman operator and the resulting effectiveness in capturing nonlinear dynamics. The only exception to this trend is the Object Relocation task, in which KODex requires 150 demonstrations to perform comparably to NGF. We speculate this is because the demonstrations for this task exhibit high variance as the hand base moves across a large space, and KODex requires more demonstrations to capture the \textcolor{black}{reference} dynamics.
\subsection{Additional Experiments}
\label{sec:additional_exp}
Additional experiments reported in the appendix suggest that KODex learns policies that i) \textcolor{black}{have inference time on par with SOTA baselines (Appendix~\ref{sec:inference})}, ii) have zero-shot out-of-distribution generalization comparable to SOTA IL methods (Appendix~\ref{sec:ood}), iii) are robust to changes in physical properties (Appendix~\ref{sec:robustness}), iv) are not overly sensitive to the choice of basis functions (Appendix~\ref{sec:basis_choice}), v) are nearly-stable linear dynamical systems that generate safe and smooth robot trajectories (Appendix~\ref{sec:Eigenvalue_Analysis}), \textcolor{black}{and vi) are significantly more efficient and scalable than a baseline BC method that directly learns state-action mappings (Appendix~\ref{sec:state_action_BC})}.
% In Appendix, we i) report the out-of-distribution generalization in Sec.~~\ref{sec:ood}, ii) demonstrate the robustness to changes in physical properties in Sec.~\ref{sec:robustness}, iii) evaluate KODex's performance on different choices of basis functions used in $\phi(\mathrm{x}(t))$ in Sec.~\ref{sec:basis_functions}, and iv) analyze the computed Koopman matrix $\mathbf{K}$ in Sec.~\ref{sec:Eigenvalue_Analysis}.
% \vspace{-0.3cm}
\section{Conclusions} 
\label{sec:conclusion}
We investigated the utility of Koopman operator theory in learning dexterous manipulation skills by encoding complex nonlinear \textcolor{black}{reference} dynamics as linear dynamical systems in higher-dimensional spaces. Our investigations conclusively show that a Koopman-based framework can i) analytically learn dexterous manipulation skills, eliminating the sensitivity to initialization and reducing the need for user expertise, and ii) match or outperform SOTA imitation learning approaches on various dexterous manipulation tasks, while being an order of magnitude faster.

\section{Limitations and Future Work}
\label{sec:limitation}
% \section{Limitation and Future Work} 
While our work offers promise for the utility of Koopman operators in dexterous manipulation, it reveals numerous avenues for further improvement.
First, we did not deploy KODex on physical robots. Although our results on robustness to changes in physical properties show promise, we plan to deploy KODex on physical platforms to translate our findings to hardware.
% In the future, we plan to conduct hardware experiments on a physical dexterous robot hand, PYSONIC Ability Hand \cite{akhtar2020touch}.
\textcolor{black}{
Second, in this work, we only considered polynomial basis functions, other non-smooth functions, e.g., ReLU \cite{agarap2018deep}, could also be beneficial to manipulation tasks involving friction and contact.}
Third, KODex has limited out-of-distribution generalization, much like most existing imitation learning approaches. 
% This necessitates the need for additional training data collected offline or online using approaches like DAGGER~\cite{ross2011reduction}. 
Future work can investigate if additional data collection~\cite{ross2011reduction} and learned lifting functions~\cite{weissenbacher2022koopman,lusch2018deep,Li2020Learning} alleviate this concern.
% if such generalization can be improved by additional training data collected offline or online using approaches like DAGGER~\cite{ross2011reduction}, and by swapping the polynomial lifting functions with learned lifting functions (e.g., \cite{weissenbacher2022koopman,lusch2018deep,Li2020Learning}), allowing for end-to-end policy learning.
% the In addition, a more robust controller may be required if deploying KODex in real world.
% Meanwhile, we would like to explore the usage of Koopman operator as the policy structure in neural networks   
% Fourth, policies learned using KODex can be subject to rigorous theoretical analysis, thanks to its linear dynamical system. For instance, a recently-discovered connection between contraction analysis and Koopman operator theory~\cite{Bowen2021Contraction} can help derive convergence and stability guarantees on learned policies. 
Fourth, KODex relies on demonstrated action trajectories to learn the tracking controller and reduce human effort. It might be possible to instead use reinforcement learning \cite{peng2020learning}, thereby enabling the ability to learn from state-only observations \cite{torabi2019recent}. \textcolor{black}{Fifth, KODex could be evaluated on other domains and robotics tasks (e.g., the benchmark tasks in \cite{majumdar2023we}) to further understand the trade-offs between KODex and other imitation learning baselines.}
% can learn skills from state-only observations \cite{torabi2019recent}, if the tracking controller is trained separately from demonstration, i.e., using reinforcement learning \cite{peng2020learning}.
Finally, Koopman operators can be used to learn system dynamics via self play to enable model-based reinforcement learning.

\newpage
\bibliography{references}

\newpage

\appendix
% \section*{Acknowledgments}
\begin{center}
    {\LARGE{Appendices}}
\end{center}

\section{KODex Pseudo-code}
The overall pseudo-code for KODex is shown below.
\label{sec:pseudo-code}
\begin{algorithm}[h] 
\SetAlgoLined
\textbf{Demonstration Data Collection} \\
Initialize $D = \varnothing$; \\
\For{$n\in \{1, ..., N\}$}{
Generate a $T^{(n)}$-horizon trajectory of states and torques $\{[\mathrm{x}^{n}(t), \mathrm{\tau}^{n}(t)]\}^{t=T^{(n)}}_{t=1}$;\\
Add $\{[\mathrm{x}^{n}(t), \mathrm{\tau}^{n}(t)]\}^{t=T^{(n)}}_{t=1}$ to $D$;\\
}
\textbf{Koopman Operator Approximation} \\
Determine lifting function $\phi(\mathrm{x}(t))$; \\
% Initialize $\bar{D} = \varnothing$; \\
% Initialize $E = [\,]$; \\
% \For{$n\in \{1, ..., N\}$}{
% Select $\{\hat{\mathrm{x}}_t^{n}\}^{t=T}_{t=1}$ from $D$ and add it to $\bar{D}$; \\
% Compute $\mathbf{K}_{\bar{D}}$ on $\bar{D}$ with $\hat{g}(\hat{\mathrm{x}}_t)$ via Eqn.~\ref{eqn:k} \ref{eqn:k_comp}; \\
% Compute the $L_2$ sum of residual error $e_n$ on $D$ using $\mathbf{K}_{\bar{D}}$; \\
% Add $e_n$ to $E$; \\
% }
Compute $\mathbf{K}$ on $D$ (\ref{eqn:k}, \ref{eqn:k_comp_new}); \\
\textbf{Controller Design} \\
Build a controller $C$ as a neural network with inputs as $(\mathrm{x}_r(t), \mathrm{x}_r(t+1))$ and output as $\mathrm{\tau}(t)$;\\
Train $C$ using state-torque pairs $(\mathrm{x}_r^n(t), \mathrm{x}_r^n(t+1),\mathrm{\tau}^n(t))$ in $D$ (\ref{eqn: controller});\\
\textbf{Execution} \\
Specify the initial states $\mathrm{x}(1)$; \\
\For{$t\in \{1, ..., T-1\}$}{
Predict the next robot states $\hat{\mathrm{x}}_r(t+1)$ using $\mathbf{K}$ (\ref{eqn:observable_system} \ref{eqn:gx}); \\ 
Read the current robot states $\mathrm{x}_r(t)$;\\
Generate the torque $\mathrm{\tau}(t)$ using $C$ on $(\mathrm{x}_r(t), \hat{\mathrm{x}}_r(t+1))$ and execute it;
}
\caption{KODex}
\label{alg: overview}
\end{algorithm} 
\section{State Design}
\label{Appendix:state_design}
% The settings are the same as DAPG project. Do we have to mention this in paper? Or reviews or readers may be confused why the DOF of hand for each task varies.
% I think for our own hand, we can set whatever it is as we want, but for the ADROIT Hand, we need to follow exactly the settings before. We can claim this in context.
In this section, we show the state design for each task in detail. It should be noted that the motion capability of the hand for each task were suggested from the work \cite{Rajeswaran2018DAPG} that originally introduced these tasks. For a decent implementation, we employed the same setting.
\newline
\textbf{Tool use}
For this task, the floating wrist base can only rotate along the $x$ and $y$ axis, so we have $\mathrm{x}_r(t) \in \mathcal{X}_r \subset \mathbb{R}^{26}$. Regarding the object states, unlike the other tasks, where the objects of interest are directly manipulated by the hand, this task requires to modify the environment itself. As a result, except for the hammer positions, orientations and their corresponding velocities $\mathrm{p}^{\text{tool}}_t, \mathrm{o}^{\text{tool}}_t, \dot{\mathrm{p}}^{\text{tool}}_t, \dot{\mathrm{o}}^{\text{tool}}_t$ ($\mathbb{R}^{3}$), we also define the nail goal position $\mathrm{p}^{\text{nail}}$
 ($\mathbb{R}^{3}$).
 % which functions similarly as $\mathrm{p}^{\text{door}}$ in the task of door opening.
 Finally, we have $\mathrm{x}_o(t) = [\mathrm{p}^{\text{tool}}_t, \mathrm{o}^{\text{tool}}_t, \dot{\mathrm{p}}^{\text{tool}}_t, \dot{\mathrm{o}}^{\text{tool}}_t, \mathrm{p}^{\text{nail}}] \in \mathcal{X}_o \subset \mathbb{R}^{15}$. As a result, $\mathrm{x}(t)$ includes 41 states in total and we use $T = 100$.
 \newline
\textbf{Door opening} For this task, the floating wrist base can only move along the direction that is perpendicular to the door plane but rotate freely, so we have $\mathrm{x}_r(t) \in \mathcal{X}_r \subset \mathbb{R}^{28}$. Regarding the object states, we define the fixed door position $\mathrm{p}^{\text{door}}$, which can provide with case-specific information (similar to $\mathrm{p}^{\text{nail}}$ in Tool Use), and the handle positions $\mathrm{p}^{\text{handle}}_t$ (both $\mathbb{R}^{3}$). In order to take into consideration the status of door being opened, we include the angular velocity of the opening angle $v_t$$
 (\mathbb{R}^{1})$. Finally, we have $\mathrm{x}_o(t) = [\mathrm{p}^{\text{handle}}_t, v_t, \mathrm{p}^{\text{door}}] \in \mathcal{X}_o \subset \mathbb{R}^{7}$. As a result, $\mathrm{x}(t)$ includes 35 states in total and we use $T = 70$.
 \newline
\textbf{Object relocation} For this task, the ADROIT hand is fully actuated, so we have $\mathrm{x}_r(t) \in \mathcal{X}^{r} \subset \mathbb{R}^{30}$ (24-DoF hand + 6-DoF floating wrist base). Regarding the object states, we define $\mathrm{p}^{\text{target}}$ and $\mathrm{p}^{\text{ball}}_t$ as the target and current positions. Then, we compute $\bar{\mathrm{p}}^{\text{ball}}_t = \mathrm{p}^{\text{ball}}_t - \mathrm{p}^{\text{target}}$, which is the component of $\mathrm{p}^{\text{ball}}_t$ in a new coordinate frame that is constructed by $\mathrm{p}^{\text{target}}$ being the origin. 
 % Similar to the in-hand reorientation task, 
 We additional include  the ball orientation $\mathrm{o}^{\text{ball}}_t$ and their corresponding velocities $\dot{\mathrm{p}}^{\text{ball}}_t$,  $\dot{\mathrm{o}}^{\text{ball}}_t$ (all $\mathbb{R}^{3}$).
Finally, we have $\mathrm{x}_o(t) = [\bar{\mathrm{p}}^{\text{ball}}_t, \mathrm{o}^{\text{ball}}_t, \dot{\mathrm{p}}^{\text{ball}}_t, \dot{\mathrm{o}}^{\text{ball}}_t] \in \mathcal{X}_o \subset \mathbb{R}^{12}$. As a result, $\mathrm{x}(t)$ includes 42 states in total and we use $T = 100$.
\newline
\textbf{In-hand reorientation} For this task, the floating wrist base is fixed, so we only consider the 24-DoF hand joints. Therefore, we have $\mathrm{x}_r(t) \in \mathcal{X}_{r} \subset \mathbb{R}^{24}$. Regarding the object states, we define $\mathrm{o}^{\text{goal}}$ and $\mathrm{o}^{\text{pen}}_t$ as the goal and current pen orientations, which are both unit direction vectors. Then, we transform $\mathrm{o}^{\text{pen}}_t$ to a new rotated coordinate frame that is constructed by $\mathrm{o}^{\text{goal}}$ being $x$ axis ([1,0,0]). Note that the vector $\bar{\mathrm{o}}^{\text{pen}}_t$ after transformation is also a unit vector and it converges to x axis if the pen is perfectly manipulated to goal orientation $\mathrm{o}^{\text{goal}}$. In addition, we also include the center of mass position $\mathrm{p}^{\text{pen}}_t$ and their corresponding velocities $\dot{\mathrm{p}}^{\text{pen}}_t$,  $\dot{\mathrm{o}}^{\text{pen}}_t$ (all  $\mathbb{R}^{3}$). Finally, we have $\mathrm{x}_o(t) = [\mathrm{p}^{\text{pen}}_t, \bar{\mathrm{o}}^{\text{pen}}_t, \dot{\mathrm{p}}^{\text{pen}}_t, \dot{\mathrm{o}}^{\text{pen}}_t] \in \mathcal{X}_o \subset \mathbb{R}^{12}$. As a result, $\mathrm{x}(t)$ includes 36 states in total and we use $T = 100$.

 In this work, we only included the joint positions as the robot states (with the only exception of NGF's second-order policy) for the following reasons: 1) Given that these tasks are not repetitive, we found that joint position information was sufficient to disambiguate the robot's next action, 2) even when ambiguity arises for a given joint position, object state information can help with disambiguation. Further, the impressive performance achieved by KODex in our experiments support this design choice. Indeed, KODex is agnostic to this specific state design. One can incorporate velocity information into the robot state space without the need of any changes to the training procedure.

\section{Task Success Criteria}
\label{appendix:Task_Success_criteria}
The task success criteria are listed below. The settings were the same as proposed in \cite{Rajeswaran2018DAPG}.
\newline
\textbf{Tool Use:} The task is considered successful if at last time step $T$, the Euclidean distance between the final nail position and the goal nail position is smaller than 0.01.
\newline
\textbf{Door Opening:} The task is considered successful if at last time step $T$, the door opening angle is larger than 1.35 rad.
\newline
\textbf{Object Relocation:} At each time step $t$, if $\sqrt{|\mathrm{p}^{\text{target}} - \mathrm{p}^{\text{ball}}_t|^2} < 0.10$, then we have $\rho(t) = 1$. The task is considered successful if $\sum_{t=1}^T \rho(t) > 10$.
\newline
\textbf{In-hand Reorientation:} At each time step $t$, if $\mathrm{o}^{\text{goal}} \cdot \mathrm{o}^{\text{pen}}_t > 0.90$ ($\mathrm{o}^{\text{goal}} \cdot \mathrm{o}^{\text{pen}}_t$ measures orientation similarity), then we have $\rho(t) = 1$. The task is considered successful if $\sum_{t=1}^T \rho(t) > 10$.

\section{Sampling Procedure}
\label{Appendix:distribution}
We describe the sampling procedure in this section. The sample distributions used for RL training and demo collection were identical, as suggested in \cite{Rajeswaran2018DAPG}. The out-of-distribution data were generated to evaluate the zero-shot out-of-distribution generalizability of each policy.
\newline
\textbf{Tool Use:} We randomly sampled the nail heights ($h$) from a uniform distributions. Within distribution: we used $h \in \mathcal{H} \sim \mathcal{U}(0.1,0.25)$; Out of distribution: we used $h \in \mathcal{H} \sim \mathcal{U}(0.05,0.1) \cup \mathcal{U}(0.25,0.3)$.
\newline
\textbf{Door Opening:} We randomly sampled the door positions ($xyz$) from uniform distributions. Within distribution: we used $x \in \mathcal{X} \sim \mathcal{U}(-0.3,0)$, $y \in \mathcal{Y} \sim \mathcal{U}(0.2,0.35)$, and $z \in \mathcal{Z} \sim \mathcal{U}(0.252,0.402)$; Out of distribution: we used $y \in \mathcal{Y} \sim \mathcal{U}(0.15,0.2) \cup \mathcal{U}(0.35,0.4)$ ($x, z$ remained unchanged).
\newline
\textbf{Object Relocation:} We randomly sampled the target positions ($xyz$) from uniform distributions. Within distribution: we used $x \in \mathcal{X} \sim \mathcal{U}(-0.25,0.25)$, $y \in \mathcal{Y} \sim \mathcal{U}(-0.25,0.25)$, and $z \in \mathcal{Z} \sim \mathcal{U}(0.15,0.35)$; Out of distribution: we used $z \in \mathcal{Z} \sim \mathcal{U}(0.35,0.40)$ ($x, y$ remained unchanged).
\newline
\textbf{In-hand Reorientation:} We randomly sampled the pitch ($\alpha$) and yaw ($\beta$) angles of the goal orientation from uniform distributions. Within distribution: we used $\alpha \in \mathcal{A} \sim \mathcal{U}(-1,1)$ and $\beta \in \mathcal{B} \sim \mathcal{U}(-1,1)$; Out of distribution: we used 
% $\alpha \in \mathcal{A} \sim \mathcal{U}(-1.2,-1) \cup \mathcal{U}(1,1.2)$ and $\beta \in \mathcal{B} \sim \mathcal{U}(-1.2,-1) \cup \mathcal{U}(1,1.2)$.
$\{(\alpha,\beta) \in (\mathcal{A}, \mathcal{B}) \sim (\mathcal{U}(-1,1.2)), \mathcal{U}(1,1.2)) \cup (\mathcal{U}(1,1.2)), \mathcal{U}(-1.2,1)) \cup (\mathcal{U}(-1.2,1)), \mathcal{U}(-1.2,-1)) \cup (\mathcal{U}(-1.2,-1)), \mathcal{U}(-1,1.2))\}.$

\section{Policy Design}
\label{sec:policy_designs}
We show the detailed policy design in this section. All the baseline policies were trained to minimize the trajectory reproduction error.
\newline
\noindent \textbf{KODex:}
The representation of the system is given as: $\mathrm{x}_r = [x_r^1, x_r^2, \cdots, x_r^n]$ and $\mathrm{x}_o = [x_o^1, x_o^2, \cdots, x_o^m]$ and superscript is used to index states. The details of the state design for each task is provided in Appendices~\ref{Appendix:state_design}. In experiments, the vector-valued lifting functions $\psi_r$ and $\psi_o$ in (\ref{eqn:gx}) were polynomial basis function defined as
\begin{equation}
\label{eqn:lift}
\begin{split}
\psi_r =& \{x_r^ix_r^j\} \cup \{(x_r^i)^3\} \text{ for }i,j=1,\cdots, n\\
\psi_o =& \{x_o^ix_o^j\} \cup \{(x_o^i)^2(x_o^j)\}  \text{ for } i,j=1,\cdots, m
\end{split}
\end{equation}
Note that $x_r^ix_r^j$/$x_r^jx_r^i$ only appears once in lifting functions (similar to $x_o^ix_o^j$/$x_o^jx_o^i$), and we ignore $t$ as the lifting functions are the same across the time horizon. 

The choice of lifting functions can be viewed as the hyper-parameter of KODex. We make this choice as inspired from 
\cite{abraham2017model} and experimental results also indicate its effectiveness. Through all the experiments, we sticked with the same set of lifting functions, which helped to relieve us from extensive efforts of tuning the hyper-parameters, e.g. network layer size, that were necessary for baseline policies as shown in Appendices~\ref{appendix:layer_size}.

\noindent \textbf{Full-connected Neural Network (NN):} The first baseline is a feedforward network that ingests the states $\mathrm{x}(1)$ and iteratively produces the predictions $\mathrm{x}(t), t=2,\cdots, T$ via the rollout of a Multilayer Perceptron (MLP).  The reference joint trajectories ($\mathrm{x}_r(t)$) are then used to execute the robot with the learned controller $C$. The significance of this baseline is to evaluate a policy that produces a high-dimensional motion without any additional structure.  
\newline
\textbf{Long Short-Term Memory (LSTM):} We create an LSTM-based policy under the same input-output flow as the NN policy. We also apply two fully-connected layers between the task input/output and the input/hidden state of the LSTM network. Similarly, the same controller $C$ is deployed to track the reference joint trajectory.
LSTM networks are known to be beneficial to imitation learning \cite{mandlekar2021matters} and suitable for sequential processing \cite{greff2016lstm}, e.g, motion generation. Therefore, we expect to evaluate the performance of the recurrent structures in these tasks.
\newline
\textbf{Neural Dynamic Policy (NDP):}
The Neural Dynamic Policy \cite{bahl2020neural} embeds desired dynamical structure as a layer in neural networks. Specifically, the parameters of the second order Dynamics Motion Primitives (DMP) are predicted as outputs of the preceding layers (MLP in \cite{bahl2020neural}). As a result, it allows the overall policy easily reason in the space of trajectories and can be utilized for learning from demonstration. We train an NDP policy following the imitation learning pipeline described in \cite{bahl2020neural}. For each task, given $\mathrm{x}(1)$, the neural network components in NDP generate the parameters of DMPs (radial basis functions (RBFs) in \cite{bahl2020neural}), which are forward integrated to produce the reference joint trajectories for tracking.  
% \newline
% \textbf{Riemannian Motion policy (RMP):}
\newline
\textbf{Neural Geometric Fabrics policy (NGF):} The Neural Geometric Fabrics~\cite{xieneural}, a structured policy class, that enables efficient skill learning for dexterous manipulation from demonstrations by leveraging structures induced by Geometric Fabrics~\cite{van2022geometric}. Geometric Fabrics is a stable class of the Riemannian Motion Policy (RMP)~\cite{cheng2020RmpflowJournalArxiv}. It has been demonstrated that NGF outperforms RMP in policy learning for dexterous manipulation task in~\cite{xieneural}. The NGF policy is defined in the configuration space of the robot, which is composed of a geometric policy, a potential policy and a damping term. More specifically, the NGF policy is constructed as follows: (1) define a geometric policy pair $[\mathbf{M}, \pi]$ and a potential policy pair $[\mathbf{M}_f, \pi_f]$ in the configuration space $\mathbf{q}$, (2) energize the geometric policy (project orthogonal to the direction of motion with $\mathbf{p}_e$) to create a collection of energy-preserving paths (the Geometric Fabric), and (3) force the Geometric Fabric with a potential defined by $[\mathbf{M}_f, \pi_f]$ and damp via $b$ applied along $\dot{\mathbf{q}}$, which ensures convergence to the potential's minima. The potential policy $\pi_f$ is the gradient of a function of position only. Note that we parameterize the geometric policy pair $[\mathbf{M}, \pi]$, the potential policy pair $[\mathbf{M}_f, \pi_f]$, and the damping scalar $b$ with MLP networks and learn them from demonstration data.
% \newline
% \textbf{Natural Policy Gradient (NPG) \& Demo Augmented Policy Gradient (DAPG):} For both methods, the policy design details can be found in \cite{Rajeswaran2018DAPG}. 
\section{Optimizing baseline model size}
\label{appendix:layer_size}
As described in Appendices~\ref{sec:policy_designs}, we sticked with the same set of lifting functions for KODex and report the task success rate when we trained KODex on training set and tested it on validation set in Table.~\ref{tab:KODEX_all_demo}. However, for baselines, the hyper-parameters were selected through a set of ablation experiments for each task using the training set over three choices of model size, including small size, median size and large size. We generated five random seeds for parameter initialization per model size, per baseline, and per task, as all learning based baseline models are sensitive to parameter initialization \cite{henderson2018deep}. For each baseline policy, we report the mean and standard deviation of the task success rate on the validation set over five random seeds in Tables.~\ref{tab:NN_size}-\ref{tab:NGF_size}.

Based on these results, we selectd the model size that offers the best performance in terms of task success rate. In addition, these results indicate that, unlike KODex, extensive hyper-parameter tuning and various trials on parameter initialization for baseline models are necessary. Note that we use $l$ to denote dim$(\mathrm{x}(t))$.
\label{Hyper_Parameters}
 \begin{table}[!h]
    \caption{Task success rate on validation set (KODex)}
    \centering
    \begin{tabular}{c|c|c|c}
    \hline
  Tool  &  Door & Relocation & Reorientation \\
        \hline
    % 62.0\% & 88.0\%  & 96.0\% & 100.0\% \\
    100.0\% & 96.0\% & 88.0\% & 62.0\%\\
        \hline
    \end{tabular}
    \label{tab:KODEX_all_demo}
\end{table}
% \begin{table}[h]
% \scriptsize
%     \caption{Hyper-parameters on NN Network Sizes}
%     \centering
%     \begin{tabular}{c|c|c|c|c}
%     \hline
%         \multirow{3}{*}{\diagbox[width=10em,height=3\line]{Model Size}{Success Rate}{Task}} & \multirow{3}{*}{Reorientation} & \multirow{3}{*}{Relocation} & \multirow{3}{*}{Door} & \multirow{3}{*}{Tool}\\ 
%             (\%) &  & &  & \\
%             &  & &  & \\
%         \hline
%     MLP: (32, 64, 32)    &  6.8($\pm$3.9) & 0.4($\pm$0.8) & 0.0($\pm$0.0) & \textbf{0.4($\pm$0.8)} \\
%         \hline
%     MLP: (64, 128, 64) & \textbf{10.4($\pm$6.6)}  & \textbf{1.2($\pm$2.4)} & \textbf{0.4($\pm$0.8)}  & 0.0($\pm$0.0) \\
%         \hline
%    MLP: (128, 256, 128)    &  6.0($\pm$1.5)  & 0.8($\pm$1.6)  &  0.0($\pm$0.0)  & 0.0($\pm$0.0) \\
%     \hline
%     \end{tabular}
%     \label{tab:NN_size}
% \end{table}
\begin{table}[!h]
    \caption{Hyper-parameters on NN Network Sizes}
    \centering
    \begin{tabular}{c|c|c|c|c}
    \hline
        \multirow{3}{*}{\diagbox[width=10em,height=3\line]{Model Size}{Success Rate}{Task}} & \multirow{3}{*}{Tool} & \multirow{3}{*}{Door} & \multirow{3}{*}{Relocation} & \multirow{3}{*}{Reorientation}\\ 
            (\%) &  & &  & \\
            &  & &  & \\
        \hline
    % MLP: (32, 64, 32)    &  6.8($\pm$3.9) & 0.4($\pm$0.8) & 0.0($\pm$0.0) & \textbf{0.4($\pm$0.8)} \\
    MLP: (32, 64, 32)    &  \textbf{0.4($\pm$0.8)} & 0.0($\pm$0.0) & 0.4($\pm$0.8) & 6.8($\pm$3.9) \\
        \hline
    % MLP: (64, 128, 64) & \textbf{10.4($\pm$6.6)}  & \textbf{1.2($\pm$2.4)} & \textbf{0.4($\pm$0.8)}  & 0.0($\pm$0.0) \\
    MLP: (64, 128, 64) & 0.0($\pm$0.0)  & \textbf{0.4($\pm$0.8)} & \textbf{1.2($\pm$2.4)}  & \textbf{10.4($\pm$6.6)} \\
        \hline
   % MLP: (128, 256, 128)    &  6.0($\pm$1.5)  & 0.8($\pm$1.6)  &  0.0($\pm$0.0)  & 0.0($\pm$0.0)
      MLP: (128, 256, 128)    &  0.0($\pm$0.0)  & 0.0($\pm$0.0)  &  0.8($\pm$1.6)  & 6.0($\pm$1.5)
   \\
    \hline
    \end{tabular}
    \label{tab:NN_size}
\end{table}
% \begin{table}[h]
% \scriptsize
%     \caption{Hyper-parameters on LSTM Network Sizes}
%     \centering
%     \begin{tabular}{c|c|c|c|c}
%     \hline
%         \multirow{3}{*}{\diagbox[width=10em,height=3\line]{Model Size}{Success Rate}{Task}} & \multirow{3}{*}{Reorientation} & \multirow{3}{*}{Relocation} & \multirow{3}{*}{Door} & \multirow{3}{*}{Tool}\\ 
%            (\%) &  & &  & \\
%             &  & &  & \\
%         \hline
%     LSTM: 200 &  \multirow{2}{*}{\textbf{56.4($\pm$7.4)}} & \multirow{2}{*}{7.6($\pm$5.9)} & \multirow{2}{*}{\textbf{87.6($\pm$10.3)}} &  \multirow{2}{*}{28.8($\pm$25.0)}\\
%     fc: ($l$, 100), (200, $l$) & & & & \\
%         \hline
%     LSTM: 250 &  \multirow{2}{*}{48.0($\pm$17.0)} & \multirow{2}{*}{7.6($\pm$7.5)} & \multirow{2}{*}{80.8($\pm$24.5)} &  \multirow{2}{*}{\textbf{60.8($\pm$36.6)}}\\
%     fc: ($l$, 175), (250, $l$) & & & & \\
%         \hline
%        LSTM: 300 &  \multirow{2}{*}{54.0($\pm$11.0)} & \multirow{2}{*}{\textbf{16.4($\pm$14.5)}} & \multirow{2}{*}{82.0($\pm$13.9)}  & \multirow{2}{*}{44.8($\pm$31.8)}\\
%     fc: ($l$, 250), (300, $l$) & & & & \\
%     \hline
%     \end{tabular}
%     \label{tab:LSTM_size}
% \end{table}

\begin{table}[!h]
    \caption{Hyper-parameters on LSTM Network Sizes}
    \centering
    \begin{tabular}{c|c|c|c|c}
    \hline
        \multirow{3}{*}{\diagbox[width=10em,height=3\line]{Model Size}{Success Rate}{Task}} & \multirow{3}{*}{Tool} & \multirow{3}{*}{Door} & \multirow{3}{*}{Relocation} & \multirow{3}{*}{Reorientation}\\ 
           (\%) &  & &  & \\
            &  & &  & \\
        \hline
    % LSTM: 200 &  \multirow{2}{*}{\textbf{56.4($\pm$7.4)}} & \multirow{2}{*}{7.6($\pm$5.9)} & \multirow{2}{*}{\textbf{87.6($\pm$10.3)}} &  \multirow{2}{*}{28.8($\pm$25.0)}\\
    % fc: ($l$, 100), (200, $l$) & & & & \\
    LSTM: 200 &  \multirow{2}{*}{28.8($\pm$25.0)} & \multirow{2}{*}{\textbf{87.6($\pm$10.3)}} & \multirow{2}{*}{7.6($\pm$5.9)} &  \multirow{2}{*}{\textbf{56.4($\pm$7.4)}}\\
    fc: ($l$, 100), (200, $l$) & & & & \\
        \hline
    % LSTM: 250 &  \multirow{2}{*}{48.0($\pm$17.0)} & \multirow{2}{*}{7.6($\pm$7.5)} & \multirow{2}{*}{80.8($\pm$24.5)} &  \multirow{2}{*}{\textbf{60.8($\pm$36.6)}}\\
    % fc: ($l$, 175), (250, $l$) & & & & \\
LSTM: 250 &  \multirow{2}{*}{\textbf{60.8($\pm$36.6)}} & \multirow{2}{*}{80.8($\pm$24.5)} & \multirow{2}{*}{7.6($\pm$7.5)} &  \multirow{2}{*}{48.0($\pm$17.0)}\\
    fc: ($l$, 175), (250, $l$) & & & & \\
        \hline
       % LSTM: 300 &  \multirow{2}{*}{54.0($\pm$11.0)} & \multirow{2}{*}{\textbf{16.4($\pm$14.5)}} & \multirow{2}{*}{82.0($\pm$13.9)}  & \multirow{2}{*}{44.8($\pm$31.8)}\\
       LSTM: 300 &  \multirow{2}{*}{44.8($\pm$31.8)} & \multirow{2}{*}{82.0($\pm$13.9)} & \multirow{2}{*}{\textbf{16.4($\pm$14.5)}}  & \multirow{2}{*}{54.0($\pm$11.0)}\\
    fc: ($l$, 250), (300, $l$) & & & & \\
    \hline
    \end{tabular}
    \label{tab:LSTM_size}
\end{table}

% \begin{table}[h]
% \scriptsize
%     \caption{Hyper-parameters on NDP Network Sizes}
%     \centering
%     \begin{tabular}{c|c|c|c|c}
%     \hline
%         \multirow{3}{*}{\diagbox[width=10em,height=3\line]{Model Size}{Success Rate}{Task}} & \multirow{3}{*}{Reorientation} & \multirow{3}{*}{Relocation} & \multirow{3}{*}{Door} & \multirow{3}{*}{Tool}\\ 
%           (\%)  &  & &  & \\
%             &  & &  & \\
%         \hline
%     MLP: (32, 64, 32) &  \multirow{2}{*}{57.2($\pm$8.6)} & \multirow{2}{*}{30.0($\pm$9.3)} & \multirow{2}{*}{8.0($\pm$2.5)} & \multirow{2}{*}{0.0($\pm$0.0)}\\
%     10 RBFs & & & & \\
%         \hline
%     MLP: (64, 128, 64) &  \multirow{2}{*}{59.2($\pm$6.5)} & \multirow{2}{*}{74.0($\pm$4.9)} & \multirow{2}{*}{40.8($\pm$8.1)} & \multirow{2}{*}{16.8($\pm$29.8)}\\
%     20 RBFs & & & & \\
%         \hline
%     MLP: (128, 256, 128) &  \multirow{2}{*}{\textbf{62.4($\pm$7.8)}} & \multirow{2}{*}{\textbf{79.2($\pm$7.7)}} & \multirow{2}{*}{\textbf{66.0($\pm$5.2)}} & \multirow{2}{*}{\textbf{18.4($\pm$31.9)}}\\
%     30 RBFs & & & & \\
%     \hline
%     \end{tabular}
%     \label{tab:NDp_size}
% \end{table}

\begin{table}[!h]
    \caption{Hyper-parameters on NDP Network Sizes}
    \centering
    \begin{tabular}{c|c|c|c|c}
    \hline
        \multirow{3}{*}{\diagbox[width=10em,height=3\line]{Model Size}{Success Rate}{Task}} & \multirow{3}{*}{Tool} & \multirow{3}{*}{Door} & \multirow{3}{*}{Relocation} & \multirow{3}{*}{Reorientation}\\ 
          (\%)  &  & &  & \\
            &  & &  & \\
        \hline
    % MLP: (32, 64, 32) &  \multirow{2}{*}{57.2($\pm$8.6)} & \multirow{2}{*}{30.0($\pm$9.3)} & \multirow{2}{*}{8.0($\pm$2.5)} & \multirow{2}{*}{0.0($\pm$0.0)}\\
    MLP: (32, 64, 32) &  \multirow{2}{*}{0.0($\pm$0.0)} & \multirow{2}{*}{8.0($\pm$2.5)} & \multirow{2}{*}{30.0($\pm$9.3)} & \multirow{2}{*}{57.2($\pm$8.6)}\\
    10 RBFs & & & & \\
        \hline
    % MLP: (64, 128, 64) &  \multirow{2}{*}{59.2($\pm$6.5)} & \multirow{2}{*}{74.0($\pm$4.9)} & \multirow{2}{*}{40.8($\pm$8.1)} & \multirow{2}{*}{16.8($\pm$29.8)}\\
    MLP: (64, 128, 64) &  \multirow{2}{*}{16.8($\pm$29.8)} & \multirow{2}{*}{40.8($\pm$8.1)} & \multirow{2}{*}{74.0($\pm$4.9)} & \multirow{2}{*}{59.2($\pm$6.5)}\\
    20 RBFs & & & & \\
        \hline
    % MLP: (128, 256, 128) &  \multirow{2}{*}{\textbf{62.4($\pm$7.8)}} & \multirow{2}{*}{\textbf{79.2($\pm$7.7)}} & \multirow{2}{*}{\textbf{66.0($\pm$5.2)}} & \multirow{2}{*}{\textbf{18.4($\pm$31.9)}}\\
    MLP: (128, 256, 128) &  \multirow{2}{*}{\textbf{18.4($\pm$31.9)}} & \multirow{2}{*}{\textbf{66.0($\pm$5.2)}} & \multirow{2}{*}{\textbf{79.2($\pm$7.7)}} & \multirow{2}{*}{\textbf{62.4($\pm$7.8)}}\\
    30 RBFs & & & & \\
    \hline
    \end{tabular}
    \label{tab:NDp_size}
\end{table}

% \begin{table}[h]
%     \caption{Hyper-parameters on NGF Network Sizes}
%     \centering
%     \begin{tabular}{c|c|c|c|c}
%     \hline
%         \multirow{3}{*}{\diagbox[width=10em,height=3\line]{Model Size}{Success Rate}{Task}} & \multirow{3}{*}{Reorientation} & \multirow{3}{*}{Relocation} & \multirow{3}{*}{Door} & \multirow{3}{*}{Tool}\\ 
%            (\%) &  & &  & \\
%             &  & &  & \\
%         \hline
%     MLP: (64, 32)    &  77.6($\pm$2.3) & 87.6($\pm$8.5) & 87.2($\pm$12.0) & 99.2($\pm$1.6) \\
%         \hline
%     MLP: (128, 64) & 72.4($\pm$4.5)  & 94.4($\pm$3.2) & 90.0($\pm$5.9)  & \textbf{100.0($\pm$0.0)} \\
%         \hline
%    MLP: (256, 128)    &  \textbf{78.4($\pm$3.4)}  & \textbf{95.2($\pm$1.6)}  &  \textbf{90.8($\pm$4.3)}  & 83.6($\pm$20.1) \\
%     \hline
%     \end{tabular}
%     \label{tab:NGF_size}
% \end{table}

\begin{table}[!h]
    \caption{Hyper-parameters on NGF Network Sizes}
    \centering
    \begin{tabular}{c|c|c|c|c}
    \hline
        \multirow{3}{*}{\diagbox[width=10em,height=3\line]{Model Size}{Success Rate}{Task}} & \multirow{3}{*}{Tool} & \multirow{3}{*}{Door} & \multirow{3}{*}{Relocation} & \multirow{3}{*}{Reorientation}\\ 
           (\%) &  & &  & \\
            &  & &  & \\
        \hline
    % MLP: (64, 32)    &  77.6($\pm$2.3) & 87.6($\pm$8.5) & 87.2($\pm$12.0) & 99.2($\pm$1.6) \\
    MLP: (64, 32)    &  99.2($\pm$1.6) & 87.2($\pm$12.0) & 87.6($\pm$8.5) & 77.6($\pm$2.3) \\
        \hline
    % MLP: (128, 64) & 72.4($\pm$4.5)  & 94.4($\pm$3.2) & 90.0($\pm$5.9)  & \textbf{100.0($\pm$0.0)} \\
    MLP: (128, 64) & \textbf{100.0($\pm$0.0)}  & 90.0($\pm$5.9) & 94.4($\pm$3.2)  & 72.4($\pm$4.5) \\
        \hline
   % MLP: (256, 128)    &  \textbf{78.4($\pm$3.4)}  & \textbf{95.2($\pm$1.6)}  &  \textbf{90.8($\pm$4.3)}  & 83.6($\pm$20.1) \\
   MLP: (256, 128)    &  83.6($\pm$20.1)  & \textbf{90.8($\pm$4.3)}  &  \textbf{95.2($\pm$1.6)}  & \textbf{78.4($\pm$3.4)} \\
    \hline
    \end{tabular}
    \label{tab:NGF_size}
\end{table}

\section{Hyper-parameters for controller learning}
\label{appendix:controller}
The hyper-parameters we used to learn the inverse dynamics controller $C$ for each task were the same as listed in Table.~\ref{tab:controller}. Note 
 that we use $l_r$ to denote dim$(\mathrm{x}_r(t))$.
 \begin{table}[!h]
    \caption{Hyper-parameters on controller learning}
    \centering
    \begin{tabular}{c|c|c|c}
    \hline
   Hidden Layer   &  Activation & Learning Rate & Iteration \\
        \hline
    $(4l_r, 4l_r, 2l_r)$ & ReLU  & 0.0001 & 300 \\
        \hline
    \end{tabular}
    \label{tab:controller}
\end{table}
\textcolor{black}{
\section{Inference Time}
\label{sec:inference}
 \begin{table}[t]
 \scriptsize
    \caption{\textcolor{black}{Report of inference time (unit: millisecond). We show the mean and standard deviation of the one-step inference time over task horizon.}}
    \centering
    \begin{tabular}{c|c|c|c|c}
    \hline
        \multirow{2}{*}{\diagbox[width=8em,height=2\line]{Policy}{Task}} & \multirow{2}{*}{Tool} & \multirow{2}{*}{Door} & \multirow{2}{*}{Relocation} & \multirow{2}{*}{Reorientation} \\ 
             &   &  &  & \\
        \hline
    NN & 1.39($\pm$0.26)  & 1.26($\pm$0.39) & 1.02($\pm$0.09) & 1.15($\pm$0.12) \\
    \hline
    LSTM & 1.71($\pm$0.28)  & 1.32($\pm$0.34) & 1.59($\pm$0.57) & 1.42($\pm$0.13) \\
    \hline
    NDP & 1.88($\pm$0.30)  & 1.08($\pm$0.22) & 1.05($\pm$0.06) & 1.32($\pm$0.21) \\
    \hline
    NGF & 1.37($\pm$0.16)  & 1.17($\pm$0.26) & 1.72($\pm$0.36) & 1.19($\pm$0.12) \\
    \hline
    KODex & 1.71($\pm$0.48)  & 1.04($\pm$0.27) & 1.12($\pm$0.24) & 1.08($\pm$0.60) \\
    \hline
    \end{tabular}
    \label{tab:inference_time}
\end{table}
We report the inference time for each method in Table.~\ref{tab:inference_time}. Our results indicate that KODex’s inference time is on par with the SOTA baselines. As such, it reveals that KODex can be translated to physical hardware and meet necessary control frequency. 
}
\section{Zero-Shot Out-of-Distribution Generalization}
\label{sec:ood}
\begin{figure*}[!h]
     \centering
     \begin{subfigure}[t]{\textwidth}
     \centering
        \raisebox{-\height}{\includegraphics[width=0.8\textwidth]{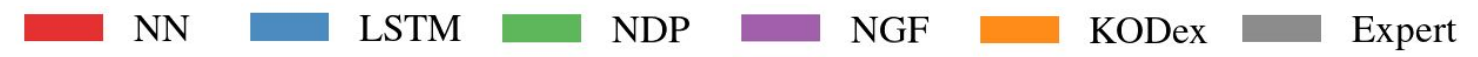}}
    \end{subfigure}
%%%%%%%%%%%%%%%%%%%%%%%%%%%%%%%%%%%%first row
        \begin{subfigure}[t]{0.5\textwidth}
        \raisebox{-\height}{\includegraphics[width=\textwidth]{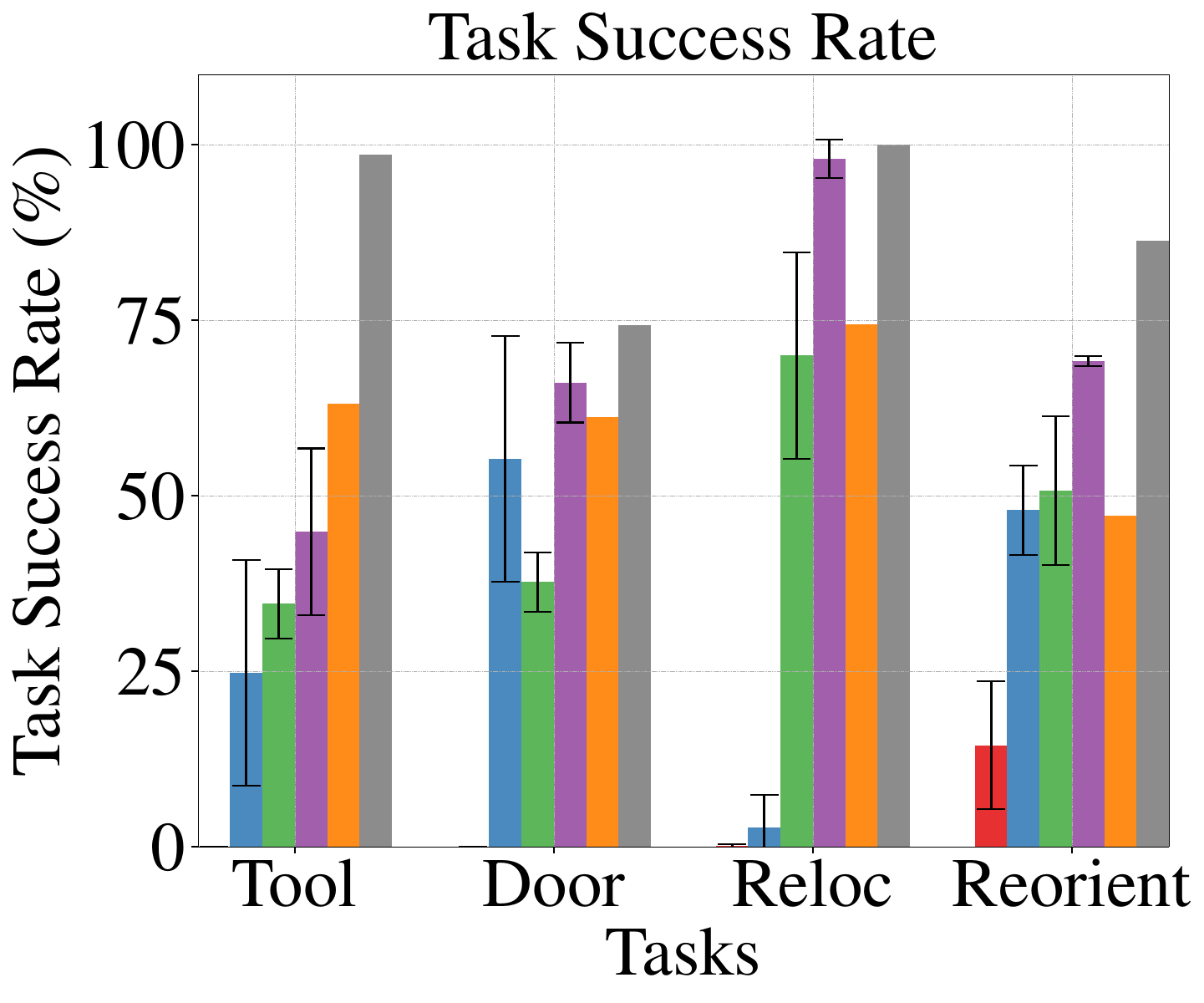}}
    \end{subfigure}
    \caption{Zero-Shot Out-of-distribution task success rates}
\label{fig:OOD_success_rate}
\end{figure*}

We generated a new set of 10,000 out-of-distribution samples to evaluate how the policies that were trained on 200 demonstrations generalize to unseen samples (see Appendices~\ref{Appendix:distribution} for details on the sampling procedure). In Fig.~\ref{fig:OOD_success_rate}, we report the task success rates of each method trained on the 200 demonstrations and tested on the 10,000 out-of-distribution samples. In addition, we also report the task success rate of the expert policy on the same 10,000 out-of-distribution samples to establish a baseline. Perhaps unsurprisingly, none of the methods are able to consistently outperform the expert policy in most tasks. We observe that KODex is able to outperform the four baselines in Tool Use task. In the other tasks, the highly-structured NGF performs the best, and KODex's performs comparably to NDP and LSTM. 

\section{Robustness to changes in physical properties}
\label{sec:robustness}
% We also evaluated the robustness of the reference dynamics KODex to variations in the environment. 
We evaluate the robustness of the reference dynamics learned by each method to changes in hand mass or object mass for each task. This experiment is motivated by the fact that sim-to-real transfer often involves changes in physical properties. Further, consistent use of robotic hardware could result in changes to physical properties. Specifically, we consider four variations per task: 

\begin{itemize}
    \item Tool Use: i) \textit{Heavy Object (Hammer)}: 0.25 (default) $\rightarrow$ 0.85 (new), ii) \textit{Light Object (Hammer)}: 0.25 (default) $\rightarrow$ 0.10 (new), iii) \textit{Light Hand (Palm)}: 4.0 (default) $\rightarrow$ 1.0 (new), and iv) \textit{Heavy Hand (Palm)}: 4.0 (default) $\rightarrow$ 8.0 (new)

    \item Door: i) \textit{Heavy Object (Latch)}: 3.54 (default) $\rightarrow$ 12.54 (new), ii) \textit{Light Object (Latch)}: 3.54 (default) $\rightarrow$ 0.54 (new), iii) \textit{Light Hand (Palm)}: 4.0 (default) $\rightarrow$ 1.5 (new), and iv) \textit{Heavy Hand (Palm)}: 4.0 (default) $\rightarrow$ 7.0 (new)

    \item Relocation: i) \textit{Heavy Object (Ball)}: 0.18 (default) $\rightarrow$ 1.88 (new), ii) \textit{Light Object (Ball)}: 0.18 (default) $\rightarrow$ 0.05 (new), iii) \textit{Light Hand (Palm)}: 4.0 (default) $\rightarrow$ 3.0 (new), and iv) \textit{Heavy Hand (Palm)}: 4.0 (default) $\rightarrow$ 5.0 (new); 

    \item Reorientation: i) \textit{Heavy Object (Pen)}: 1.5 (default) $\rightarrow$ 9.5 (new), ii) \textit{Light Object (Pen)}: 1.5 (default) $\rightarrow$ 0.2 (new), iii) \textit{Light Hand (Finger Knuckles)}: 0.008 (default) $\rightarrow$ 0.0001 (new), and iv) \textit{Heavy Hand (Finger Knuckles)}: 0.008 (default) $\rightarrow$ 0.20 (new)
\end{itemize}
It is important to note we held the reference dynamics learned by each method constant for this experiment, irrespective of the changes to the hand or the object. Instead, we relearned the tracking controller using 200 rollouts from the expert agent, following the procedure detailed in Section.~\ref{sec:control}. 
% To achieve this, we collected another 200 rollouts from the expert agent deployed in the new environment to relearn the inverse dynamics controller through the procedures shown in Section.~\ref{sec:control}. 
% In order to evaluate the influence of demonstration quality to controller learning, we used noiseless and noisy expert agents to collect demonstrations and trained the inverse dynamics controller respectively.

% \vspace{3pt}
% \noindent \textbf{Analysis}:
In Tables.~\ref{tab:Variation_test_hammer}-\ref{tab:Variation_test_pen}, we report the task success rate of KODex, and other baseline policies (all trained on 200 demonstrations) before and after relearning the controller. We also report the task success rates of the expert agents to establish baselines.

% It is no surprise that all methods suffer a drop in performance when using the original controller to handle the variations. 

We find that the Light Hand variation results in the lowest drop in performance across all methods and all tasks, thus consequently relearning controllers does not offer any considerable improvements.
In contrast, all methods benefit from relearning the controller in the Heavy Hand variations, as evidenced by the increased task success rates. Overall, we find that KODex outperforms all baselines, with the exception of NGF which performs better than KODex under a few variations and tasks. Surprisingly, KODex (and some baselines) when used with the original controller outperform the expert policy under a few variations (e.g., Heavy Object in Relocation task, and Heavy Object in Door task). We believe this is due to the fact that KODex and the baselines learn to generate and track desired trajectories separately, while the expert RL directly generates control inputs from state information. In particular, the learned desired trajectories for a given tasks are likely invariant to slight changes in physical properties. On rare occasions where this is not the case, we indeed find that fine-tuning the tracking controllers worsens the performance.

These results demonstrate that changes to the robot/system dynamics can be handled by fine tuning the tracking controller without the need for relearning the reference dynamics. Once again, KODex is able to perform comparably to or outperform SOTA approaches despite its simplicity.
% since expert's decision during the grasping phase has shorter duration compared to that of relocation phase, 

% All methods struggle the most with the Heavy Object variation, perhaps due to the complexity of learning to grasp a considerably heavier object. Though these methods with the controllers tuned by the noisy expert perform worse than the ones trained by the expert, they are still able to  outperform the noisy expert itself. 

% We are not including a variation involving a lighter object as we found that this variation did not impact the performance of the default controller.

\setlength{\tabcolsep}{1.5pt} % Default value: 6pt
 \begin{table}[!h]
    \caption{Robustness to variations in the physical properties (Tool Use)}
    \centering
    \begin{tabular}{c|c|c|c|c}
    \hline
        \multirow{3}{*}{\diagbox[width=18em,height=3\line]{Controller}{Success Rate}{Variation}} &  \multirow{3}{*}{Heavy Object} & \multirow{3}{*}{Light Object} & \multirow{3}{*}{Light Hand} & \multirow{3}{*}{Heavy Hand} \\ 
            (\%) &  &  &  & \\
            &  &  &  & \\
        \hline
    Expert agent & 93.5  & 66.2 & 65.4 & 71.2  \\
    \hline
    KODex + Original controller &  46.0 & 64.0 & 99.5 & 46.5 \\
    NN + Original controller &  0.0($\pm$0.0) & 0.0($\pm$0.0) & 0.0($\pm$0.0) & 0.7($\pm$1.4) \\
    LSTM + Original controller &  32.7($\pm$18.7) & 35.0($\pm$22.1) & 44.3($\pm$23.1) & 52.7($\pm$27.5) \\
    NDP + Original controller &  0.0($\pm$0.0) & 68.0($\pm$20.8) & 45.4($\pm$37.4) & 0.0($\pm$0.0) \\
    NGF + Original controller &  33.4($\pm$11.5) & 62.9($\pm$27.5) & 83.2($\pm$26.3) & 40.3($\pm$20.2) \\
    \hline
    KODex + Expert-tuned controller &  53.5 & 44.0 & 89.0 & 92.5  \\
    NN + Expert-tuned controller &  0.0($\pm$0.0) & 0.0($\pm$0.0) & 0.2($\pm$0.4) & 0.0($\pm$0.0)  \\
    LSTM + Expert-tuned controller &  42.4($\pm$34.3) & 33.7($\pm$14.9) & 52.2($\pm$22.7) & 69.9($\pm$19.4)  \\
    NDP + Expert-tuned controller &  33.3($\pm$20.0) & 23.8($\pm$24.4) & 29.4($\pm$37.1) & 39.8($\pm$24.5)  \\
    NGF + Expert-tuned controller &  48.2($\pm$18.0) & 48.7($\pm$12.2) & 94.6($\pm$8.9) & 82.1($\pm$7.5)  \\

    \hline
    \end{tabular}
    \label{tab:Variation_test_hammer}
\end{table}

 \begin{table}[!h]
    \caption{Robustness to variations in the physical properties (Door)}
    \centering
    \begin{tabular}{c|c|c|c|c}
    \hline
        \multirow{3}{*}{\diagbox[width=18em,height=3\line]{Controller}{Success Rate}{Variation}} &  \multirow{3}{*}{Heavy Object} & \multirow{3}{*}{Light Object} & \multirow{3}{*}{Light Hand} & \multirow{3}{*}{Heavy Hand} \\ 
            (\%) &  &  &  & \\
            &  &  & & \\
        \hline
    Expert agent & 45.2  &  91.7 & 82.0 & 74.9  \\
    \hline
    KODex + Original controller &  57.0 & 97.0 & 56.5 & 33.5 \\
    NN + Original controller &  0.0($\pm$0.0) & 0.2($\pm$0.4) & 1.3($\pm$2.1) & 0.0($\pm$0.0) \\
    LSTM + Original controller &  34.4($\pm$8.7) & 75.8($\pm$19.5) & 38.1($\pm$10.4) & 33.5($\pm$11.4) \\
    NDP + Original controller &  22.1($\pm$1.9) & 62.8($\pm$5.2) & 51.1($\pm$4.9) & 3.1($\pm$2.3) \\
    NGF + Original controller &  48.7($\pm$6.7) & 95.0($\pm$2.1) & 42.1($\pm$11.0) & 33.8($\pm$10.0) \\
    \hline
    KODex + Expert-tuned controller &  39.0 & 94.0 & 54.0 & 81.5  \\
    NN + Expert-tuned controller &  0.0($\pm$0.0) & 0.0($\pm$0.0) & 0.7($\pm$0.9) & 0.0($\pm$0.0)  \\
    LSTM + Expert-tuned controller &  21.2($\pm$5.3) & 75.4($\pm$18.0) & 49.2($\pm$8.1) & 56.9($\pm$18.7)  \\
    NDP + Expert-tuned controller &  15.5($\pm$3.0) & 36.2($\pm$10.6) & 25.5($\pm$4.4) & 8.8($\pm$3.0)  \\
    NGF + Expert-tuned controller &  36.6($\pm$5.1) & 95.5($\pm$1.8) & 57.7($\pm$4.7) & 77.1($\pm$6.7)  \\

    \hline
    \end{tabular}
    \label{tab:Variation_test_door}
\end{table}

 \begin{table}[!h]
    \caption{Robustness to variations in the physical properties (Relocation)}
    \centering
    \begin{tabular}{c|c|c|c|c}
    \hline
        \multirow{3}{*}{\diagbox[width=18em,height=3\line]{Controller}{Success Rate}{Variation}} &  \multirow{3}{*}{Heavy Object} & \multirow{3}{*}{Light Object} & \multirow{3}{*}{Light Hand} & \multirow{3}{*}{Heavy Hand} \\ 
            (\%) &  &  &  & \\
            &  &  &  & \\
        \hline
    Expert agent & 77.0  & 100.0 & 100.0 & 100.0  \\
    \hline
    KODex + Original controller &  19.5 & 89.5 & 82.5 & 21.5 \\
    NN + Original controller &  0.1($\pm$0.2) & 1.6($\pm$2.5) & 1.5($\pm$2.1) & 1.7($\pm$2.2) \\
    LSTM + Original controller &  0.4($\pm$0.4) & 15.4($\pm$10.7) & 9.5($\pm$8.1) & 7.7($\pm$9.4) \\
    NDP + Original controller &  13.5($\pm$5.0) & 85.6($\pm$8.1) & 72.1($\pm$9.6) & 31.6($\pm$10.0) \\
    NGF + Original controller &  25.8($\pm$4.9) & 96.4($\pm$1.4) & 96.6($\pm$0.97) & 19.3($\pm$3.8) \\
    \hline
    KODex + Expert-tuned controller &  34.0 & 93.0 & 85.0 & 89.0  \\
    NN + Expert-tuned controller &  0.2($\pm$0.4) & 0.6($\pm$0.7) & 1.4($\pm$1.8) & 1.5($\pm$2.3)  \\
    LSTM + Expert-tuned controller &  5.8($\pm$4.7) & 15.2($\pm$12.5) & 15.5($\pm$10.7) & 14.1($\pm$9.3)  \\
    NDP + Expert-tuned controller &  19.9($\pm$5.8) & 84.5($\pm$8.9) & 63.2($\pm$15.0) & 92.4($\pm$1.2)  \\
    NGF + Expert-tuned controller &  52.6($\pm$3.6) & 98.1($\pm$1.2) & 95.6($\pm$2.2) & 94.5($\pm$0.9)  \\

    \hline
    \end{tabular}
    \label{tab:Variation_test}
\end{table}

 \begin{table}[!h]
    \caption{Robustness to variations in the physical properties (Reorientation)}
    \centering
    \begin{tabular}{c|c|c|c|c}
    \hline
        \multirow{3}{*}{\diagbox[width=18em,height=3\line]{Controller}{Success Rate}{Variation}} &  \multirow{3}{*}{Heavy Object} & \multirow{3}{*}{Light Object} & \multirow{3}{*}{Light Hand} & \multirow{3}{*}{Heavy Hand} \\ 
            (\%) &  &   &  & \\
            &  &  & & \\
        \hline
    Expert agent & 46.8  & 69.0 & 95.2 & 89.7  \\
    \hline
    KODex + Original controller &  53.5 & 55.0 & 66.5 & 61.5 \\
    NN + Original controller &  4.7($\pm$2.6) & 9.6($\pm$8.1) & 9.5($\pm$6.4) & 7.9($\pm$6.5) \\
    LSTM + Original controller &  34.5($\pm$7.8) & 52.3($\pm$10.6) & 60.3($\pm$6.0) & 55.6($\pm$7.8) \\
    NDP + Original controller &  49.4($\pm$3.6) & 58.4($\pm$6.4) & 59.8($\pm$7.6) & 55.7($\pm$9.7) \\
    NGF + Original controller &  39.9($\pm$1.9) & 57.1($\pm$2.2) & 81.6($\pm$1.8) & 73.4($\pm$3.8) \\
    \hline
    KODex + Expert-tuned controller &  52.0 & 63.0 & 71.5 & 65.5  \\
    NN + Expert-tuned controller &  1.5($\pm$0.9) & 5.2($\pm$4.2) & 3.8($\pm$1.7) & 3.7($\pm$2.6)  \\
    LSTM + Expert-tuned controller &  43.5($\pm$7.9) & 47.7($\pm$8.8) & 61.4($\pm$4.2) & 54.4($\pm$5.5)  \\
    NDP + Expert-tuned controller &  55.5($\pm$5.9) & 59.0($\pm$5.5) & 63.0($\pm$6.5) & 57.0($\pm$7.5)  \\
    NGF + Expert-tuned controller &  49.1($\pm$2.6) & 59.7($\pm$3.2) & 79.4($\pm$1.9) & 72.6($\pm$1.2)  \\

    \hline
    \end{tabular}
    \label{tab:Variation_test_pen}
\end{table}

% \subsection{Baseline Training details}
% \label{Train_detail}
% We provide with additional training details for all the baseline models, namely batch size, training iteration, learning rate.
% \begin{table}[h]
%     \caption{Hyper-parameters on baseline training}
%     \centering
%     \begin{tabular}{c|c|c|c}
%     \hline
%      &Batch Size& Training Iteration& Learning Rate  \\
%         \hline
%    NN   &  8 & 200 & 0.001  \\
%         \hline
%     LSTM & 4  & 200 & 0.001 \\
%         \hline
%    NDP  &  4  & 200 & 0.001   \\
%     \hline
%     \end{tabular}
%     \label{tab:NN_size}
% \end{table}

%% Use plainnat to work nicely with natbib. 
% plainnat may cause some wrong orderings
\section{The impact of the choice of basis functions}
\label{sec:basis_choice}
\noindent We evaluate if KODex's performance is impacted by different sets of polynomial functions that are used as the lifting function. We trained all policies on 200 demos and tested them on 10,000 unseen initial conditions. 
% The design choices of each set are shown in Section.~\ref{design}, and the results and discussions are reported in Section.~\ref{result}.
% \section{Design Choice}
% In this section, we show the design choice of each set.

\noindent \textbf{Design}: Specifically, we define four sets of observables (one of which was used in the original submission). Let robot state: $\mathrm{x}_r = [x_r^1, x_r^2, \cdots, x_r^n]$ and $\mathrm{x}_o = [x_o^1, x_o^2, \cdots, x_o^m]$ denote the robot and the object state, respectively, with superscript indexing the states. We then define four vector-valued lifting functions $\psi_r$ and $\psi_o$ in (\ref{eqn:gx}) as follows
\label{design}
\begin{itemize}
    \item Set 1
\begin{equation*}
\begin{split}
\psi_r =& \{(x_r^i)^2\} \text{ for }i=1,\cdots, n\\
\psi_o =& \{(x_o^i)^2\} \text{ for }i=1,\cdots, m
\end{split}
\end{equation*}
    \item Set 2
\begin{equation*}
\begin{split}
\psi_r =& \{x_r^ix_r^j\} \text{ for }i,j=1,\cdots, n\\
\psi_o =& \{x_o^ix_o^j\} \text{ for } i,j=1,\cdots, m
\end{split}
\end{equation*}
    \item \textbf{Set 3 (used in this work)}
\begin{equation*}
\begin{split}
\psi_r =& \{x_r^ix_r^j\} \cup \{(x_r^i)^3\} \text{ for }i,j=1,\cdots, n\\
\psi_o =& \{x_o^ix_o^j\} \cup \{(x_o^i)^2(x_o^j)\}  \text{ for } i,j=1,\cdots, m
\end{split}
\end{equation*}
    \item Set 4
\begin{equation*}
\begin{split}
\psi_r =& \{x_r^ix_r^j\} \cup \{(x_r^i)^2(x_r^j)\} \text{ for }i,j=1,\cdots, n\\
\psi_o =& \{x_o^ix_o^j\} \cup \{(x_o^i)^2(x_o^j)\}  \text{ for } i,j=1,\cdots, m
\end{split}
\end{equation*}
\end{itemize}
\noindent We report the number of observables for each set and task combination in Table.~\ref{tab:tabel_dim}.
 \begin{table}[h]
    \caption{Number of observables}
    \centering
    \begin{tabular}{c|c|c|c|c}
    \hline
        \multirow{3}{*}{\diagbox[width=8em,height=2.5\line]{Set}{Task}} & \multirow{3}{*}{Tool} & \multirow{3}{*}{Door} & \multirow{3}{*}{Relocation} & \multirow{3}{*}{Reorientation} \\ 
            &  &  &  & \\
             & n=26,m=15  & n=28,m=7 & n=30,m=12 & n=24,m=12 \\
        \hline
    Set 1 & 82  & 70 & 84 & 72 \\
    \hline
    Set 2 & 512  & 469 & 585 & 414 \\
    \hline
    \textbf{Set 3} (ours) & 763  & 546 & 759 & 582 \\
    \hline
    Set 4 & 1413  & 1302 & 1629 & 1134 \\
    \hline
    \end{tabular}
    \label{tab:tabel_dim}
\end{table}

% \section{Result \& Discussion}
% \label{result}
% In this section, we include the results and discussions. 
% We use the same metrics (Training time, Imitation error, and Task success rate) to quantify the performance. 
\label{sec:basis_functions}
\begin{figure*}[!h]
\centering
\includegraphics[width=\columnwidth]{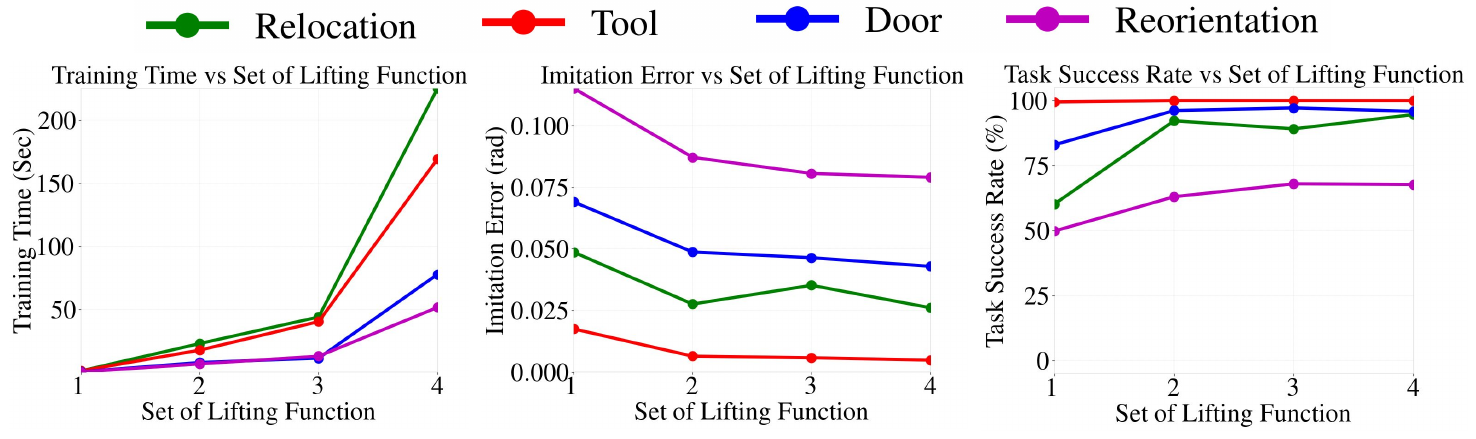}
\caption{The effects of lifting function on training time (left), imitation error (center), and success rate (right).}
\label{fig:obs_exp}
\end{figure*}
\noindent \textbf{Discussion}: As shown in Fig.~\ref{fig:obs_exp}, it is clear that training time increases with the number of observables since the Moore–Penrose inverse requires more computation for higher-dimension matrices. Importantly, KODex’s success rate across all tasks remained roughly the same for Sets 2, 3, and 4. In general, as one would expect, increasing the number of observables tends to decrease imitation error and increase task success rate. The only exception to this trend is observed for the Object Relocation task, in which KODex performs marginally better when trained on Set 2 (585 observables) compared with it trained on Set 3 (759 observables). Taken together, these results suggest that KODex's performance is not highly sensitive to the specific choice of lifting function, as long as sufficient expressivity is ensured.

\section{Stability Analysis}
\label{sec:Eigenvalue_Analysis}
Another unique advantage of utilizing Koopman Operators to model the \textcolor{black}{reference dynamics} for dexterous manipulation tasks is that the learned policy is a linear dynamical system which can be readily inspected and analyzed, in stark contrast to SOTA methods built upon deep neural networks.

We analyzed the stability of the learned policy. For a linear dynamical system with complex conjugate eigenvalues $\lambda_i = \theta_i \pm j\omega_i$, i.e., KODex with Koopman matrix $\mathbf{K}$, the system is asymptotically stable if all of the eigenvalues have magnitude ($\rho_i = \sqrt{\theta_i^2 + \omega_i^2}$) less than one. From the standpoint of control theory, it is beneficial to have a asymptotically stable system because of the guarantee that all system states will converge. However, from the standpoint of dexterous manipulation tasks considered in this work, strict stability might not be preferable because the final desired hand poses and object poses are not identical for different initial conditions. This represents a natural trade-off between safety and expressivity. As such, understanding how KODex addresses this trade-off can be illuminating.

\begin{figure*}[!h]
     \centering
%%%%%%%%%%%%%%%%%%%%%%%%%%%%%%%%%%%%first row
        \begin{subfigure}[t]{0.40\textwidth}
        \raisebox{-\height}{\includegraphics[width=\textwidth]{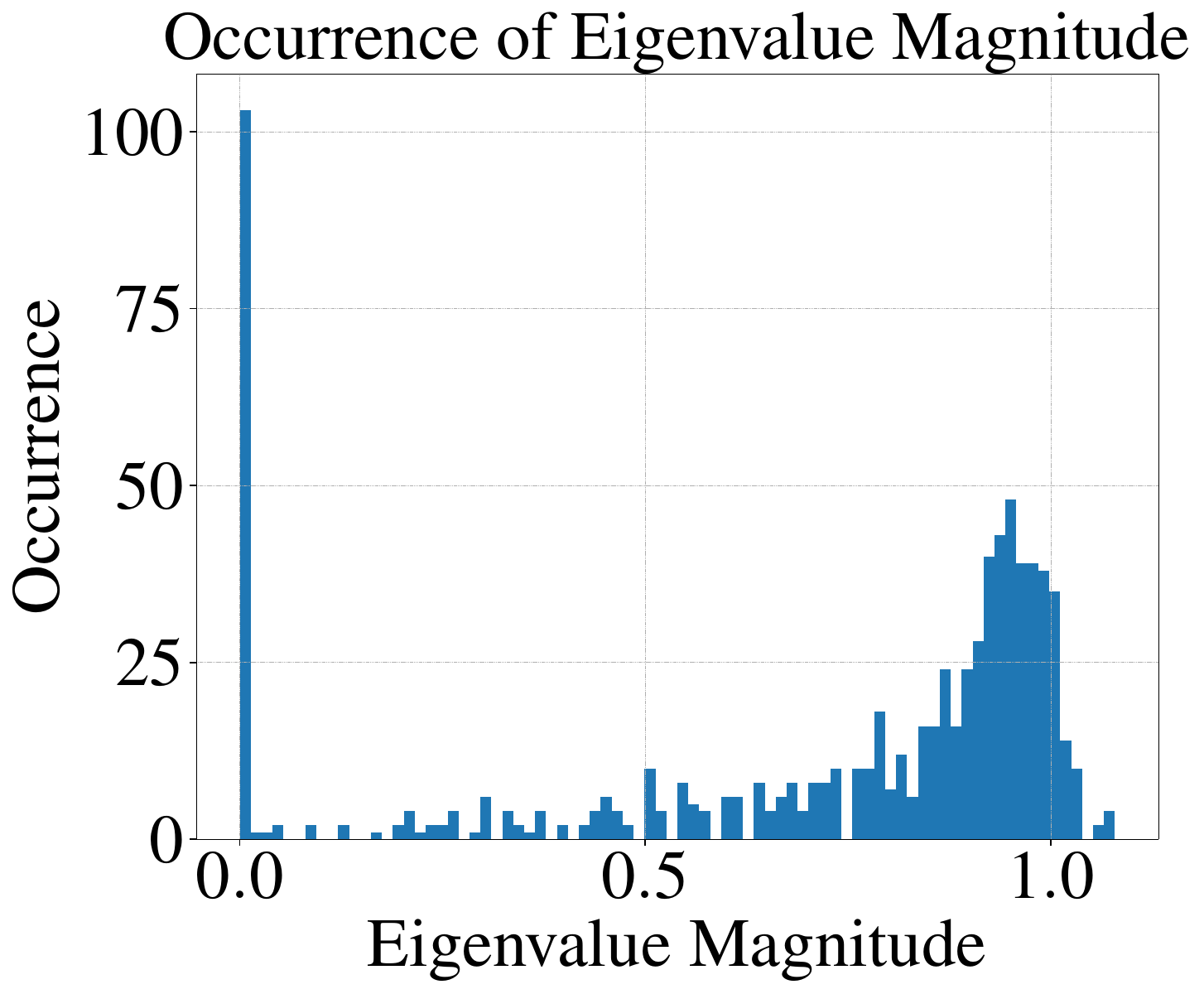}}
        \caption{Tool Use} 
    \end{subfigure}
        \begin{subfigure}[t]{0.40\textwidth}
        \raisebox{-\height}{\includegraphics[width=\textwidth]{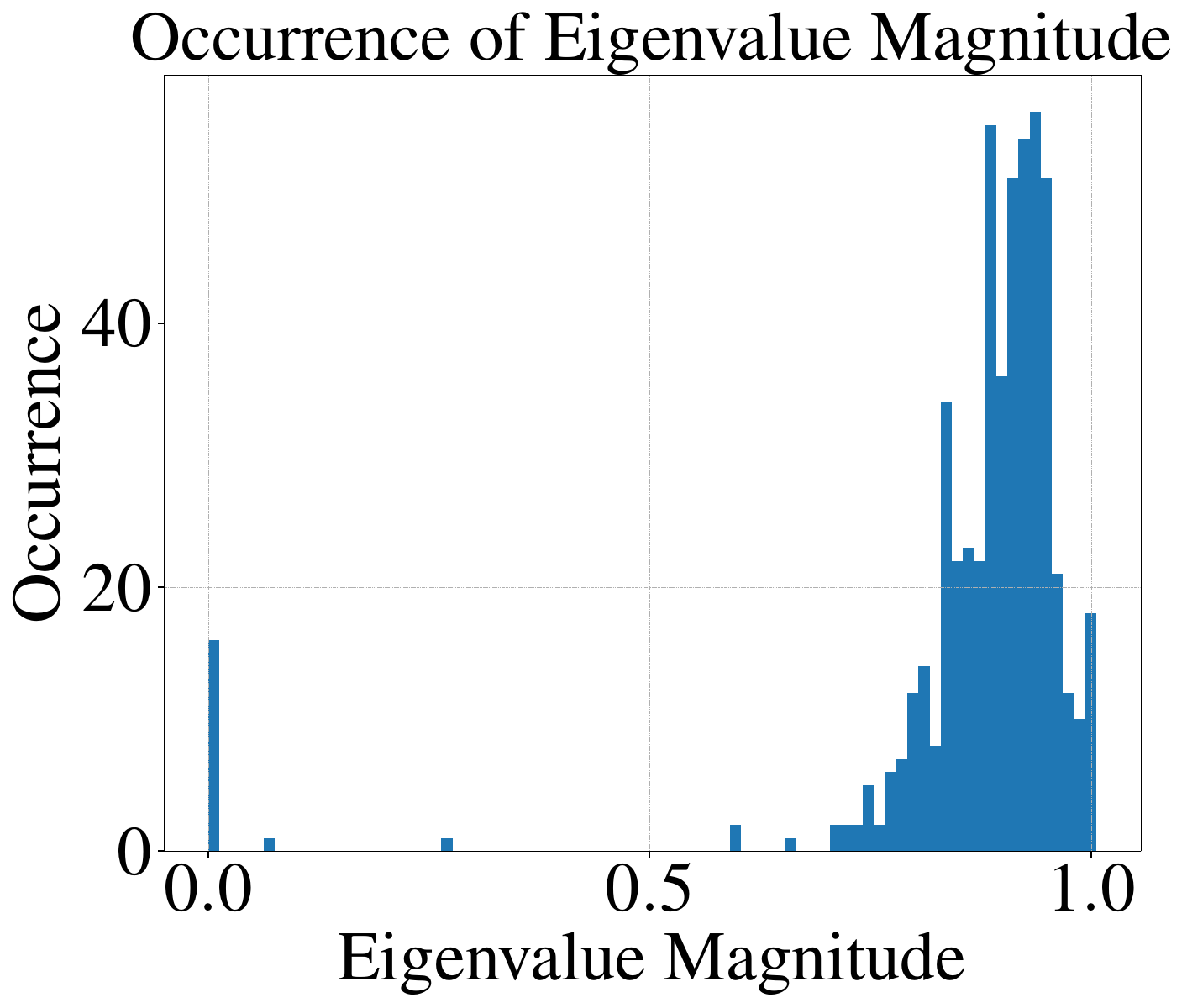}}
        \caption{Door Opening} 
    \end{subfigure}
    \begin{subfigure}[t]{0.40\textwidth}
        \raisebox{-\height}{\includegraphics[width=\textwidth]{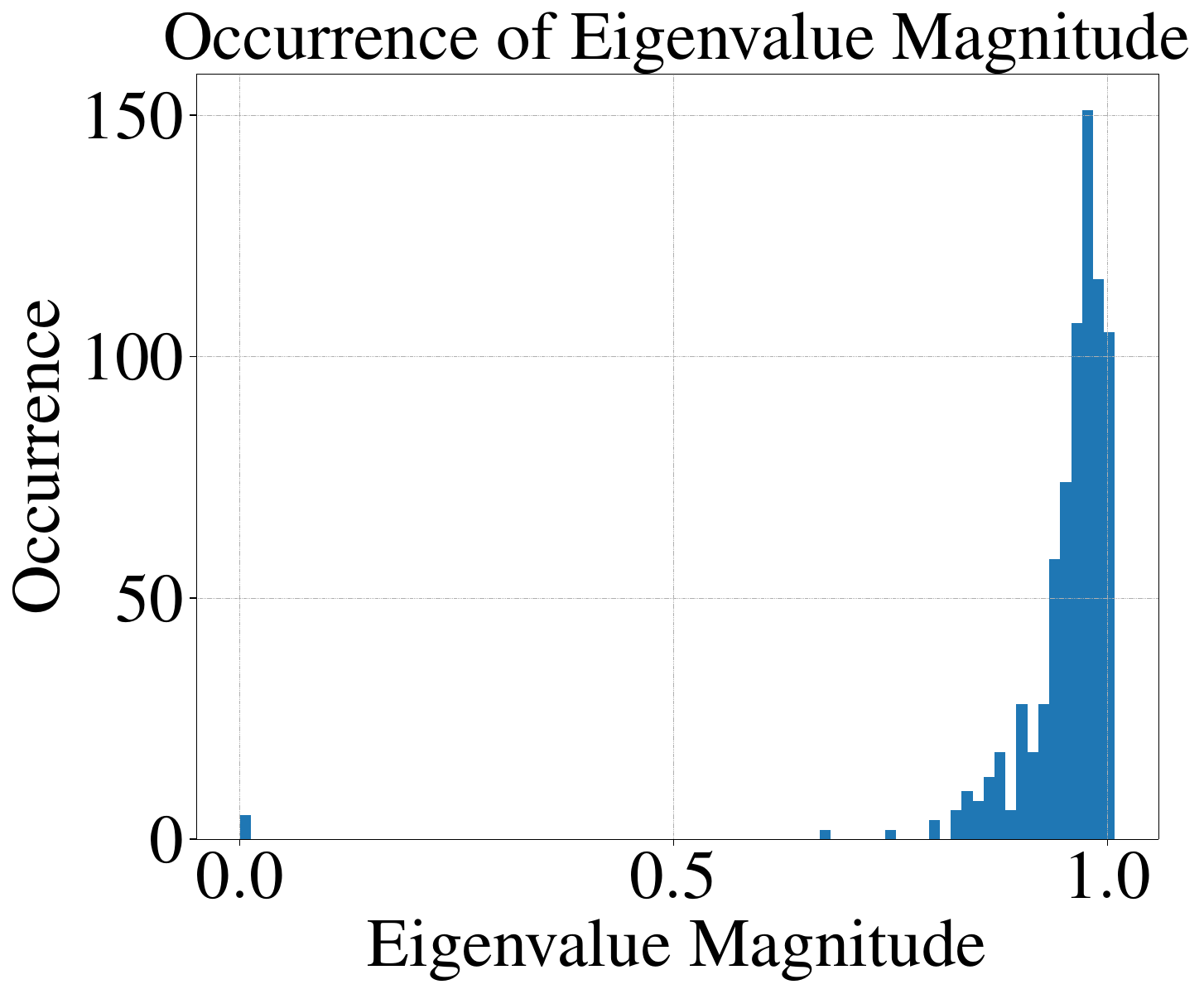}}
        \caption{Object Relocation}
    \end{subfigure}
    \begin{subfigure}[t]{0.40\textwidth}
        \raisebox{-\height}{\includegraphics[width=\textwidth]{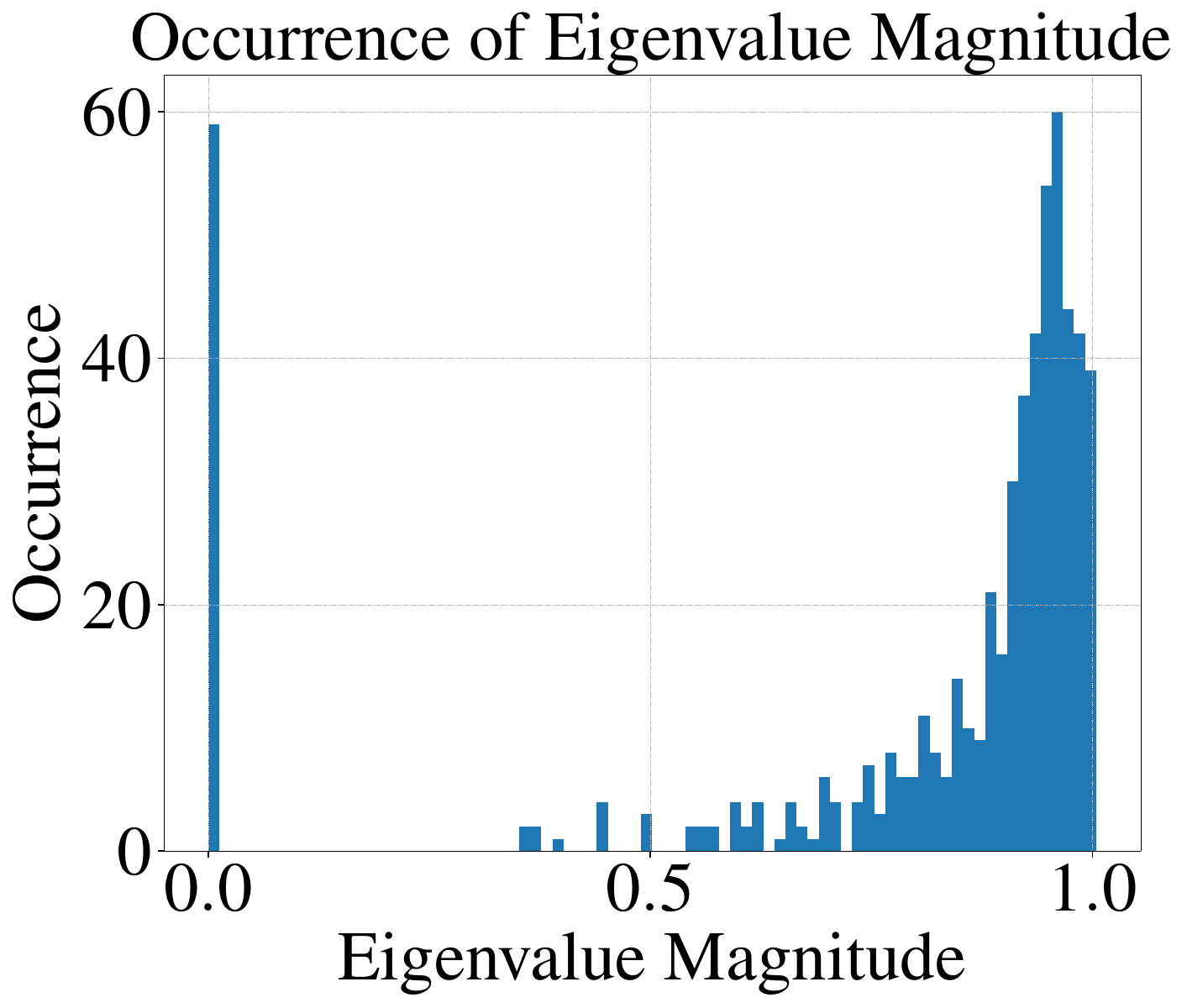}}
        \caption{In-hand Reorientation}
    \end{subfigure}
    \caption{Occurrence of Eigenvalue Magnitude}
\label{fig:eigenvalue_occur}
\end{figure*}
 \begin{table}[h]
    \caption{Maximun Eigenvalue Magnitude}
    \centering
    \begin{tabular}{c|c|c|c}
    \hline
   Tool Use   &  Door Opening & Object Relocation & In-hand Reorientation \\
        \hline
    1.07888 & 1.00553  & 1.00859 & 1.00413 \\
        \hline
    \end{tabular}
    \label{tab:maximum_eigen}
\end{table}

In Fig.~\ref{fig:eigenvalue_occur}, we report a histogram of the Koopman matrix's eigenvalue magnitudes in each task. In addition, we report the maximum eigenvalue magnitude in Table.~\ref{tab:maximum_eigen}.
Based on these results, we can see that i) most eigenvalues' magnitudes are less than one, suggesting that KODex tends to learn nearly-stable policies that generate \textit{safe trajectories} during execution, and ii) a few eigenvalues have magnitude larger than one, suggesting KODex does not prioritize stability, at the expense of expressivity required to achieve the reported performance.

% \begin{figure*}[hbt!]
% \centering
% \includegraphics[scale=0.65]{images/Result.pdf}
% \caption{The effects of lifting function on training time (left), imitation error (center), and success rate (right).}
% \label{fig:result}
% \end{figure*}

% In the future, we can explore other non-polynomial functions to verify if performance can be further improved.
\textcolor{black}{
\section{An additional baseline - State-action Behaviour Cloning Policy}
\label{sec:state_action_BC}
We conducted an additional experiment involving a new neural network based baseline policy that learns the mapping directly from states $\mathrm{x}(t)$ to actions $\mathrm{\tau}(t)$ instead of learning the reference dynamics and the tracking controller. The new policy (State-action BC) was built upon the MLP architecture with three hidden layers ([64, 128, 64]), and was trained over three random seeds to minimize the state-action reproduction error. For a fair comparison, these policies were trained and tested on the same set of demonstrations and testing instances as in Section.~\ref{sec:sample_efficiency}.
\begin{figure*}[t]
     \centering
%%%%%%%%%%%%%%%%%%%%%%%%%%%%%%%%%%%%first row
        \begin{subfigure}[t]{0.24\textwidth}
        \raisebox{-\height}{\includegraphics[width=\textwidth]{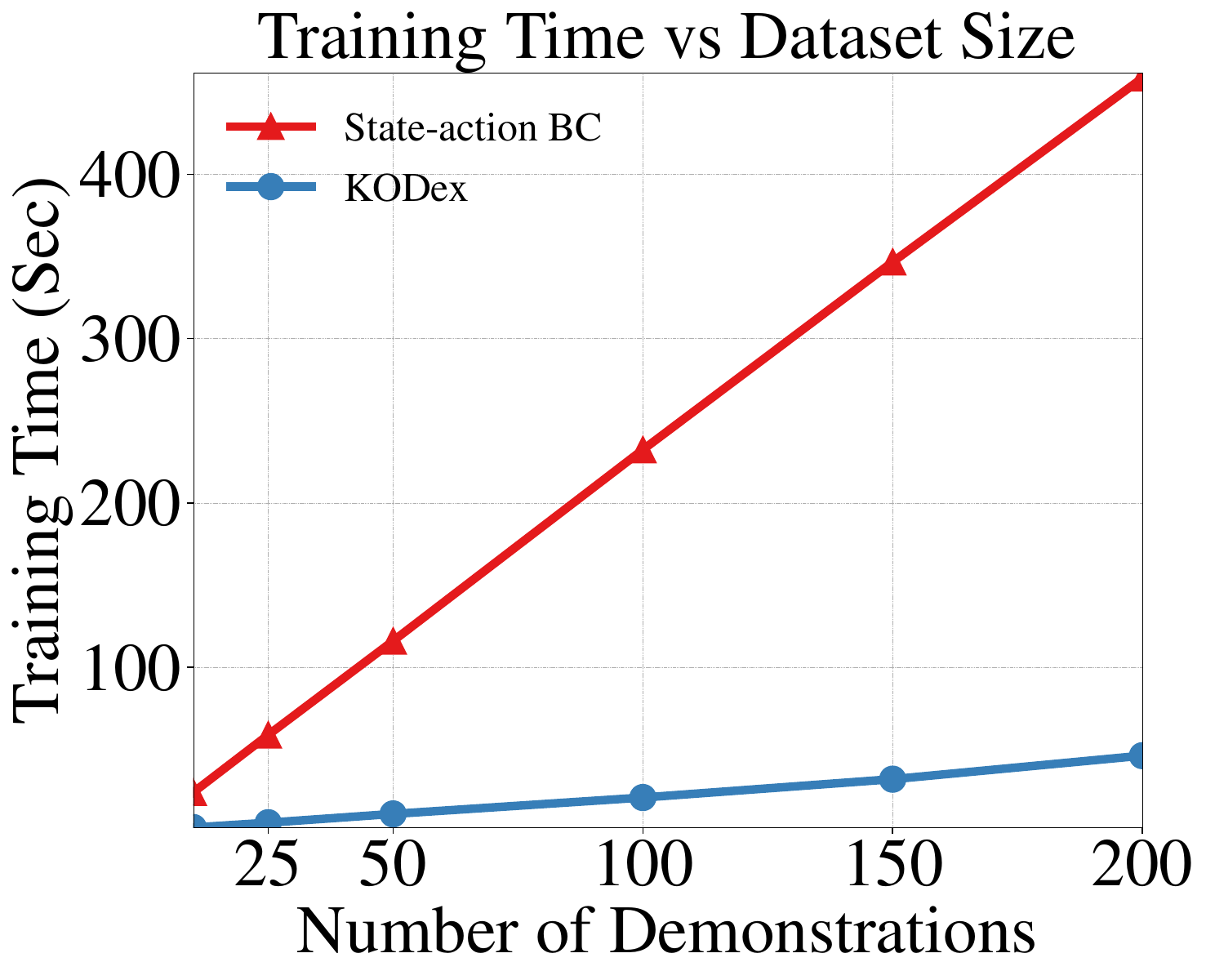}}
    \end{subfigure}
        \begin{subfigure}[t]{0.24\textwidth}
        \raisebox{-\height}{\includegraphics[width=\textwidth]{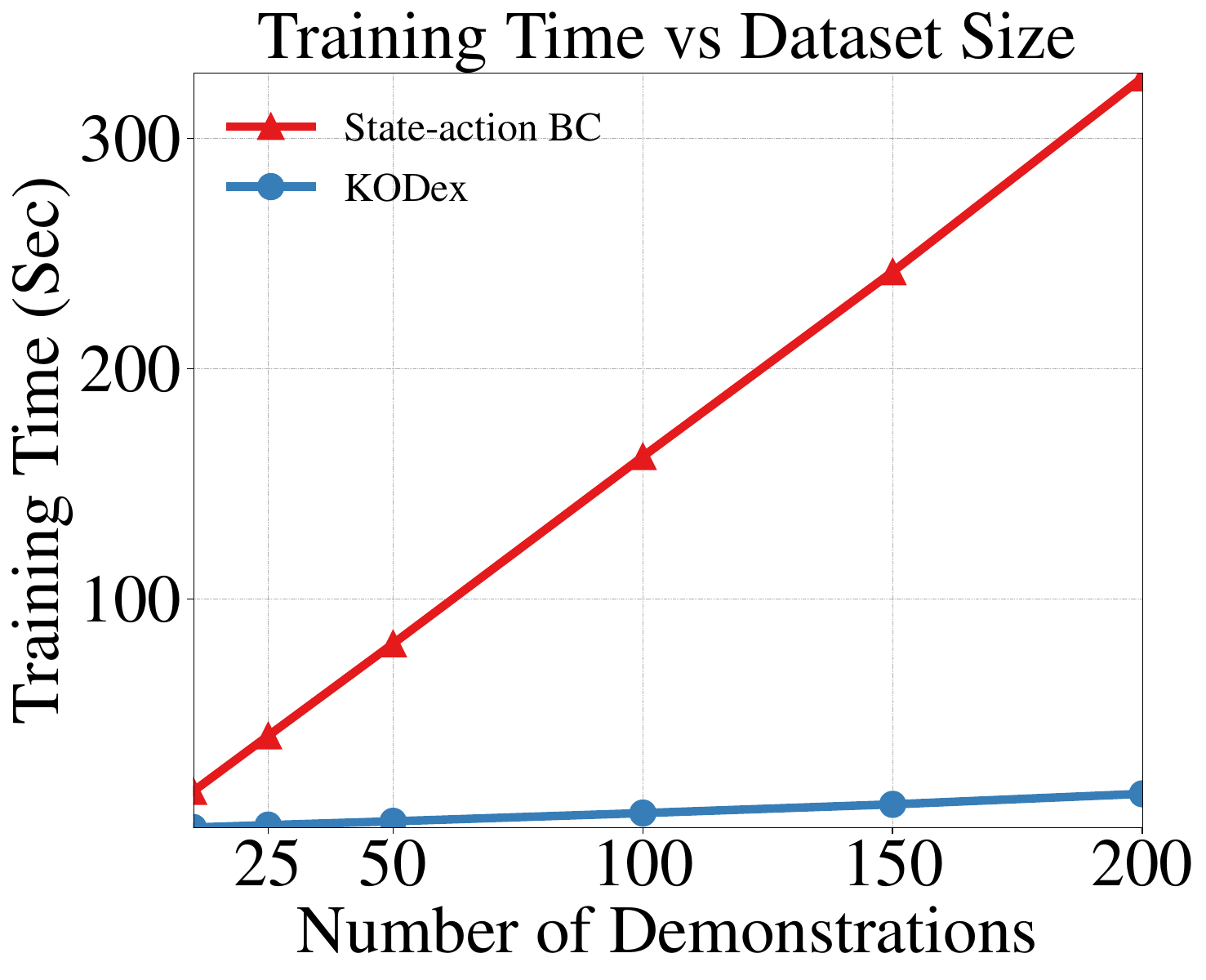}}
    \end{subfigure}
    \begin{subfigure}[t]{0.24\textwidth}
        \raisebox{-\height}{\includegraphics[width=\textwidth]{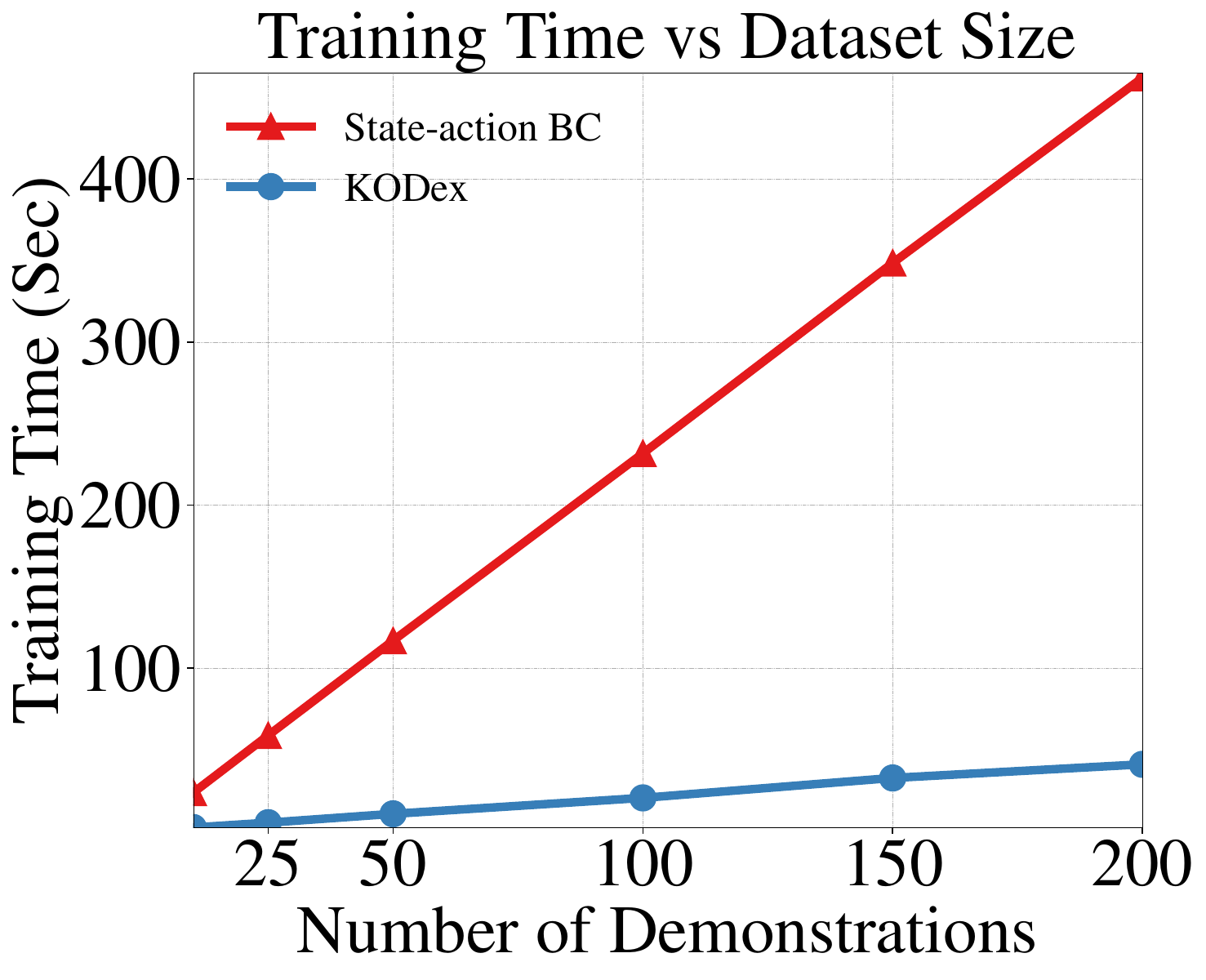}}
    \end{subfigure}
    \begin{subfigure}[t]{0.24\textwidth}
        \raisebox{-\height}{\includegraphics[width=\textwidth]{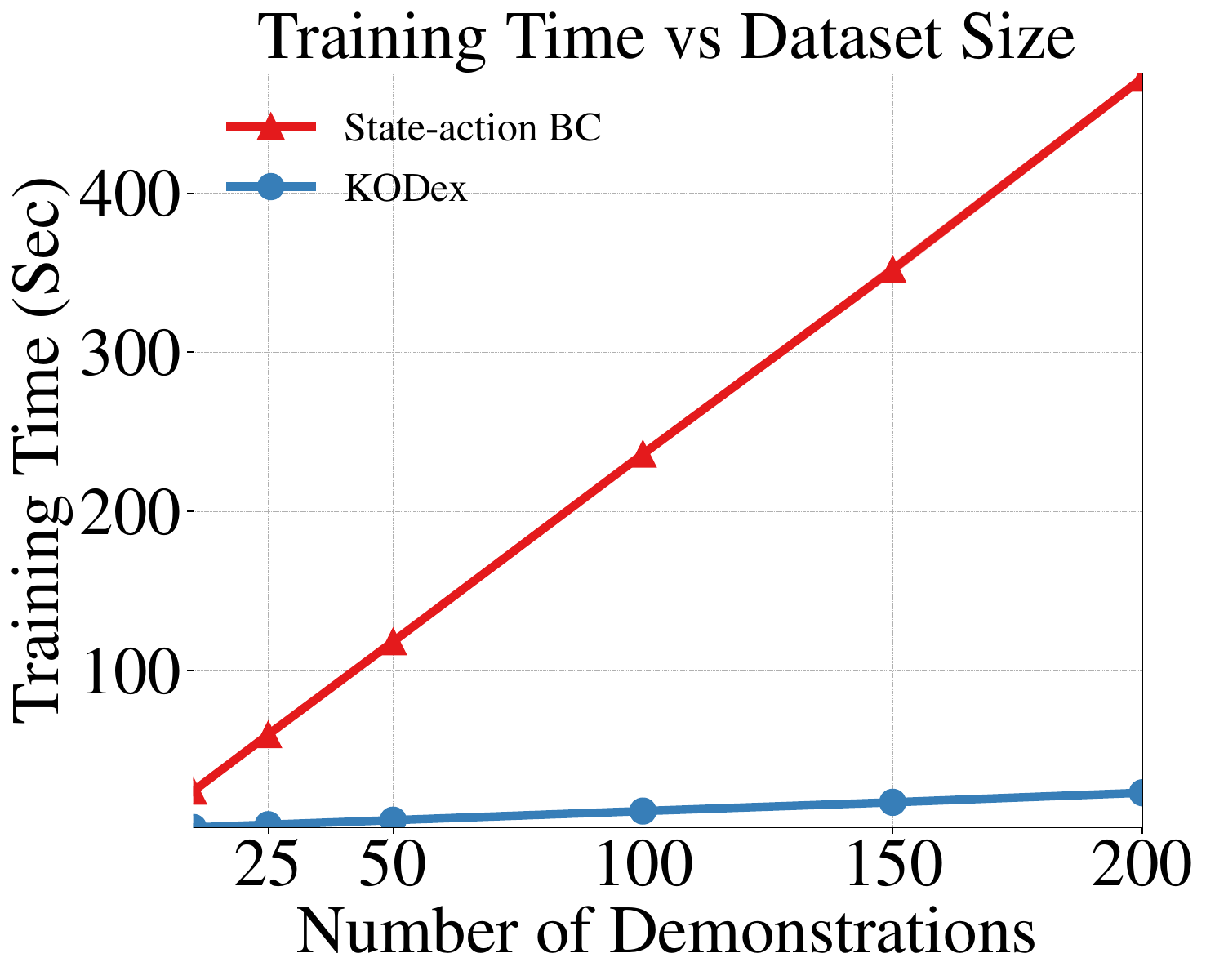}}
    \end{subfigure}
        %%%%%%%%%%%%%%%%%%%%%%%%%%%%%%%%%%%%third row
        \begin{subfigure}[t]{0.24\textwidth}
        \raisebox{-\height}{\includegraphics[width=\textwidth]{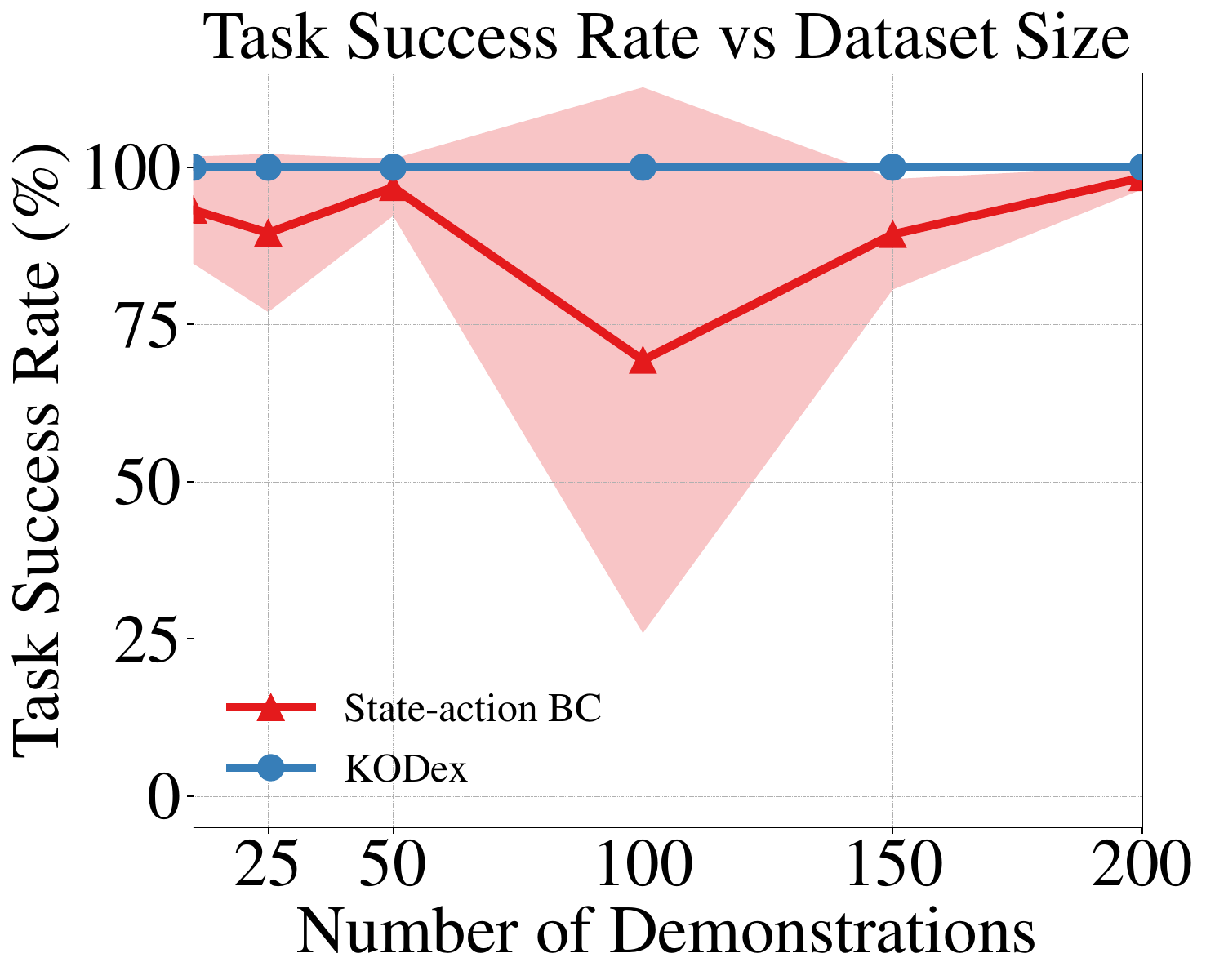}}
        \caption{Tool Use (Hammer)} 
    \end{subfigure}
        \begin{subfigure}[t]{0.24\textwidth}
        \raisebox{-\height}{\includegraphics[width=\textwidth]{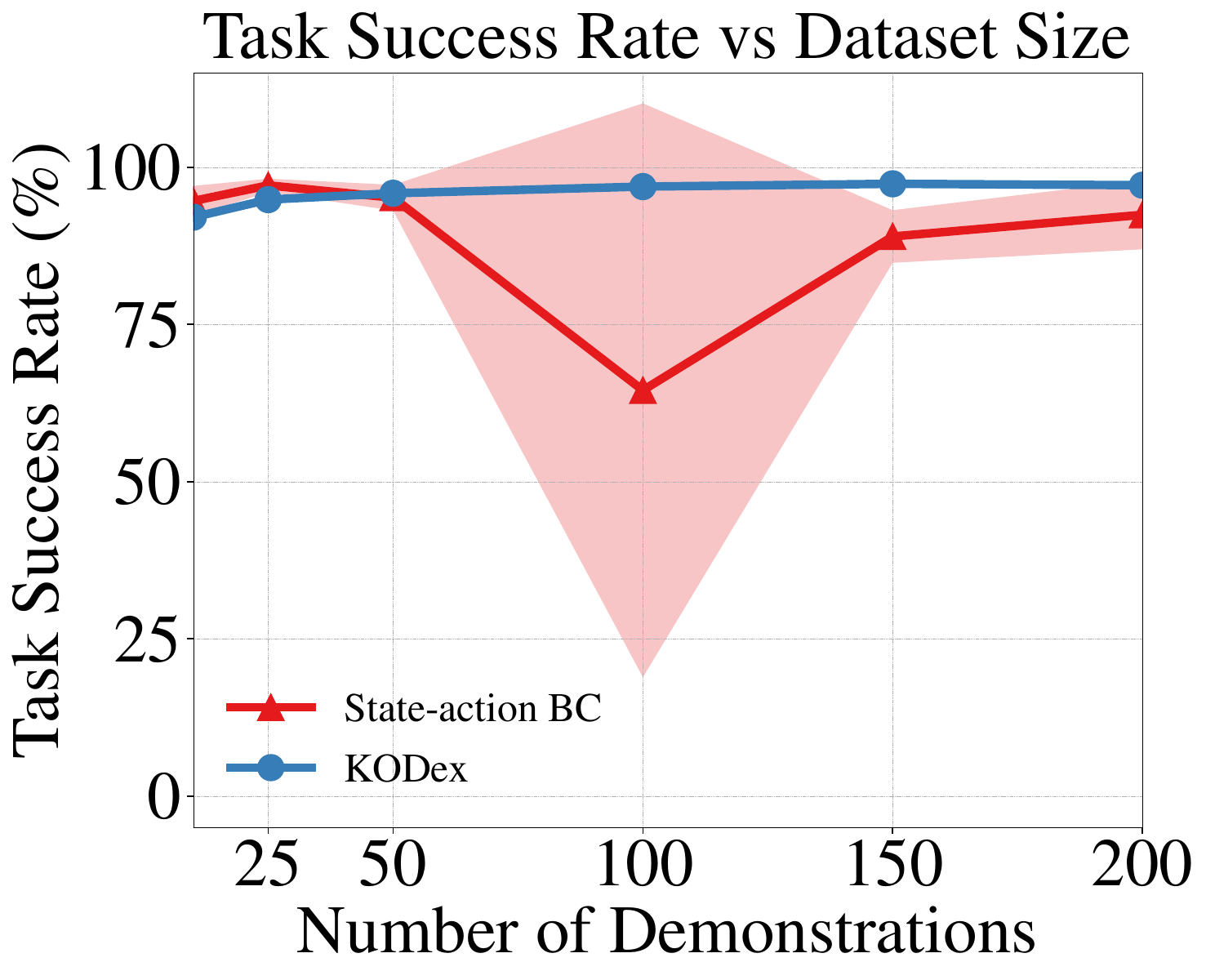}}
        \caption{Door Opening} 
    \end{subfigure}
    \begin{subfigure}[t]{0.24\textwidth}
        \raisebox{-\height}{\includegraphics[width=\textwidth]{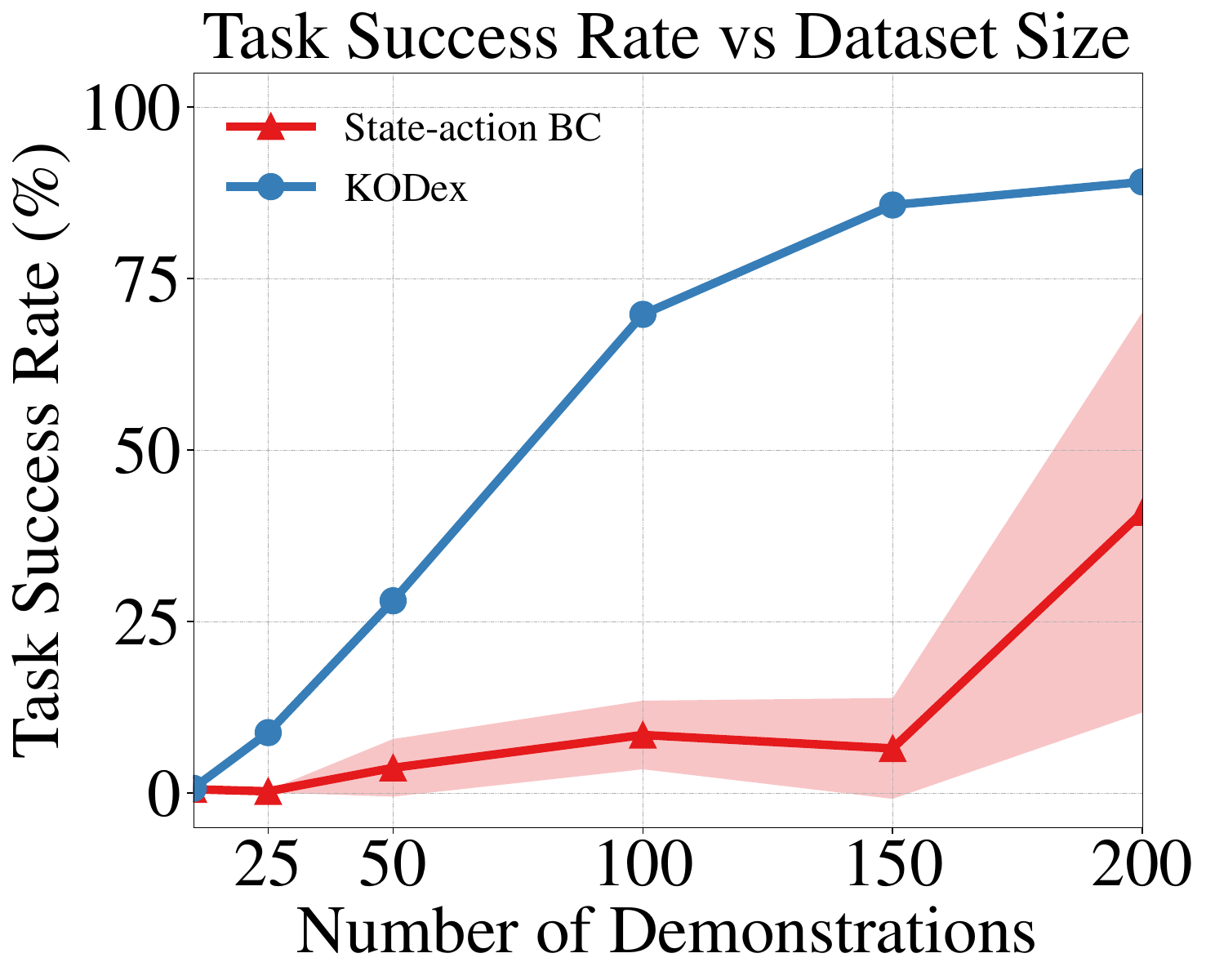}}
        \caption{Object Relocation} 
    \end{subfigure}
    \begin{subfigure}[t]{0.24\textwidth}
        \raisebox{-\height}{\includegraphics[width=\textwidth]{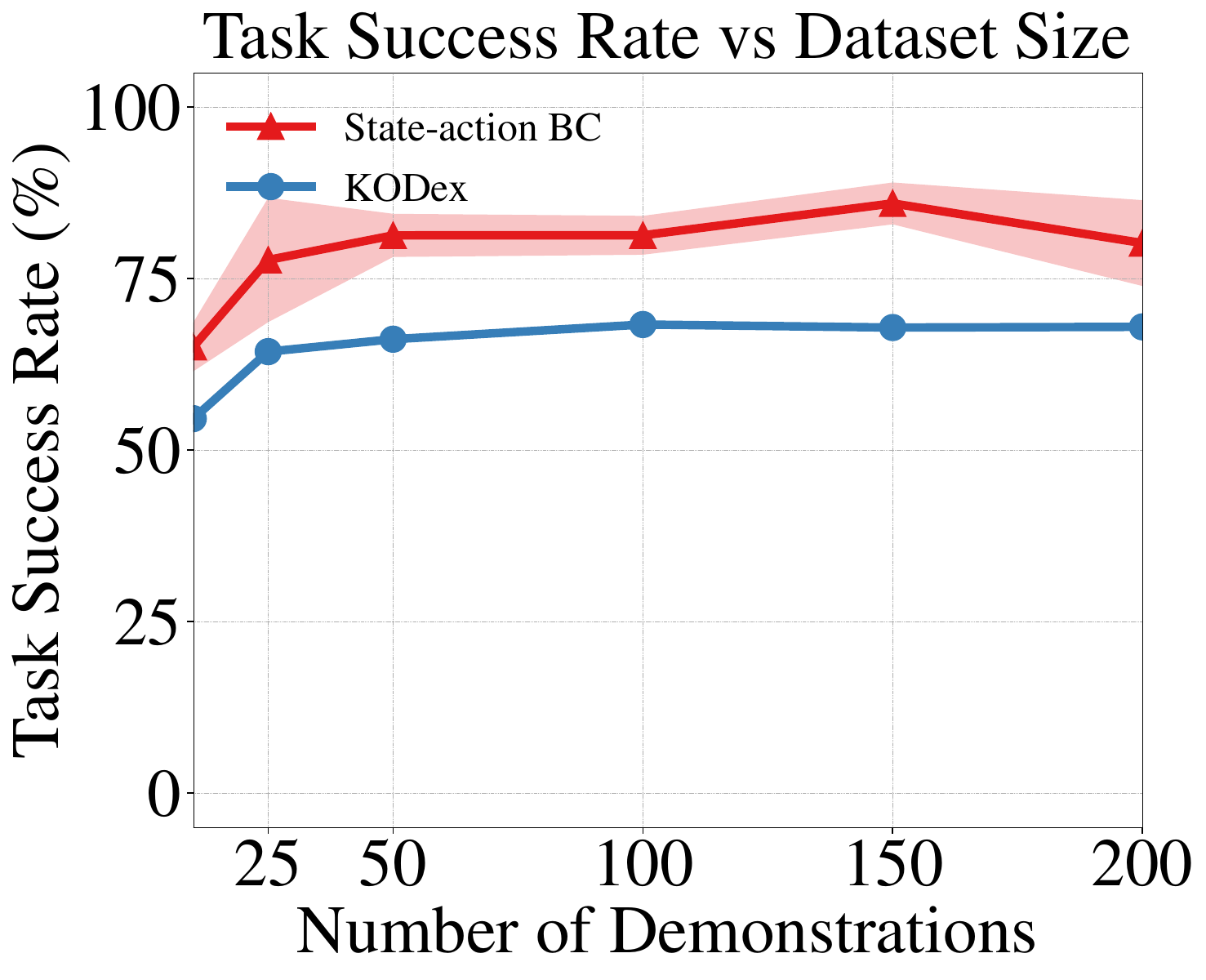}}
        \caption{In-hand Reorientation} 
    \end{subfigure}
    \caption{The effects of number of demonstrations on training time (top row), and success rate (bottom row) for KODex and State-action BC on each task. Solid lines indicate mean trends and shaded areas show $\pm$ standard deviation over three random seeds. 
    % Note that we did not plot NN's imitation errors as they were out of scale.
    }
\label{fig:new_baseline}
\end{figure*}
\newline
In Fig.~\ref{fig:new_baseline}, we report the training time and the task success rate on each task for KODex and State-Action BC.
The results reveal a familiar trend: across all tasks, KODex is drastically more computationally efficient than State-Action BC, while performing comparably  (if not better) in terms of success rate.
% \newline
}

\textcolor{black}{
In addition, we would like to highlight two other advantages of KODex over the state-action policies: 1). KODex could be potentially applied on state-only demonstrations, with a manually-tuned PD controller replacing the learned tracking controller (one could also learn the controller via reinforcement learning \cite{peng2020learning}). On the contrary, state-action imitation learning methods inevitably need action labels. 2). KODex is safer for online execution. Since KODex separates the motion generation and tracking, it tends to take less risky actions. Due to the curse of covariate shift, the state-action policies may make problematic decisions when they encounter unseen states caused by perturbations. In Fig.~\ref{fig:pen_throw_out}, we report the pen throw-out rate from the In-hand Reorientation task. It can be seen that the State-action BC policy is more likely to generate undesirable behaviours, resulting in complete task failures. This implies that KODex may be safer for hardware implementations, thanks to the separation of reference motion and tracking. Although there are a few other state-action policies that better address covariate shift (e.g., GAIL \cite{ho2016generative}), such comparisons are outside of scope of this work.
\begin{figure*}[t]
     \centering
        \begin{subfigure}[t]{0.5\textwidth}
        \raisebox{-\height}{\includegraphics[width=\textwidth]{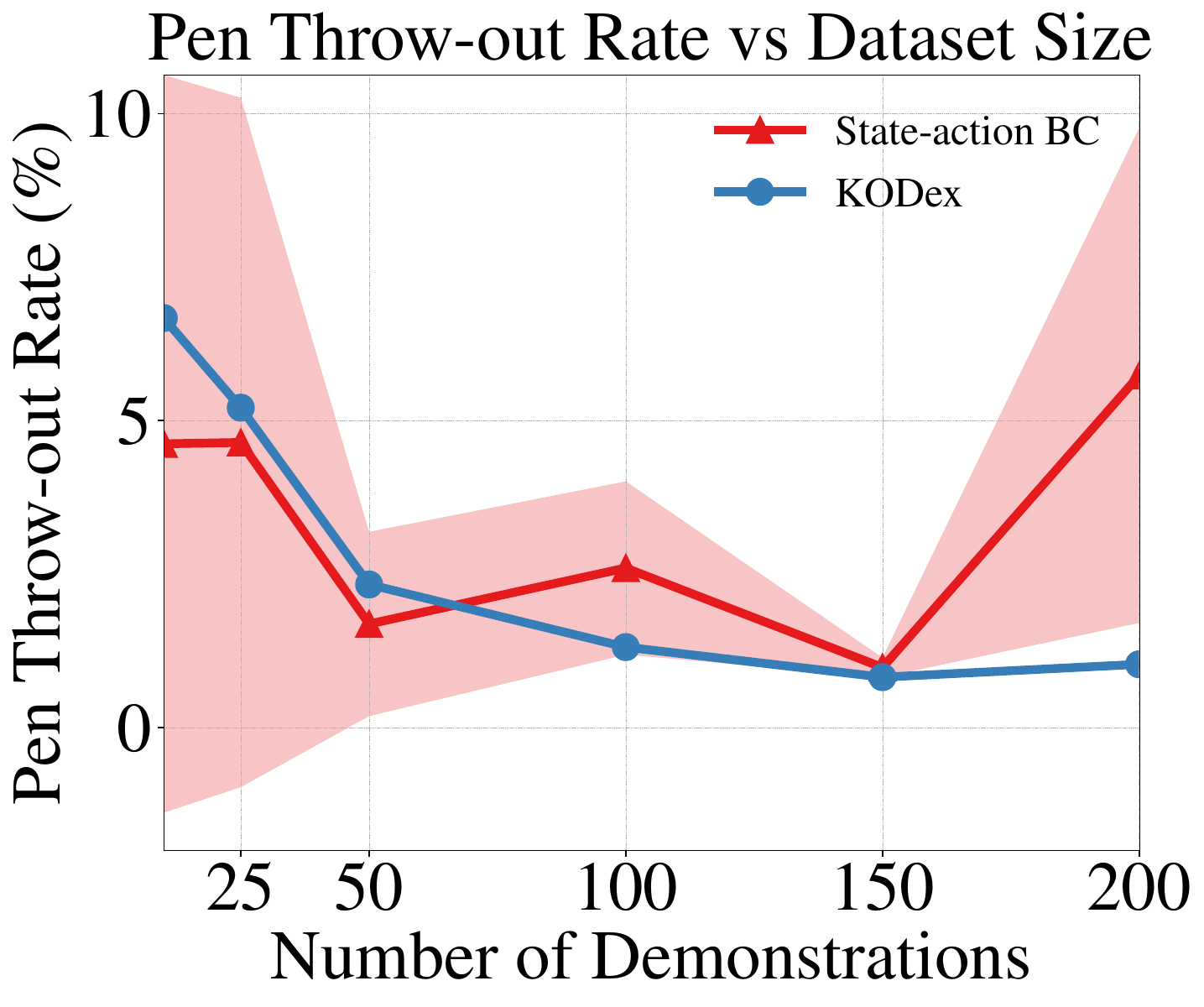}}
    \end{subfigure}
    \caption{The effects of number of demonstrations on pen throw-out rate for KODex and State-action BC.}
\label{fig:pen_throw_out}
\end{figure*}
}
\end{document}